\documentclass{article}

\PassOptionsToPackage{numbers, compress}{natbib}

\usepackage[preprint]{neurips_2024}

\usepackage[utf8]{inputenc} 
\usepackage[T1]{fontenc}    
\usepackage{hyperref}       
\usepackage{url}            
\usepackage{booktabs}       
\usepackage{amsfonts}       
\usepackage{nicefrac}       
\usepackage{microtype}      
\usepackage{xcolor}         
\usepackage{xspace}
\usepackage{fixltx2e} 
\usepackage{graphicx}
\usepackage{caption}
\usepackage{subcaption}
\usepackage{multirow}
\usepackage{colortbl}
\usepackage{tcolorbox}
\usepackage{inconsolata}
\usepackage{amsmath}
\usepackage{arydshln}
\usepackage{enumitem}
\usepackage{listings}
\usepackage{fancyvrb}
\usepackage{framed}
\usepackage{algorithm}
\usepackage{algorithmic}
\usepackage{amsmath}
\usepackage{authblk}

\definecolor{lightred}{rgb}{1, 0.8, 0.8}
\definecolor{lightredalpha}{rgb}{1, 0.8, 0.8}
\colorlet{lightredalpha}{lightred!50}

\definecolor{lightblue}{rgb}{0.8, 0.8, 1}
\definecolor{lightbluealpha}{rgb}{0.8, 0.8, 1}
\colorlet{lightbluealpha}{lightblue!50}

\newcommand{\agent}{$\textsc{Husky}$\xspace}
\newcommand{\dataset}{$\textsc{HuskyQA}$\xspace}

\newcommand{\critic}{$\textsc{Critic}$\xspace}
\newcommand{\finma}{$\textsc{FinMA}$\xspace}
\newcommand{\fireact}{$\textsc{FireAct}$\xspace}
\newcommand{\chameleon}{$\textsc{Chameleon}$\xspace}
\newcommand{\lila}{$\textsc{Lila}$\xspace}
\newcommand{\lumos}{$\textsc{Lumos}$\xspace}
\newcommand{\react}{$\textsc{ReAct}$\xspace}
\newcommand{\rewoo}{$\textsc{ReWOO}$\xspace}

\usepackage{dsfont}

\definecolor{verylightgray}{rgb}{0.9,0.9,0.9}

\title{
\includegraphics[width=0.035\textwidth, scale=0.001, clip, trim=5cm 7.0cm 5cm 2.8cm]{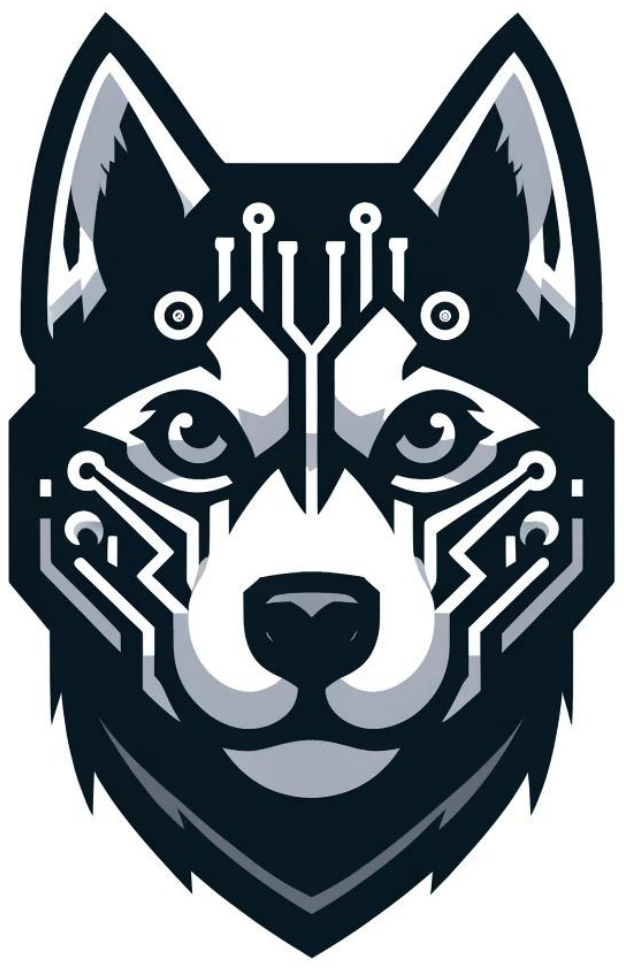}
\agent: A Unified, Open-Source Language Agent for Multi-Step Reasoning}

\author{%
\textbf{Joongwon Kim}$^{\dagger}$ \quad \textbf{Bhargavi Paranjape}$^{\ddagger}$ \quad \textbf{Tushar Khot}$^\S$ \quad \textbf{Hannaneh Hajishirzi}$^{\dagger\S}$ \\
$^\dagger$University of Washington \quad $^\ddagger$Meta AI \quad $^\S$Allen Institute for AI\\
}

\begin{document}

\maketitle

\begin{figure}[!h]
    \centering
    \includegraphics[scale=0.55, clip, trim=3cm 3.4cm 3cm 3.9cm]{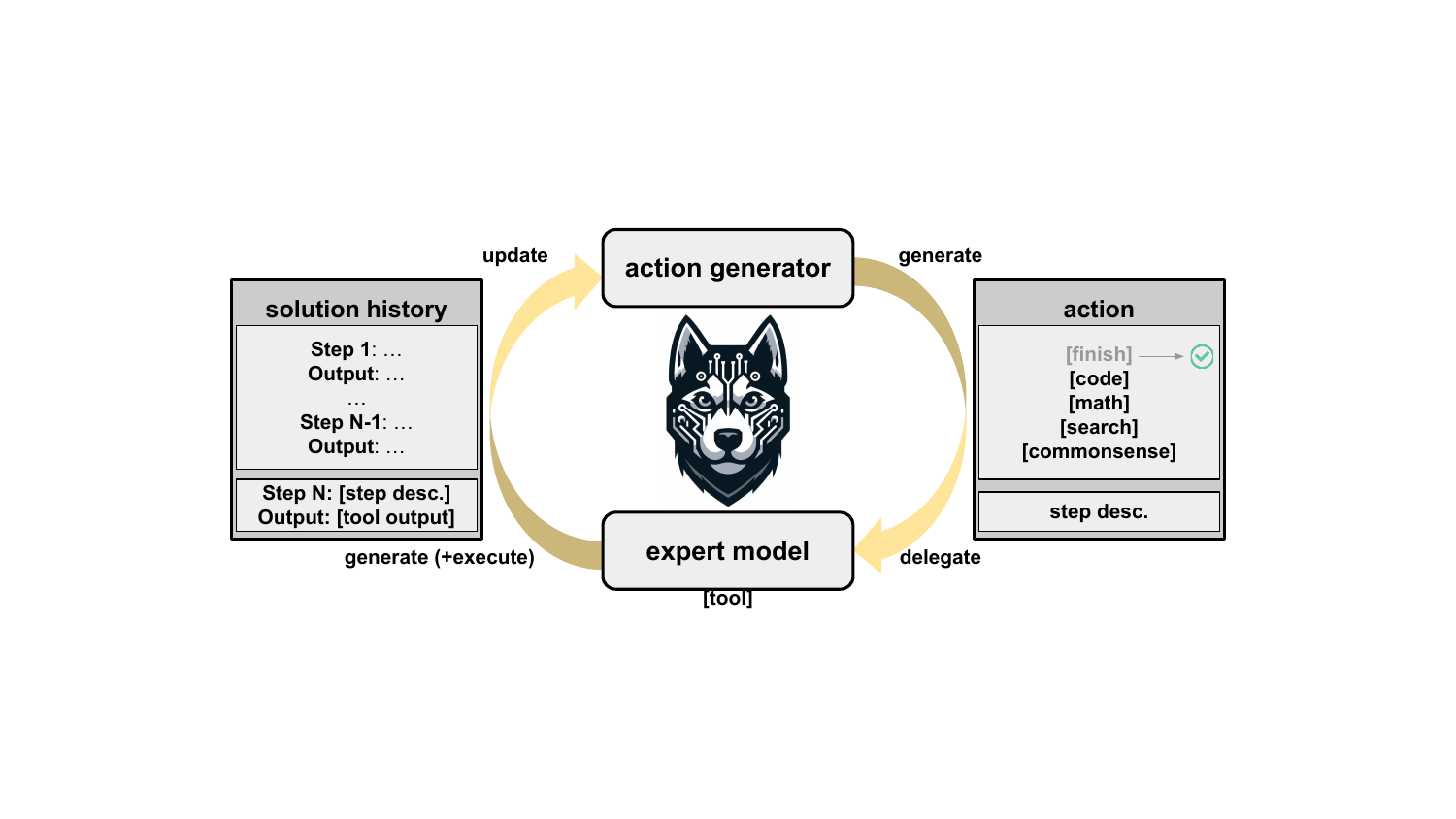}
    \caption{
    Schematic of \agent. \agent iterates between \textit{action generation} where it generates a tool call and the corresponding high-level step description, and \textit{action execution} where it uses the tool-associated expert model to execute the action, repeating this until it arrives at the terminal state.
    }
    \label{fig:husky-teaser}
\end{figure}

\begin{abstract}
Language agents perform complex tasks by using tools to execute each step precisely. 
However, most existing agents are based on proprietary models or designed to target specific tasks, such as mathematics or multi-hop question answering.
We introduce \agent, a holistic, open-source language agent that learns to reason over a unified action space to address a diverse set of complex tasks involving  numerical, tabular, and knowledge-based reasoning.
\agent iterates between two stages: 1) generating the next action to take towards solving a given task  
and 2) executing the action using expert models and updating the current solution state.
We identify a thorough ontology of actions for addressing complex tasks and curate high-quality data to train expert models for executing these actions.
Our experiments show that \agent outperforms prior language agents across 14 evaluation datasets.
Moreover, we introduce \dataset, a new evaluation set which stress tests language agents for mixed-tool reasoning, with a focus on retrieving missing knowledge and performing numerical reasoning.
Despite using 7B models, \agent matches or even exceeds frontier LMs such as GPT-4 on these tasks, showcasing the efficacy of our holistic approach in addressing complex reasoning problems. 
Our code and models are available at \url{https://github.com/agent-husky/Husky-v1}.

\end{abstract}

\section{Introduction}
\label{sec:intro}
\begin{figure}
    \centering
    \includegraphics[scale=0.55, clip, trim=0cm 0cm 0.5cm 1cm]{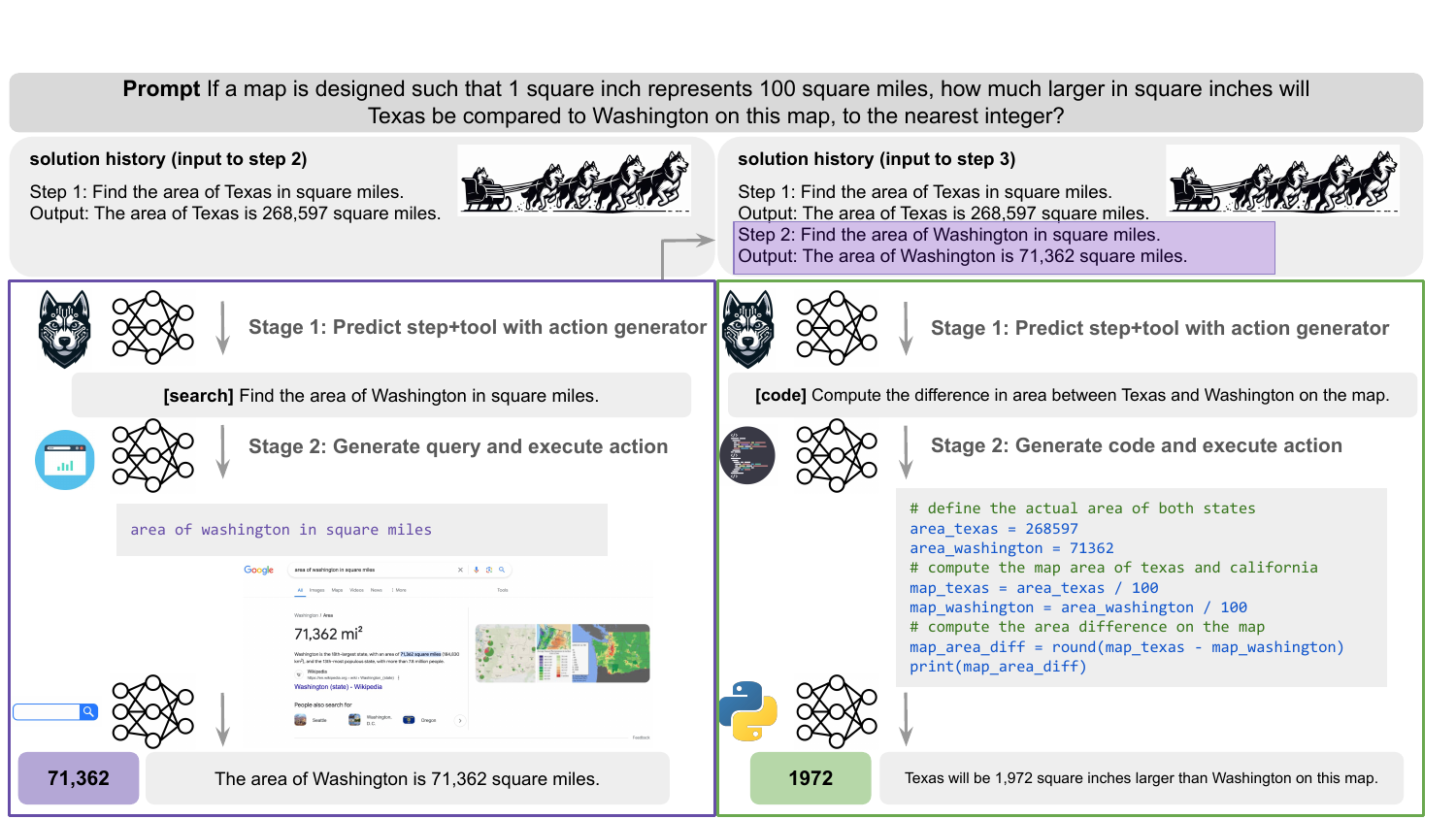}
    \caption{
    Overview of \agent.
    \agent solves multi-step tasks for numerical, tabular and knowledge-based reasoning by jointly predicting the next high-level step and tool with an action generator, and executing the action with the assigned expert model. 
    This process repeats until it arrives at the final answer.
    As shown above, \agent employs multiple LMs in parallel to solve a complex task, with the action generator coordinating the expert models, similar to how several Huskies pull a sleigh together.
    }
    \label{fig:husky-overview}
\end{figure}
Recent advances in the capabilities of large language models (LLMs) have led to the development of \textit{language agents} to address complex, multi-step tasks~\cite{DBLP:conf/icml/HuangAPM22, DBLP:conf/iclr/YaoZYDSN023, DBLP:journals/corr/abs-2305-18323, DBLP:conf/nips/ShinnCGNY23, DBLP:journals/corr/abs-2310-04406}.
As LLMs are limited in their computational or factual accuracies~\cite{DBLP:conf/naacl/PatelBG21, DBLP:conf/acl/MallenAZDKH23, DBLP:conf/emnlp/MinKLLYKIZH23}, language agents provide a compelling alternative by interacting with external tools to solve complex tasks step-by-step~\cite{DBLP:journals/corr/abs-2305-11738, DBLP:journals/corr/abs-2307-16789, DBLP:conf/nips/HaoLWH23}.
A significant number of existing language agents employ proprietary models through API calls~\cite{DBLP:journals/corr/abs-2303-09014, DBLP:conf/nips/0001ST00Z23, DBLP:conf/nips/LuPCGCWZG23}, which are difficult to scale due to the associated costs and latencies.
Meanwhile, recent open-source LLMs focus on specific tasks such as multi-hop QA~\cite{DBLP:journals/corr/abs-2310-05915, DBLP:journals/corr/abs-2402-11451} or employ convoluted procedures for training data curation and agent inference~\cite{, DBLP:journals/corr/abs-2311-05657, DBLP:journals/corr/abs-2401-05268}.
In this work we introduce \textbf{\agent}, a unified, open-source language agent that solves complex, multi-step reasoning tasks by decomposing each task into a series of executable actions and performing each action with a tool until it reaches a terminal state.
Unlike other previous language agents that build on open LMs~\cite{DBLP:journals/corr/abs-2310-05915, DBLP:journals/corr/abs-2311-05657}, \agent provides a distinctively generalizable yet efficient approach to train and deploy open language agents on a wide variety of tasks while maintaining a unified action space.

\agent jointly addresses numerical, tabular and knowledge-based reasoning tasks in two stages: 1) \textit{action generation} where the action generator predicts the action which consists of the high-level step to take and the tool to execute, and 2) \textit{action execution} where the designated expert model and tool performs the action and updates the solution state.
Our agent uses highly performant 7B LMs~\cite{DBLP:journals/corr/abs-2307-09288, DBLP:journals/corr/abs-2311-10702, DBLP:journals/corr/abs-2401-14196, DBLP:journals/corr/abs-2402-03300} to initialize the action generator and expert models in order to test the current limits of open language agents.
\agent, as it updates its solution state with a pre-defined ontology of actions until it arrives at a terminal state, can be viewed as a modern, LLM-based reformulation of classical planning systems such as STRIPS~\cite{DBLP:journals/ai/FikesN71}.

As shown in Figure~\ref{fig:husky-teaser}, \agent iterates between the two stages until it arrives at a terminal state.
The first module in \agent is the \textit{action generator}. 
Given the input question and the solution generated so far, the action generator jointly predicts the next high-level step to take and the associated tool.
The tools forming the ontology of our actions are \textbf{[code]}, \textbf{[math]}, \textbf{[search]} and \textbf{[commonsense]}.
If the final answer to the question has been reached in the solution history, then the action generator returns the answer.
Based on the tool assigned by the action generator, \agent calls the corresponding tool, executes the tool and re-writes the tool outputs optionally into natural language.
Each tool is associated with an expert model - a \textit{code generator} for \textbf{[code]}, a \textit{math reasoner} for \textbf{[math]}, a \textit{query generator} for \textbf{[search]} and a \textit{commonsense reasoner} for \textbf{[commonsense]}.

To train \agent, we use a simplified and generalizeable pipeline where we few-shot prompt a teacher model to generate tool-integrated solution trajectories, which are crucial to endowing tool use abilities to \agent. 
We rearrange our set of solution trajectories to build training data for the action generator and all expert models in \agent.
Our inference procedure requires no assumption about a specific task, allowing us to use the same action generator and expert models across all downstream tasks.

Additionally, we extensively evaluate \agent using tasks that require leveraging multiple tools.
We present decontextualized versions of DROP~\cite{DBLP:conf/naacl/DuaWDSS019} and IIRC~\cite{DBLP:conf/emnlp/FergusonGHKD20} where the question is re-written to be understood without its passage in the original dataset.
Moreover, we build \dataset, a new benchmark that tests the abilities of models and agents to combine knowledge-based and numerical reasoning to solve complex queries.
\agent matches or outperforms frontier models~\cite{DBLP:journals/corr/abs-2303-08774} on these evaluation sets, demonstrating existing gaps in frontier models for mixed-tool reasoning and the potential of open language agents that are trained on high-quality instances of planning and tool use.

Our experiments indicate that \agent generalizes across multiple tasks better than other language agents including \fireact~\cite{DBLP:journals/corr/abs-2310-05915} and \lumos~\cite{DBLP:journals/corr/abs-2311-05657}, and outperforms other agents in tasks of their own expertise.
For example, \agent outperforms \lumos on GSM-8K~\cite{DBLP:journals/corr/abs-2110-14168} by more than 20 points and \fireact on HotpotQA~\cite{DBLP:conf/emnlp/Yang0ZBCSM18} by 5 points.
\agent also outperforms \finma~\cite{DBLP:journals/corr/abs-2306-05443} on FinQA~\cite{DBLP:conf/emnlp/ChenCSSBLMBHRW21} by 9 points and \critic-70B~\cite{DBLP:journals/corr/abs-2305-11738} on TabMWP~\cite{DBLP:conf/iclr/Lu0CWZRCK23} by 1.8 points.
Finally, \agent outperforms CoT-prompted \texttt{gpt-4-0125-preview}~\cite{DBLP:journals/corr/abs-2303-08774} on our decontextualized subsets of DROP and IIRC by 3 and 5 points, using only 7B models.
On \dataset, \agent with a 13B action generator scores within 1 points behind \texttt{gpt-4o}.
Our experiments also hint that \agent can be extended to address a wider variety of tasks without performance loss by scaling the action space and the number of expert models.
These results showcase our robust recipe for developing \agent, an open-source language agent that generalizes and achieves competitive performance across a wide array of multi-step reasoning tasks.




\section{Related Work}
\label{sec:related-work}
\paragraph{Language agents.}
Language agents have recently gained popularity for solving complex tasks.
These agents utilize language models to perform specialized tasks such as generating a high-level plan to solve the task~\cite{DBLP:journals/corr/abs-2305-18323, DBLP:conf/nips/LuPCGCWZG23}, or assigning a specific tool to execute the given step of the solution~\cite{DBLP:conf/iclr/YaoZYDSN023, DBLP:journals/corr/abs-2311-05657}.
Such agents can largely be classified into two categories depending on the property of the underlying LM used to perform planning and tool assignment -- closed-source and open-source agents.

Earlier generations of language agents have used proprietary LMs~\cite{DBLP:journals/corr/abs-2303-08774} to perform core reasoning operations for solving multi-step problems.
This includes generating a particular action to take given the current solution state~\cite{DBLP:conf/iclr/YaoZYDSN023, DBLP:journals/corr/abs-2303-09014}, writing entire sequences of tools to execute~\cite{DBLP:conf/nips/0001ST00Z23, DBLP:conf/nips/LuPCGCWZG23}, or performing self-reflection~\cite{DBLP:conf/nips/MadaanTGHGW0DPY23, DBLP:journals/corr/abs-2311-07961} to evaluate the solution generated so far.
Such agents harness the raw reasoning capabilities of proprietary models to decompose a complex task into simpler subtasks and faithfully execute each subtask.
However, these agent systems incur high costs and lack in efficiency due to the API costs and latencies associated with proprietary models.
The lack of controllability of such models also hinders scientific analyses into building performant and efficient language agent systems.

More recent work has addressed this limitation by adapting open-source LMs to perform reasoning operations for solving complex tasks~\cite{DBLP:journals/corr/abs-2310-05915, DBLP:journals/corr/abs-2311-05657, DBLP:journals/corr/abs-2401-05268}.
Such agents are often trained on solution trajectories generated by a teacher model~\cite{DBLP:journals/corr/abs-2309-17452, DBLP:journals/corr/abs-2310-03731, DBLP:journals/corr/abs-2310-05915, DBLP:journals/corr/abs-2311-05657}, effectively distilling the reasoning and tool use capabilities of the teacher into a smaller, open language model~\cite{DBLP:conf/naacl/WestBHHJBLWC22}.
While demonstrating success in addressing complex problems, current open agents are only specialized for specific datasets or task domains~\cite{DBLP:journals/corr/abs-2309-17452, DBLP:journals/corr/abs-2310-03731, DBLP:journals/corr/abs-2310-05915, DBLP:journals/corr/abs-2402-11451}, require multifaceted rounds of data curation~\cite{DBLP:journals/corr/abs-2401-05268} or involve multiple layers of inference with task-specific toolsets~\cite{DBLP:journals/corr/abs-2311-05657} which hampers generalizability and efficiency.
In contrast, \agent works across multiple tasks of varying domains and involves a very straightforward data curation procedure which can be handled via few-shot prompting.
Moreover, \agent, being loosely inspired by classical planning systems~\cite{DBLP:journals/ai/FikesN71}, features an intuitive two-stage inference pipeline which involves generating and executing actions taken from a carefully-defined ontology of tools.

\paragraph{Tool use.}
Language agents utilize tools to address specific tasks with better precision~\cite{DBLP:conf/nips/SchickDDRLHZCS23, DBLP:journals/corr/abs-2303-09014, DBLP:journals/corr/abs-2305-15334}.
This can involve using code interpreters for numerical computations~\cite{DBLP:journals/corr/abs-2309-17452, DBLP:journals/corr/abs-2310-03731} or retrievers/search engines for factual grounding~\cite{DBLP:journals/corr/abs-2310-11511, DBLP:journals/corr/abs-2312-10003}.
Agents are often evaluated on multi-step reasoning tasks such as math~\cite{DBLP:journals/corr/abs-2110-14168, DBLP:conf/nips/HendrycksBKABTS21} or question-answering ~\cite{DBLP:conf/emnlp/Yang0ZBCSM18, DBLP:journals/tacl/GevaKSKRB21}.
While such tasks measure multi-step reasoning abilities, in practice they can often be solved with single types of tools and are not suitable for measuring an agent's capability to utilize multiple tools.
Meanwhile, there exist benchmarks that feature thousands of highly task-specific APIs~\cite{DBLP:journals/corr/abs-2307-16789, DBLP:conf/emnlp/LiZ000YLHL23} or a broad scope of tools such as PDF readers~\cite{DBLP:journals/corr/abs-2311-12983}, but such benchmarks either 1) require shallow forms of reasoning due to the synthetic combination of different APIs, or 2) require a wide ontology of tools that brings the scope out of this work.
In this work we introduce a new set of evaluations to measure the ability of language agents to simultaneously retrieve missing information and perform complex numerical reasoning.

\section{\agent: A Modular Framework for Solving Multi-Step Reasoning Tasks}
\label{sec:overview}
We introduce \agent, with its overview shown in Figure~\ref{fig:husky-overview}.
\agent is a language agent which solves complex, multi-step reasoning tasks by iterating between two stages: 1) predicting the next action to take, and 2) executing the action with the designated expert model.
This process is repeated by \agent until the agent arrives at a terminal state.
During the first stage, the action generator $\mathcal{A}$ jointly predicts each step $s$ and tool $t$ for solving the task, with $t$ represented by special tokens as summarized in Table~\ref{tab:husky-tools}.
The second stage involves executing each step with an expert model, which is either 1) a model that directly generates the output to each step (\textbf{[math]}, \textbf{[commonsense]}), or 2) a model that generates the inputs to be executed by the actual tools (\textbf{[code]}, \textbf{[search]}).
The terminal state is recognized by $\mathcal{A}$ which returns the final answer if it is found in the solution history.

\begin{table}[]
    \centering
    \begin{tabular}{lllll}
        \toprule
        Tool & Model & Input & Output & Functionality \\\midrule
        \textbf{[code]} & $\mathcal{M}_c$ & $x$, $h$, $s$ & $c$ & Generates code for precise computations. \\
        \textbf{[math]} & $\mathcal{M}_m$ & $x$, $h$, $s$ & $o$ & Performs mathematical reasoning.\\
        \textbf{[search]} & $\mathcal{M}_q$ & $x$, $h$, $s$ & $q$ & Writes search query to retrieve knowledge. \\
        \textbf{[commonsense]} & $\mathcal{M}_r$ & $x$, $h$, $s$ & $o$ & Performs commonsense reasoning.\\
        \bottomrule
    \end{tabular}
    \vspace{0.5em}
    \caption{
    Tools used by \agent. 
    All tools accept the same input format (task instruction $x$, solution history $h$ and current step $s$). 
    The \textbf{[math]} and \textbf{[commonsense]} tools directly generate the output to the step, and the \textbf{[code]} and \textbf{[search]} tools generate inputs to be executed by the code interpreter and search engine, respectively.
    Note that $\mathcal{M}_m$, $\mathcal{M}_c$ and $\mathcal{M}_q$ are finetuned models, and $\mathcal{M}_r$ is not.
    }
    \label{tab:husky-tools}
\end{table}

\subsection{Problem Formulation and Overview}
\label{sec:husky-problem-overview}
Given a task instruction $x$ and the solution history $h$ generated so far, we train an action generator $\mathcal{A}$ to predict the next step $s$ and its associated tool $t$.
We also train expert models $\mathcal{M}_t$ which accept $x$, $h$ and $s$ as inputs and function as the tool $t$ or the preprocessors for $t$.
In this work we use $t \in \{c, m, q\}$, which are LM-based modules that generate code snippets, math solutions and search queries.
Each tool assigned by $\mathcal{A}$ is represented as a special token (Table~\ref{tab:husky-tools}), which is used to route \agent to the correct expert model for action execution.

\paragraph{Inference overview.} 
Figure ~\ref{fig:husky-overview} provides an overview of the inference procedure taken by \agent.
 During inference, \agent assumes that the action generator $\mathcal{A}$, the expert models $\mathcal{M}_m$, $\mathcal{M}_c$, $\mathcal{M}_q$ and the base reasoning model $\mathcal{M}_r$ are ready. 
Given the input $x$, \agent iterates over a two-stage pipeline.
The first stage involves $\mathcal{A}$ generating the next step $s_i$ to take along with its tool $t_i$ for step $i$, and the second stage involves using an expert model routed from $t_i$ to generate the output $o_i$.
The math and commonsense reasoners ($\mathcal{M}_m$, $\mathcal{M}_r$) directly generate the natural language output $o_i$, while the code and query generators ($\mathcal{M}_c$, $\mathcal{M}_q$) generate an input $c_i$ to a code interpreter, or $q_i$ to a search engine.
In the latter case, $\mathcal{M}_r$ is used to re-write the tool-executed output into the natural language output $o_i$.
After this stage, $s_i$ and $o_i$ are concatenated to the solution history from step $i-1$ such that $h_{i}$ = "$h_{i-1}\texttt{\textbackslash n}s_{i}\texttt{\textbackslash n}o_{i}$" and the process repeats again for step $i+1$ with $x$ and $h_i$ as inputs to $\mathcal{A}$.
This process repeats until $\mathcal{A}$ identifies the final answer to $x$ in $h$.

\begin{figure}
    \centering
    \includegraphics[scale=0.55, clip, trim=0.1cm 4.5cm 0.8cm 1.5cm]{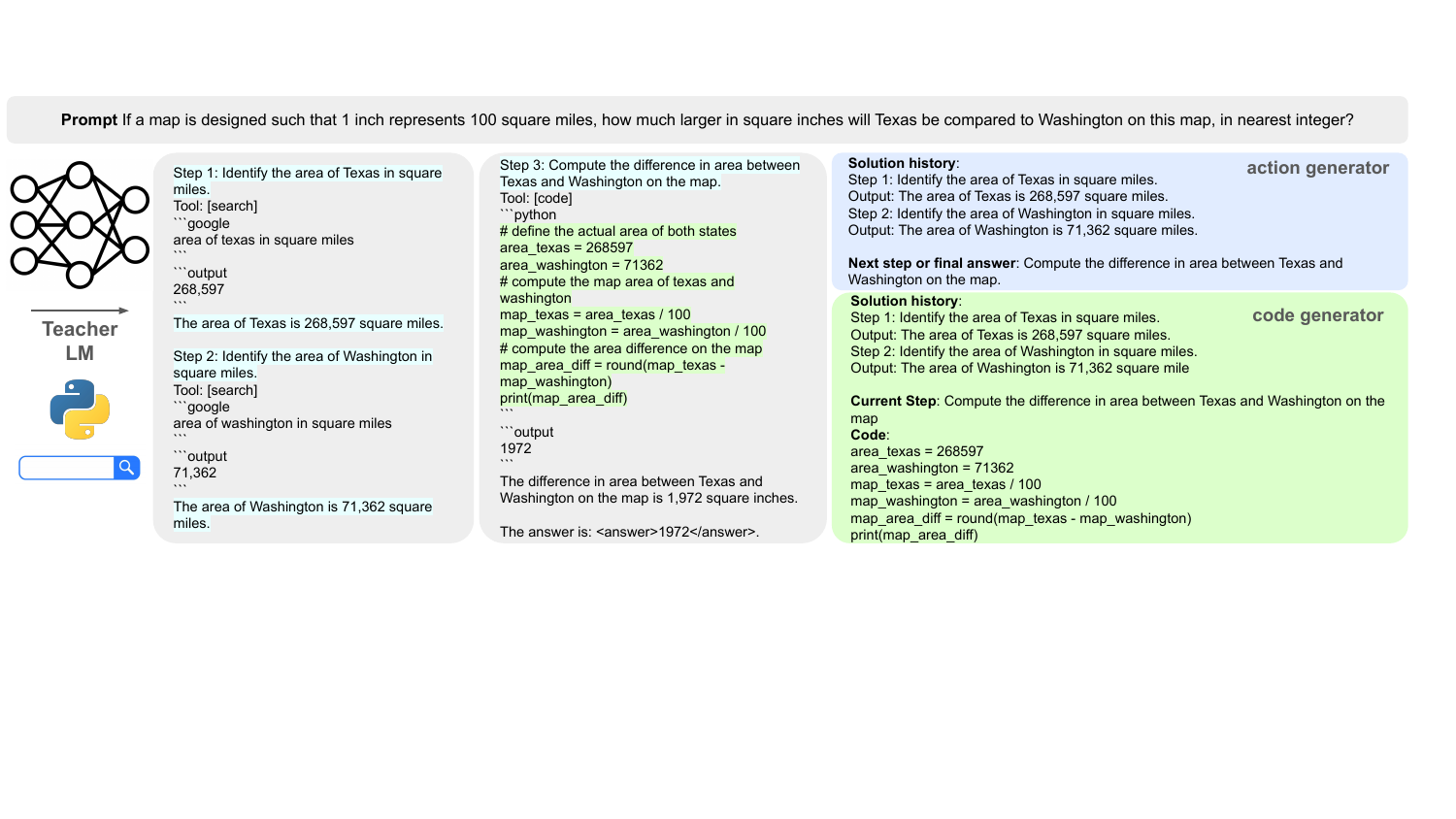}
    \caption{ Training data synthesis for \agent. A teacher LM is few-shot prompted to generate an initial trajectory for a question given in a training task. Then, each solution is parsed to extract steps and their outputs, which are used to construct training sets for $\mathcal{A}$, $\mathcal{M}_m$, $\mathcal{M}_c$ and $\mathcal{M}_q$.}
    \label{fig:husky-training} 
\end{figure}

\paragraph{Training overview.} 
All modules in \agent ($\mathcal{A}$, $\mathcal{M}_m$, $\mathcal{M}_c$ and $\mathcal{M}_q$) are trained using synthetic data (Section~\ref{sec:husky-training-synthetic-data})\footnote{We do not specifically train $\mathcal{M}_r$ to maintain its functionality as a general-purpose reasoner addressing any case where \textbf{[code]}, \textbf{[math]} or \textbf{[search]} are not used. Instead, we use few-shot prompting so that additional reasoning operations that appear with new tasks, such as table lookups, can be quickly added as demonstrations.}.
Refer to Figure~\ref{fig:husky-training} for a visualization of how the training data is constructed.
Given a set of seed training tasks, a teacher LM is first used to construct a set of tool-integrated solution trajectories.
Each step $i$ of the trajectory consists of the high-level step $s_i$ to execute, the associated tool $t_i$ chosen from the toolset in Table~\ref{tab:husky-tools}, the tool-executed output if applicable -- $c_i$ for code, $q_i$ for search -- and the natural language output $o_i$, which is directly generated for math and commonsense reasoning, and re-written from the tool-executed output for code and search.
For each input, our system generates the tool-integrated solutions until $t_i \in \{\text{\textbf{[code]}, \textbf{[search]}}\}$ is called.
Our system executes the tool and integrates its natural language output $o_i$ back into the solution trajectory until the solution reaches the final answer.
The solution trajectory is then used as a seed dataset from which training instances for $\mathcal{A}$, $\mathcal{M}_m$, $\mathcal{M}_c$ and $\mathcal{M}_q$ are extracted (Section~\ref{sec:husky-training-modules}).
For the action generator $\mathcal{A}$, the individual steps and their natural language outputs are extracted to form a training instance.
At step $i$, the prompt is formatted as a concatenation of $x$ and $h_{i-1}$ where $h_{i-1} = s_1\texttt{\textbackslash n}o_1\texttt{\textbackslash n}s_2\texttt{\textbackslash n}o_2\texttt{\textbackslash n}\cdots\texttt{\textbackslash n}s_{i-1}\texttt{\textbackslash n}o_{i-1}$
, and the completion is the next step $s_i$.
Meanwhile, for $\mathcal{M}_c$, $\mathcal{M}_m$ and $\mathcal{M}_q$, the prompt is formatted as a concatenation of $x$, $h_i$ and $s_i$, and the completion is a code snippet, math solution or a search query.

\subsection{\agent Training}
\label{sec:husky-training}
We describe in detail the procedure for obtaining synthetic data used to train ($\mathcal{A}$, $\mathcal{M}_m$, $\mathcal{M}_c$ and $\mathcal{M}_q$).
The training data is obtained in two stages: 1) synthesizing tool-integrated solution trajectories with a teacher LM, and 2) extracting training prompts for each module from the tool-integrated solutions.

\subsubsection{Synthesizing Tool-Integrated Solutions}
\label{sec:husky-training-synthetic-data}
As \agent is a new language agent framework, there does not exist training data to perform next step prediction and tool calls, along with code, math or search query generation for \agent.
Previous studies have demonstrated that smaller models can learn from distilling the generated outputs of larger \textit{teacher} models that have better reasoning capabilities \cite{DBLP:conf/naacl/WestBHHJBLWC22, DBLP:conf/acl/WangKMLSKH23} and have distilled their multi-step tool use abilities into open LMs to adapt them as language agents~\cite{DBLP:journals/corr/abs-2310-05915, DBLP:journals/corr/abs-2310-12823, DBLP:journals/corr/abs-2311-05657}.
Therefore, we follow existing approaches and utilize a teacher LM $\mathcal{T}$ to first synthesize tool-integrated solution trajectories for a given set of training tasks.
These high-quality solution trajectories serve as the blueprint from which we automatically extract and rearrange their components to build training sets for $\mathcal{A}$, $\mathcal{M}_m$, $\mathcal{M}_c$ and $\mathcal{M}_q$ -- we use each solution trajectory to synthesize training examples for multiple modules.

\textbf{Data synthesis.} Refer to Figure~\ref{fig:husky-training} for an example of the tool-integrated solution and the module-specific training data extracted from the trajectory.
Our solution trajectory is formatted as a sequence of steps $s_i$ ($i \in [1\ldots]$), the tool $t_i$ assigned to each step $i$, and the output $o_i$ from solving each step $i$. 

Each iteration of the solution begins with the high-level step to take, written as "Step $i$: $s_i$" for the $i$th step, which is then followed by a tool call formatted as "Tool: $t_i$".
The tool $t_i$ is chosen from Table~\ref{tab:husky-tools}.

If $t_i \in \{\text{\textbf{[code]}, \textbf{[search]}}\}$, it is accompanied by a code snippet $c_i$ or a search query $q_i$ generated by $\mathcal{T}$ and enclosed within their identifier tags \texttt{python} or \texttt{google} along with quotation marks.
$c_i$ or $q_i$ is then executed by a code interpreter or a search engine, and its output is rewritten by $\mathcal{T}$ into natural language.
For math or commonsense reasoning whose execution output is the LM-generated output, $t_i$ is directly followed by $\mathcal{T}$'s generated output.
We generate each solution with $\mathcal{T}$ until it identifies the final answer to $x$ with its identifier tag \texttt{<answer>}.

We few-shot prompt $\mathcal{T}$ to generate such tool-integrated solutions for all training instances in the set of training tasks we designate for developing \agent.
After generating the trajectories, we evaluate the final answer with respect to the label in the training set and keep the solution instances with the correct answer.
We provide example trajectories and few-shot prompts in Appendix~\ref{sec:tool-integrated-solution-trajectories} and~\ref{sec:solution-trajectory-prompts}.

\subsubsection{Training \agent modules}
\label{sec:husky-training-modules}
We use the tool-integrated solution trajectories as a blueprint to construct training data for all modules in \agent that require training.

\paragraph{Action generator.}
The input to the action generator $\mathcal{A}$ at step $i$ consists of the task instruction $x$ and the solution history $h_{i-1}$ concatenated as 
"$\{x$\}\texttt{\textbackslash n}$\{h_{i-1}\}$" $=$ "$x$\texttt{\textbackslash n}[$s_1$]\texttt{\textbackslash n}[$o_1$]\texttt{\textbackslash n}$\dots$\texttt{\textbackslash n}[$s_{i-1}$]\texttt{\textbackslash n}[$o_{i-1}$]".
Given this input, $\mathcal{A}$ generates an output "\textbf{[}$t_i$\textbf{]} $s_i$".
For example, if $x=$ "Find the birthday of the first president of the United States", then a possible output from $\mathcal{A}$ would be "\textbf{[search]} Identify the first president of the United States."
We extract the steps and their outputs from the solution trajectory via regex matching and organize them according to the format above to train $\mathcal{A}$.

\paragraph{Expert models.}
The input to each expert model consists of the task instruction $x$ and the solution history $h_{i-1}$, as well as the current step, $s_i$.
Given this input, $\mathcal{M}_c$ returns code snippet $c_i$, $\mathcal{M}_m$ returns a math solution $m_i$, and $\mathcal{M}_q$ returns a search query $q_i$ as their outputs.
For example, given the task instruction above, $\mathcal{M}_q$ would be trained to generate $q_i=$ \textit{first president of the united states}.
Again, we extract the steps and the tool-specific outputs from the solution trajectory and format them accordingly to build the training data for the expert models.

We finetune all \agent modules independently on their respective training datasets, jointly constructed from all training tasks, using the standard next token prediction objective.

\subsection{\agent Inference}
\label{sec:husky-inference}
We integrate the modules trained according to the procedure above into \agent to solve new, multi-step tasks at inference time.
Figure~\ref{fig:husky-overview} provides an example of \agent performing inference for a complex, multi-step task.

For task instruction input $x$, \agent first assigns the action generator $\mathcal{A}$ to generate the first step $s_1$ that needs to be taken, as well as its corresponding tool $t_1$.
With the example task given in Figure~\ref{fig:husky-overview}, $\mathcal{A}$ returns "\textbf{[search]} Find the area of Texas in square miles."
In many cases, the solution history $h$ is provided along with $x$ to $\mathcal{A}$.
For example, the solution history $h_1=$ "Step 1: Find the area of Texas in square miles.\texttt{\textbackslash n}Output: The area of Texas is 268,597 square miles" is provided to $\mathcal{A}$, which returns "\textbf{[search]} Find the area of Washington in square miles." for step 2.

Given the current step $s_i$ and tool $t_i$ generated by $\mathcal{A}$, \agent then uses the tool token $t_i$ to assign a expert model $\mathcal{M}$.
The connections between $t$ and $\mathcal{M}$ are: \textbf{[code]} token to the code generator $\mathcal{M}_c$, \textbf{[math]} token to the math reasoner $\mathcal{M}_m$, \textbf{[search]} token to the query generator $\mathcal{M}_q$ and \textbf{[commonsense]} token to the commonsense reasoner $\mathcal{M}_r$. 
While $\mathcal{M}_m$ and $\mathcal{M}_r$ directly generate the natural language outputs to the current step, $\mathcal{M}_c$ and $\mathcal{M}_q$ generate the code snippet and search query, which are then fed into a code interpreter and a search engine.
As the outputs from these code interpreters and web browsers are neither contextualized upon $x$ or $s$ nor are natural language, we few-shot prompt $\mathcal{M}_r$ to re-write the tool-executed output.
For example, given the step "Compute the difference in area between Texas and Washington on the map." in Figure~\ref{fig:husky-overview} and its Python output 1972, $\mathcal{M}_r$ generates "Texas will be 1,972 square inches larger than Washington on this map."

After an iteration of the two-stage process above, the current step $s_i$ and its natural language output $o_i$ is concatenated to the solution history $h_{i-1}$ to be fed as input to the next iteration of the inference along with $x$.
This procedure is repeated until $\mathcal{A}$ identifies the final answer to $x$ in its solution history $h_i$, at which point it arrives at the terminal state and returns "The answer is $a$." for the given answer $a$ and terminates the process.
In the example in Figure~\ref{fig:husky-overview}, $\mathcal{A}$ recognizes 1,972 to be the answer after \agent completes step 3 and returns 1,972 as the final answer.

\subsection{\agent Evaluation}
\label{sec:husky-evaluation}
Evaluating \agent is a process of performing inference with \agent modules upon complex, multi-step reasoning questions and scoring the final answers using the appropriate metric for each dataset.
While there are numerous datasets that involve multi-step reasoning in their solutions such as mathematics or multi-hop QA~\cite{DBLP:journals/corr/abs-2110-14168, DBLP:conf/nips/HendrycksBKABTS21, DBLP:conf/emnlp/Yang0ZBCSM18, DBLP:journals/tacl/TrivediBKS22}, not many datasets require the use of diverse tools. To complement this deficiency, we build \dataset, a numerical reasoning dataset with incomplete information that the language agent must first identify in order to solve the problem.

For example, consider the question \textit{If a map is designed such that 1 inch represents 100 square miles, how much larger in square inches will Texas be compared to Washington on this map, to the nearest integer?}.
While this question fundamentally requires mathematical reasoning to solve correctly, the information about the area of Texas and Washington is omitted and must be retrieved by the agent.

To build \dataset, we first generate synthetic, factoid questions from a seed list of 40 diverse, manually curated set of topics spanning economics, geography, sports, science, etc.
For the question above, the topic would be \textit{geography (state area)} and the factoid questions would be \textit{What is the area of Texas?} and \textit{What is the area of Washington?}
Then, we search the Web to identify the answer to each factoid question and convert the question-answer pair into a factoid statement.
According to the running example, the statements would be \textit{The area of Texas is 268,597 square miles.} and \textit{The area of Washington is 71,362 square miles.}
Finally, we few-shot prompt GPT-4 to generate synthetic questions from pairs of factoid statements in the same topic set.

The resulting questions are conditioned upon the information from the pair of factoid statements used to generate the question, and require mathematical reasoning at middle-school or high-school level to solve correctly.
We label all answers in \dataset automatically, and validate all our answers in the test set of 292 examples with human annotations.
We provide more details in Appendix~\ref{sec:husky-curation}.

We also provide decontextualized versions of a subset of DROP~\cite{DBLP:conf/naacl/DuaWDSS019} and IIRC~\cite{DBLP:conf/emnlp/FergusonGHKD20}, which we refer to as DROP* and IIRC*.
DROP and IIRC consist of questions contextualized on Wikipedia passages that require multi-step reasoning with knowledge-based and numerical operations.
We decontextualize each question by few-shot prompting \texttt{gpt-3.5-turbo} to integrate question-specific information from the passage.
More details regarding DROP* and IIRC* can be found in Appendix~\ref{sec:drop-iirc-details}.

\section{Experiments}
\label{sec:experiments}
\subsection{Tasks and Datasets}
\label{sec:husky-datasets}
We train and evaluate \agent, along with other baseline language agents, on a diverse set of tasks that require multi-step reasoning and benefit from tool use.
About half of the tasks are used to train \agent modules based on their tool-integrated solution trajectories, while the others are held out and introduced at inference time. 
All tasks are evaluated in a zero-shot manner.

\textbf{Numerical reasoning tasks.} We include mathematics datasets with difficulties ranging from elementary school to high school competition-math problems, including GSM-8K~\cite{DBLP:journals/corr/abs-2110-14168}, MATH~\cite{DBLP:conf/nips/HendrycksBKABTS21}, and the Google DeepMind mathematics~\cite{DBLP:conf/iclr/SaxtonGHK19} and MathQA~\cite{DBLP:conf/naacl/AminiGLKCH19} tasks taken from the \lila benchmark~\cite{DBLP:conf/emnlp/MishraFLTWBRTSC22}.
For Google DeepMind mathematics, we measure the performance on the Algebra, Basic Math, Calculus, Mulplication/Division and Number Theory subsets, and the same for the Gain, General, Geometry, Physics and Probability subsets in MathQA.
We use GSM-8K and MATH for training and hold out the others for evaluation.
GSM-8K and MATH contribute 7.4K and 6.3K corrrect solution trajectories from their training sets, resulting in a total of 13.7K tool-integrated solution trajectories.

\textbf{Tabular reasoning tasks.} We use TabMWP, a tabular math-word problem dataset~\cite{DBLP:conf/iclr/Lu0CWZRCK23}, FinQA~\cite{DBLP:conf/emnlp/ChenCSSBLMBHRW21} and TAT-QA~\cite{DBLP:conf/acl/ZhuLHWZLFC20} which are both finance question-answering datasets, and a subset of test questions from MultimodalQA~\cite{DBLP:conf/iclr/TalmorYCLWAIHB21} which require multi-hop reasoning and understanding of both text and tabular modalities.
We use TabMWP and FinQA for both training and evaluation, and we hold out TAT-QA and MultimodalQA (MMQA) for evaluation.
We collect 2.6K correct solutions from TabMWP and 4.6K solutions from FinQA, resulting in a total of 7.2K tool-integrated solution trajectories.

\textbf{Knowledge-based reasoning tasks.} We incorporate HotpotQA~\cite{DBLP:conf/emnlp/Yang0ZBCSM18}, CWQ~\cite{DBLP:conf/naacl/TalmorB18} and Musique~\cite{DBLP:journals/tacl/TrivediBKS22} which contain complex queries with multi-step solutions, Bamboogle~\cite{DBLP:conf/emnlp/PressZMSSL23} which contain questions that can only be solve via multiple Google searches, and StrategyQA~\cite{DBLP:journals/tacl/GevaKSKRB21} which involves retrieving and reasoning over multiple facts.
We hold out Bamboogle and HotpotQA for evaluation, use CWQ and Musique only for training, and use StrategyQA for both training and evaluation.
We collect about 2.8K solutions each from CWQ and MusiQue and 1.3K solutions from StrategyQA, resulting in a total of 7K tool-integrated trajectories.

textbf{Mixed-tool reasoning tasks.} While the datasets above involve diverse multi-step reasoning strategies, they often can be solved with a single type of tool and risk contamination by proprietary models.
Therefore we add three custom tasks that better measure the model or agent's ability to invoke multiple tools.
These consist of our modified versions of DROP and IIRC, where the questions are decontextualized from the passages, as well as \dataset (Section~\ref{sec:husky-evaluation}).
We use DROP~\cite{DBLP:conf/naacl/DuaWDSS019} and IIRC~\cite{DBLP:conf/emnlp/FergusonGHKD20} for both training and evaluation, and collect 3K solution trajectories from each dataset for training.
For \dataset, we split the dataset by topic and assign 8 manually selected topics to the evaluation set and the rest to the training set, which yields a total of 1.3K tool-integrated trajectories.

Refer to Appendix~\ref{sec:husky-training-details} for a detailed breakdown of the statistics regarding the solution trajectories.

\subsection{Models}
\label{sec:husky-models}
\textbf{Action generator.}
We use \textsc{Llama-2-7B} and \textsc{13B}~\cite{DBLP:journals/corr/abs-2307-09288}, and \textsc{Llama-3-8B} for training the action generator $\mathcal{A}$.
Using the tasks listed in Section~\ref{sec:husky-datasets}, we remove the solution trajectories with incorrect answers and collect a total of 110K training instances for $\mathcal{A}$ across numerical, tabular, knowledge-based and mixed-tool reasoning tasks.
We fully finetune on this multi-task training set~\cite{DBLP:conf/iclr/SanhWRBSACSRDBX22}.

\textbf{Code generator.}
We choose \textsc{DeepSeekCoder-7B-Instruct-v1.5}~\cite{DBLP:journals/corr/abs-2401-14196} as the starting point for finetuning $\mathcal{M}_c$, due to its robust coding abilities at 7B parameters.
As described in Section~\ref{sec:husky-training}, we extract all code used in the solution trajectories that return the correct answers, and remove any code that fails to compile.
We collect a total of 44K code instances and fully finetune on this training set.

\textbf{Math reasoner.}
We select \textsc{DeepSeekMath-7B-Instruct}~\cite{DBLP:journals/corr/abs-2402-03300} as the model to finetune $\mathcal{M}_m$ from, due to its advanced mathematical reasoning capabilities.
In a similar manner to $\mathcal{M}_c$, we extract all math solutions in the tool-integrated trajectories with the correct answers.
This results in a total of 30K math solution instances, upon which we fully finetune $\mathcal{M}_m$.

\textbf{Query generator.}
As there is no model specialized for search query generation in particular, we use \textsc{Llama-2-7B}~\cite{DBLP:journals/corr/abs-2307-09288} as the base model for finetuning $\mathcal{M}_q$.
Again, we extract all search query instances in the tool-integrated trajectories resulting in the correct answers.
We fully finetune $\mathcal{M}_q$ on a total of 22K search query instances.

Refer to Appendix~\ref{sec:training-data-examples} for examples of the training data for each module.

\subsection{Baselines}
\label{sec:husky-baselines}
\textbf{Numerical, tabular and knowledge-based reasoning tasks.} 
We evaluate \agent along with other existing open-source language agents including \react~\cite{DBLP:conf/iclr/YaoZYDSN023}, \rewoo~\cite{DBLP:journals/corr/abs-2305-18323}, \chameleon~\cite{DBLP:conf/nips/LuPCGCWZG23}, \fireact~\cite{DBLP:journals/corr/abs-2310-05915} and \lumos~\cite{DBLP:journals/corr/abs-2311-05657} on 11 different tasks for numerical, tabular and knowledge-based reasoning.
Our baselines consist of two categories.
1) \textbf{few-shot prompted agents} (\react, \rewoo, \chameleon):
We construct few-shot prompts specific to each task category and generate the corresponding outputs using \textsc{Tulu-2-7B} during inference.
We generate thoughts and actions step-by-step for \react and plans in a single pass for \rewoo and \chameleon.
2) \textbf{finetuned agents} (\fireact, \lumos): we execute the model checkpoints according to their original prompts.
To keep comparisons fair, we map each tool call made by the agents to a corresponding \agent module and execute the module.
For example, we map a \textsc{calculator} call made by the agent to $\mathcal{M}_c$ to perform computation.
By fixating the toolset, we control for the confounding error associated with the tools employed by each agent and directly compare the agents' planning and tool use capabilities.

\textbf{Mixed-tool reasoning tasks.}
While language agents effectively address various multi-step reasoning tasks, many such tasks can be solved by state-of-the-art proprietary models with Chain-of-Thought (CoT)~\cite{DBLP:conf/nips/Wei0SBIXCLZ22} with better performance.
This is due to a combination of their superiority in core reasoning capabilities, long-context handling, and training data coverage with possibilities of contamination.
As such, we measure the performance of \texttt{gpt-3.5-turbo} and \texttt{gpt-4-turbo}\footnote{We use \texttt{gpt-3.5-turbo-0125}, \texttt{gpt-4-0125-preview}, \texttt{gpt-4-turbo-0409} and \texttt{gpt-4o}.} along with \react, \lumos and \agent on our set of mixed-tool reasoning tasks consisting of DROP*, IIRC* and \dataset.
We use the same inference procedure for the language agents as the other evaluation tasks, and we measure the performance of proprietary models using Chain-of-Thought.

\section{Results}
\label{sec:results}
We evaluate \agent and other baseline agent systems across 14 different evaluation tasks and present the results in Table \ref{tab:main-results}.
Overall, we find that \agent consistently outperforms all other baseline agents while using a jointly trained action generator and the same set of tools for all of the evaluation tasks.
We divide our analyses according to the main type of reasoning employed by each task, spanning numerical, tabular, knowledge-based and mixed reasoning.

\begin{table}[ht]
    \centering
     
    \begin{subtable}{1\textwidth}
        \centering
        \footnotesize
        \setlength{\tabcolsep}{3pt}
        \begin{tabular}{lcccccccc}
        \toprule
        Agent & \multicolumn{4}{c}{\scriptsize \textbf{Numerical reasoning}} & \multicolumn{4}{c}{\scriptsize \textbf{Tabular reasoning}} \\ 
         & \cellcolor{lightredalpha}GSM8K & \cellcolor{lightredalpha}MATH & \cellcolor{lightbluealpha}GDM Math. & \cellcolor{lightbluealpha}MathQA & \cellcolor{lightredalpha}TabMWP & \cellcolor{lightredalpha}FinQA & \cellcolor{lightbluealpha}TAT-QA & \cellcolor{lightbluealpha}MMQA \\
         & acc & acc & acc & acc & \scriptsize EM & \scriptsize EM & \scriptsize EM & \scriptsize EM / F1 \\
        \midrule
        CoT$_{\text{Tulu2-7B}}$ & 28.8 & 3.1 & 10.8 & 5.2 & 48.6 & 8.7 & 19.9 & 39.8/43.1    \\
        \react$_{\text{Tulu2-7B}}$ & 55.4 & 21.0 & 27.8 & 31.9 & 72 & 7.6 & 15.0 & 34.0/39.3    \\
        \rewoo$_{\text{Tulu2-7B}}$ & 52.2 & 28.7 & 53.7 & 25.4 & 67.0 & 14.7 & 38.2 & 32.6/37.5    \\
        \chameleon$_{\text{Tulu2-7B}}$ & 64.4 & 30.3 & 41.9 & 39.3 & 71.0 & 14.7 & 36.5 & 39.2/44.6    \\
        \fireact$_{\text{Llama2-7B}}$ & 56.1 & 19.7 & 29.6 & 22.6 & 29.2 & 8.2 & 12.7 & 44.2/49.6    \\
        \lumos$_{\text{Llama2-7B}}$ & 54.9 & 30.3 & 42.4 & 30.7 & 70.9 & 18.7 & 41.0 & 37.9/41.3    \\
        \midrule
        \agent$_{\text{Llama2-7B}}$ & 77.9 & 40.9 & \textbf{58.2} & 49.1 & \textbf{77.6} & 20.8 & 42.2 & \textbf{46.7}/\textbf{52.5}    \\
        \hdashline
        \agent$_{\text{Llama2-13B}}$ & 79.4 & 41.9 & 57.9 & \textbf{51.2} & 75.6 & 20.0 & \textbf{42.3} &     41.2/48.2    \\
        \agent$_{\text{Llama3-8B}}$ & \textbf{79.9} & \textbf{42.1} & 55.8 & 51.0 & 76.6 &  \textbf{20.9} & 41.8 & 43.7/50.5   \\
        \bottomrule
      \end{tabular}
        \caption{Numerical and tabular reasoning task results}
        \label{tab:results-numerical-tabular}
    \end{subtable}
    
    \vspace{0.2cm} 
    
    \begin{subtable}{1\textwidth}
        \centering
        \footnotesize
        \setlength{\tabcolsep}{0.8pt}
        \begin{minipage}{0.4\textwidth}
            \centering
            \begin{tabular}{lccc}
        \toprule
        Agent & \multicolumn{3}{c}{\scriptsize \textbf{Knowledge-based reasoning}}\\
        & \cellcolor{lightbluealpha}Bamboogle & \cellcolor{lightbluealpha}HotpotQA & \cellcolor{lightredalpha}Strat.QA \\
        & \scriptsize EM / F1 & \scriptsize EM / F1 & acc \\
        \midrule
        CoT$_{\text{Tulu2-7B}}$ & 28.0/37.7 & 20.5/29.0 & 68.0 \\
        \react$_{\text{Tulu2-7B}}$ & 37.6/48.7 & 24.4/34.0 & 62.0 \\
        \rewoo$_{\text{Tulu2-7B}}$ & 33.6/42.4 & 22.1/29.4 & 58.0 \\
        \chameleon$_{\text{Tulu2-7B}}$ & 39.2/50.8 & 24.8/34.7 & 60.7 \\
        \fireact$_{\text{Llama2-7B}}$ & 36.8/45.6 & 27.7/37.8 & 63.0 \\
        \lumos$_{\text{Llama2-7B}}$ & 52.0/64.5 & 26.8/36.3 & 66.3 \\
        \midrule
        \agent$_{\text{Llama2-7B}}$ & 54.4/65.8 & 32.7/46.6 & 70.0 \\
        \hdashline
        \agent$_{\text{Llama2-13B}}$ & 56.0/66.8 & \textbf{35.1}/48.8 & \textbf{71.3} \\
        \agent$_{\text{Llama3-8B}}$ & \textbf{58.4}/\textbf{70.2} & 33.7/\textbf{49.0} & 67.3 \\
        \bottomrule
      \end{tabular}
            \caption{Knowledge-based reasoning task results}
            \label{tab:results-knowledge-based}
        \end{minipage}
        \hspace{0.07\textwidth} 
        \begin{minipage}{0.48\textwidth}
            \centering
            \footnotesize
            \setlength{\tabcolsep}{0.8pt}
            \begin{tabular}{lccc}
        \toprule
        Agent/Model & \multicolumn{3}{c}{\scriptsize \textbf{Mixed-tool reasoning}}\\
        & \cellcolor{lightredalpha}DROP* & \cellcolor{lightredalpha}IIRC* & \dataset \\
        & \scriptsize EM / F1 & \scriptsize EM / F1 & acc \\
        \midrule
        \react$_{\text{Tulu2-7B}}$ & 15.0/19.8 & 25.5/31.1 & 7.19 \\
        \lumos$_{\text{Llama2-7B}}$ & 21.0/23.3 & 30.5/35.1 & 9.59 \\
        \scriptsize \texttt{gpt-3.5-turbo-0125} & 20.0/24.6 & 24.5/30.6 & 15.1 \\
        \scriptsize \texttt{gpt-4-0125-preview} & 22.0/25.5 & 28.0/33.7 & 20.2 \\
        \scriptsize \texttt{gpt-4-turbo-0409} & 25.0/29.5 & 27.5/32.9 & 25.3 \\
        \scriptsize \texttt{gpt-4o} & \textbf{28.0}/32.1 & 31.0/37.3 & \textbf{26.0} \\
        \midrule
        \agent$_{\text{Llama2-7B}}$ & 26.0/30.9 & 33.0/37.9 & 20.9 \\
        \hdashline
        \agent$_{\text{Llama2-13B}}$ & 27.5/\textbf{32.1} & 32.5/37.2 & 25.0 \\
        \agent$_{\text{Llama3-8B}}$ & 26.0/31.2 & \textbf{33.5}/\textbf{39.1} & 20.9 \\
        \bottomrule
      \end{tabular}
            \caption{Mixed-tool reasoning task results}
            \label{tab:results-mixed}
        \end{minipage}
    \end{subtable}
    
    \caption{
      Overall results across 14 different evaluation tasks. \agent outperforms or is on par with existing language agents while using an action generator jointly trained across different tasks on the numerical, tabular and knowledge-based reasoning tasks, and matches or outperforms state-of-the-art proprietary models on the mixed reasoning tasks.
      Tasks whose training sets are used for training \agent are indicated in \colorbox{lightredalpha}{\textbf{red}}, and unseen tasks introduced during evaluation are indicated in \colorbox{lightbluealpha}{\textbf{blue}}. 
}
    \label{tab:main-results}
\end{table}

\subsection{Results for Numerical, Tabular and Knowledge-based Reasoning}
\textbf{Numerical reasoning tasks.} The first four columns of Table \ref{tab:results-numerical-tabular} demonstrate the results for the numerical reasoning tasks, which include GSM-8K, MATH, Google-DeepMind Mathematics and MathQA.
While the training sets of GSM-8K and MATH are incorporated to create the tool-integrated trajectories, the Google-DeepMind Mathematics and MathQA datasets are unseen tasks introduced during evaluation.
\agent demonstrates significant performance improvements over other language agents by 10 to 20 points across all four tasks, indicating its ability to call and execute the correct tools required to solve each step of a given math question.
Notably, it shows consistency in its high performance even for unseen math tasks.
Of the baseline language agents, \chameleon delivers the highest and most consistent performance, corroborating a prior study indicating that agents tend to perform better in math tasks with single-pass planning rather than iterative planning~\cite{DBLP:journals/corr/abs-2311-05657}.

\agent's superior performance on numerical tasks can be attributed to its effective usage of $\mathcal{M}_m$ based on DeepSeekMath-7B-Instruct~\cite{DBLP:journals/corr/abs-2402-03300}.
As numerical reasoning tasks can be solved with CoT and bypass stepwise execution, we compare the performance of using a state-of-the-art CoT prompt and our iterative executions with $\mathcal{M}_m$ as shown in Table~\ref{tab:cot_vs_agent}.
For GSM-8K and MATH, we observe that CoT slightly edges over \agent. 
We hypothesize that DeepSeekMath-7B-Instruct is specialized in the two tasks by extensively incorporating their training data during instruction tuning in CoT format.
On the other hand, eight out of ten subtasks in Google-DeepMind mathematics and MathQA benefit from stepwise execution -- these results indicate that numerical reasoning tasks unseen during training can often be more accurately solved by iteratively addressing simpler subproblems.

\textbf{Tabular reasoning tasks.} The latter four columns of Table \ref{tab:results-numerical-tabular} display the results for the tabular reasoning tasks.
Note that TabMWP and FinQA are integrated for training \agent, while TAT-QA and MultimodalQA are newly introduced during evaluation.
Again, \agent consistently outperforms all other baseline agents across all four tasks due to its exposure to tabular data during training.
Some baseline agents demonstrate competitive performance on some benchmarks, such as \fireact for MMQA or \lumos for FinQA and TAT-QA, though the performance nevertheless slightly lags behind that of \agent.

\textbf{Knowledge-based reasoning tasks.} Table \ref{tab:results-knowledge-based} summarizes the evaluation results of the language agents on three different knowledge reasoning tasks.
The observed task for \agent includes StrategyQA, while the unseen tasks include Bamboogle and HotpotQA.
\agent outperforms other language agents across all evaluation tasks, even outperforming task-specific agents without incorporating the task in its own training data.
For example, \agent outperforms \fireact on HotpotQA by five EM points despite having never seen HotpotQA while \fireact has been trained on the dataset.
\agent achieves competitive scores across all knowledge-based tasks without using highly task-specific tools such as a Wikipedia passage retriever, capturing both precision and efficiency during its inference.
Refer to Appendix~\ref{sec:husky-inference-examples} for examples of solutions generated by \agent.

\begin{table}[]
    \centering
    \small
    \setlength{\tabcolsep}{2.2pt} 
    \renewcommand{\arraystretch}{1.0}
    \begin{tabular}{lcccccccccccc}
        \toprule
        Agent & \cellcolor{lightredalpha}GSM8K & \cellcolor{lightredalpha}MATH & \multicolumn{5}{c}{\cellcolor{lightbluealpha}GDM Mathematics} & \multicolumn{5}{c}{\cellcolor{lightbluealpha}MathQA} \\
        & & & alg & basic & calc & muldiv & numth & gain & gen & geom & phys & prob \\\midrule
        CoT$_{\text{DeepSeekMath-7B-Inst}}$ & \textbf{83.2} & 42.1 & 39.61 & 44.33 & 28.25 & 64.13 & 44.39 & 55.5 & 53.28 & \textbf{57.26} & \textbf{51.23} & 45.18\\
        \agent$_{\text{Llama2-7B}}$ & 77.86 & 40.94 & 43.66 & 62.0 & 44.61 & \textbf{86.77} & \textbf{53.87} & 56.78 & 51.74 & 46.15 & 45.9 & 44.85\\
        \agent$_{\text{Llama2-13B}}$ & 79.38 & 41.92 & \textbf{46.22} & \textbf{68.33} & 44.24 & 78.03 & 52.42 & 57.29 & \textbf{55.56} & 52.51 & 51.02 & 39.53 \\
        \agent$_{\text{Llama3-8B}}$ & 79.91 & \textbf{42.12} & 43.66 & 58.0 & \textbf{46.47} & 83.86 & 47.2 & \textbf{58.57} & 53.8 & 49.57 & 47.34 & \textbf{45.85}\\
        \bottomrule
    \end{tabular}
    \vspace{0.3cm}
    \caption{Effects of stepwise execution on numerical reasoning tasks compared to Chain-of-Thought (CoT). While using the optimized CoT prompt with DeepSeekMath-7B-Instruct even outperforms agent executions on GSM-8K and MATH, the agent executions outperform CoT in many cases where the evaluation task has not been observed in the model's instruction tuning stage.}
    \label{tab:cot_vs_agent}
\end{table}

\subsection{Results for Mixed-Tool Reasoning}
Table \ref{tab:results-mixed} demonstrates  evaluation results on the mixed reasoning tasks, which consist of DROP*, IIRC* and \dataset.
Despite only using open-source 7B LMs for its modules, \agent outperforms all other language agents and proprietary models except for \texttt{gpt-4o} on DROP*, and outperforms all baselines for IIRC*.
Similarly for \dataset, \agent outperforms other language agents by a significant margin.
It also outperforms \texttt{gpt-3.5-turbo} and on par with \texttt{gpt-4-0125-preview} with a 7B action generator and on par with \texttt{gpt-4-turbo-0409} with a 13B action generator.

These results showcase \agent's ability to iteratively solve complex questions via precise decomposition of the task into simpler subtasks with appropriate tool choices.
While \agent has integrated a subset of \dataset into its training, this split is done by different real-world topics, indicating that \agent learns to generalize the multi-tool reasoning strategy across various domains such as geography, sports, chemistry and more.
Our experiment results show that it is possible to develop open-source agents that compete with the best proprietary models for solving complex questions through a careful fine-tuning procedure that endows advanced planning and multi-tool use abilities.
Refer to Appendix~\ref{sec:husky-inference-examples} for examples of solutions generated by \agent.

\section{Analysis}
\label{sec:analysis}
Having showcased \agent's multi-step reasoning capabilities across 14 different tasks, we further examine design choices that affect \agent's performance.
Our analysis is summarized as follows.
\textbf{(1) Cross-domain generalization}: we investigate whether the action generator responsible for decomposing the problem into simpler steps benefits from domain-specific or cross-domain training.
\textbf{(2) Tool choice}: we study the effect of tool capabilities on \agent's downstream performance.
\textbf{(3) \agent$_{\text{Llama3-8B-all}}$}: we share a variant of \agent that is built upon the same base model (Llama-3-8B) for the action generator and the specialized modules $\mathcal{M}_c$, $\mathcal{M}_m$ and $\mathcal{M}_q$.

\subsection{Cross-task Generalization}
One notable feature of \agent is its use of an action generator that is jointly trained across different tasks involving numerical, tabular and knowledge-based reasoning to predict the next step to take.
In constrast, many existing language agents are either finetuned on tasks within a specific domain (e.g., \fireact), or provide different modules each trained on a specific domain (e.g., \lumos).
We explore this design choice in a more systematic way by training variants of \agent's action generator separately on numerical, tabular and knowledge reasoning tasks and comparing their performance on in-domain tasks to the jointly trained action generator.

\begin{table}[]
    \centering
    \setlength{\tabcolsep}{5.0pt} 
    \renewcommand{\arraystretch}{1.0}
    \begin{tabular}{lcccccc}
    \toprule
        Action generator & \multicolumn{2}{c}{Numerical (Acc)} & \multicolumn{2}{c}{Tabular (EM)} & \multicolumn{2}{c}{Knowledge (EM)}\\
         & GDM Math. & MathQA & TAT-QA & MMQA & Bamboogle & HotpotQA \\\midrule
        Domain-Specific & 58.07 & 48.82 & \textbf{44.85} & 40.06 & \textbf{55.2} & \textbf{34.1}\\
        Joint (110K) & \textbf{58.18} & \textbf{49.08} & 42.2 & \textbf{46.69} & 54.4 & 32.7 \\
    \bottomrule
    \end{tabular}
    \vspace{0.3cm}
    \caption{Comparison between \agent w/ action generator trained on domain-specific tasks, and \agent w/ action generator jointly trained on all tasks in the original training set.}
    \label{tab:action-ablations}
\end{table}

Table~\ref{tab:action-ablations} shows the results of our analyses.
For numerical reasoning tasks, we observe that the cross-domain action generator score equally to the domain-specific action generator.
On the other hand, our results on Bamboogle and HotpotQA indicate that knowledge-based reasoning tasks slightly benefit from domain-specific training.
The results for tabular tasks are mixed, with MultimodalQA greatly benefitting from joint training of $\mathcal{A}$ while TAT-QA slightly decreases in performance.

These results hint that the benefit of domain-specific or cross-domain training for the action generator depends on the domain of the downstream task being evaluated upon.
However, the difference in the metrics is small enough to conclude that joint training across different task domains mostly preserve the action generator's performance in each domain, indicating a positive signal towards scaling \agent with an even larger and more diverse array of tasks.

\begin{table}[h]
  \centering
  \caption{Ablations for the code generator ($\mathcal{M}_c$) and the math reasoner ($\mathcal{M}_m$). For each tool, we use a set of five tasks that display high frequencies of invoking the tool, change the tools with different models and run \agent. We use a subset of 1K examples for MATH to reduce the inference time.}
  \label{tab:tool-ablations}
  \footnotesize
  \begin{tabular}{c}
      \begingroup
      \setlength{\tabcolsep}{5.0pt}
        \subcaptionbox{Results for code generator ($\mathcal{M}_c$)}{
            \begin{tabular}{lccccc}
                \toprule
                Tool & MATH (1K) & GDM Math. & TabMWP & FinQA & TAT-QA \\
                \midrule
                CodeTulu-7B & 25.6 & 54.54 & 42.6 & 6.88 & 40.53\\
                Llama-3-8B & 39.4 & 57.05 & 74.5 & 20.74 & 41.23\\
                DeepSeekCoder-7B-Instruct-v1.5 & \textbf{41.8} & \textbf{58.18} & \textbf{77.6} & \textbf{20.83} & \textbf{42.2}\\
                \bottomrule
            \end{tabular}
        }
      \endgroup
    \end{tabular}
    \begingroup
    \setlength{\tabcolsep}{6.0pt}
    \subcaptionbox{Results for math reasoner ($\mathcal{M}_m$)}{
        \begin{tabular}{lccccc}
            \toprule
            Tool & GSM8K & MATH (1K) & GDM Math. & MathQA & HuskyQA\\
            \midrule
            Tulu-2-7B & 49.28 & 23.5 & 49.05 & 17.63 & 14.04\\
            Llama-3-8B & 67.7 & 33.3 & 52.59 & 31.3 & 18.15\\
            DeepSeekMath-7B-Instruct & \textbf{77.86} & \textbf{41.8} & \textbf{58.18} & \textbf{49.08} & \textbf{20.89}\\
            \bottomrule
        \end{tabular}
    }
    \endgroup
\end{table}

\subsection{Tool Choice}
Another key component of \agent is the tools or modules used for executing each step predicted by the action generator.
We conduct our ablations by switching out the models used for the code generator $\mathcal{M}_c$ and the math reasoner $\mathcal{M}_m$ while keeping everything else in \agent fixed.
We ablate $\mathcal{M}_c$ with CodeTulu-7B and Llama-3-8B in addition to DeepSeekCoder-7B-Instruct-v1.5, and we ablate $\mathcal{M}_m$ with Tulu-2-7B and Llama-3-8B in addition to DeepSeekMath-7B-Instruct.

Table~\ref{tab:tool-ablations} shows the results of our ablations.
For code generators, we confirm that DeepSeekCoder-7B-Instruct-v1.5 provides the best initialization for $\mathcal{M}_c$ with its state-of-the-art coding capabilities.
Llama-3-8B also demonstrates strong coding capabilities that brings the agent's performance very close to the best system's performance across all ablation tasks.
For math reasoners, we also find that DeepSeekMath-7B-Instruct results in the best performance for the agent.
We observe a significant performance gap with Tulu-2-7B due to its relative lack of math reasoning capabilities.
Llama-3-8B closes this gap about halfway through according to our evaluation results, but it still lacks in comparison to DeepSeekMath which is pre-trained and fine-tuned specifically for math tasks.

\subsection{\agent$_{\text{Llama3-8B-all}}$}
While \agent benefits from careful decisions about which models to use for tools, in some cases it is difficult to know in advance which model serves as the best candidate for a new tool functionality requested by a user.
Therefore we build a variant of \agent where the action generator and the tools $\mathcal{M}_c$, $\mathcal{M}_m$ and $\mathcal{M}_q$ are all trained from Llama-3-8B, a state-of-the-art LM in the 7-8B size family.

\begin{table}
  \centering
  \caption{
  Results for \agent$_{\text{Llama3-8B-all}}$, which shares Llama-3-8B as the base model for $\mathcal{A}$, $\mathcal{M}_c$, $\mathcal{M}_m$ and $\mathcal{M}_q$. The performance remains competitive despite switching out the code and math expert LMs into a general-purpose base LM for fine-tuning the tool  modules.
  }
  \vspace{0.2cm}
  \label{tab:fixed-model}
  \footnotesize
  \begin{tabular}{c}
    \begingroup
    \setlength{\tabcolsep}{2.5pt} 
    \renewcommand{\arraystretch}{1.0}
    {
      \begin{tabular}{lcccccccc}
        \toprule
        \multirow{3}{*}{Agent} & \multicolumn{4}{c}{\scriptsize \textbf{Numerical reasoning}} & \multicolumn{4}{c}{\scriptsize \textbf{Tabular reasoning}} \\ 
         & GSM8K & MATH & GDM Math. & MathQA & TabMWP & FinQA & TAT-QA & MMQA \\
         & acc & acc & acc & acc & \scriptsize EM & \scriptsize EM & \scriptsize EM & \scriptsize EM / F1 \\
        \midrule
        \agent$_{\text{Llama3-8B}}$ & \textbf{79.91} & \textbf{42.12} & \textbf{55.84} & \textbf{51.03} & \textbf{76.6} &  20.92 & 41.78 & \textbf{43.65}/\textbf{50.45}   \\
        \agent$_{\text{Llama3-8B-all}}$ & 70.58 & 30.0 & 46.03 & 31.27 & 74.1 & \textbf{23.21} & \textbf{42.9} & 42.82/49.91   \\
        \bottomrule
      \end{tabular}
      }
    \endgroup
    \\
    \\
    \begingroup
    \setlength{\tabcolsep}{5.0pt} 
    \renewcommand{\arraystretch}{1.0}
    {
      \begin{tabular}{lcccccc}
        \toprule
        \multirow{3}{*}{Agent} & \multicolumn{3}{c}{\scriptsize \textbf{Knowledge reasoning}} & \multicolumn{3}{c}{\scriptsize \textbf{Mixed-tool reasoning}} \\
        & Bamboogle & HotpotQA & StrategyQA & DROP* & IIRC* & \dataset \\
        & \scriptsize EM / F1 & \scriptsize EM/F1 & acc & \scriptsize EM / F1 & \scriptsize EM / F1 & acc \\
        \midrule
        \agent$_{\text{Llama3-8B}}$ & \textbf{58.4}/\textbf{70.2} & \textbf{33.7}/\textbf{48.98} & 67.33 & \textbf{26.0}/\textbf{31.24} & 33.5/39.12 & \textbf{20.89} \\
        \agent$_{\text{Llama3-8B-all}}$ & 56.0/68.94 & 33.5/48.14 & \textbf{68.67} & 25.0/29.47 & \textbf{34.5}/\textbf{40.25} & 17.12       \\
        \bottomrule
      \end{tabular}
      }
    \endgroup
  \end{tabular}
\end{table}

Our results are summarized in Table~\ref{tab:fixed-model}.
\agent$_{\text{Llama3-8B-all}}$ demonstrates similar performance to that of \agent$_{\text{Llama3-8B}}$ on most downstream tasks, except for numerical reasoning tasks due to its lack of a math-specialized model such as DeepSeekMath-Instruct for $\mathcal{M}_m$.

These results demonstrate that fine-tuning all \agent modules using a fixed base LM with strong reasoning capabilities yields robust performances across a wide array of tasks, bypassing the need to identify and download specialized models to fine-tune each module from.
We expect \agent's performances to further improve with scaling in terms of the parameter count, as well as integrating new open LMs with stronger reasoning capabilities as they continue to be released.

\section{Conclusion}
\label{sec:conclusion}
We introduce \agent, a unified, open-source language agent that jointly solves multi-step reasoning tasks which require numerical, tabular, knowledge-based, and mixed-tool reasoning.
First, we train an action generator that iteratively predicts the high-level steps and the associated tools for solving tasks across different domains.
We also finetune strong base LMs with high quality training data to integrate highly performant expert models into \agent.
Moreover, we develop \dataset and other evaluation sets that stress-test language agents' abilities to address mixed-tool, multi-step reasoning tasks, and show existing gaps in frontier models such as GPT-4 for such tasks.
We perform additional experiments to provide scientific insights on the effects of model choice and dataset composition on \agent's performance.
Here, we discover that further scaling the action space and expert models will allow \agent to address an even wider range of tasks.
Our work presents a robust recipe for building open-source language agents that generalize across different types of multi-step reasoning tasks.

\bibliographystyle{plainnat}
\bibliography{neurips_2024}
\clearpage
\appendix

\section{Dataset Details}
We utilize the following list of datasets in this work, across both training and evaluation.
\begin{itemize}[left=0pt, itemsep=0.5em]
    \item \textbf{GSM-8K}~\cite{DBLP:journals/corr/abs-2110-14168}: We use the publicly available train split as the part of the training set for generating tool-integrated solution trajectories, and the test split as part of the evaluation set.
    \item \textbf{MATH}~\cite{DBLP:conf/nips/HendrycksBKABTS21}: We use the publicly available train split as the part of the training set for generating tool-integrated solution trajectories, and the test split as part of the evaluation set.
    \item \textbf{Google DeepMind Mathematics}~\cite{DBLP:conf/iclr/SaxtonGHK19}: We use the entire train and test split as part of the evaluation set. We use the algebra, basic math, calculus, multiplication-division and number theory subsets. We do not use the dataset for training.
    \item \textbf{MathQA}~\cite{DBLP:conf/naacl/AminiGLKCH19}: We use the publicly available test split as part of the evaluation set. We use the gain, general, geometry, physics and probability subsets. We do not use the dataset for training.
    \item \textbf{TabMWP}~\cite{DBLP:conf/iclr/Lu0CWZRCK23}: We use the publicly available train split as part of the training set or generating tool-integrated solution trajectories, and the 1K version of the test split as part of the evaluation set.
    \item \textbf{FinQA}~\cite{DBLP:conf/emnlp/ChenCSSBLMBHRW21}: We use the publicly available train split as part of the training set or generating tool-integrated solution trajectories, and the test split as part of the evaluation set.
    \item \textbf{TAT-QA}~\cite{DBLP:conf/acl/ZhuLHWZLFC20}: We use the publicly available test split as part of the evaluation set. We do not use the dataset for training.
    \item \textbf{MultimodalQA}~\cite{DBLP:conf/iclr/TalmorYCLWAIHB21}: We use part of the publicly available development split as part of the evaluation set. We filter for compositional questions that use both text and table modalities. We do not use the dataset for training.
    \item \textbf{ComplexWebQuestions}~\cite{DBLP:conf/naacl/TalmorB18}: We use part of the publicly available train split as the part of the training set for generating tool-integrated solution trajectories. We only use the dataset for training.
    \item \textbf{MusiQue}~\cite{DBLP:journals/tacl/TrivediBKS22}: We use part of the publicly available train split as the part of the training set for generating tool-integrated solution trajectories. We only use the dataset for training.
    \item \textbf{Bamboogle}~\cite{DBLP:conf/emnlp/PressZMSSL23}: We use the test split as part of the evaluation set. We do not use the dataset for training.
    \item \textbf{HotpotQA}~\cite{DBLP:conf/emnlp/Yang0ZBCSM18}: We use part of the test split as part of the evaluation set. We use 1,000 examples for the test split. We do not use the dataset for training.
    \item \textbf{StrategyQA}~\cite{DBLP:journals/tacl/GevaKSKRB21}: We use the publicly available train split as the part of the training set for generating tool-integrated solution trajectories, and part of the test split as part of the evaluation set.
    \item \textbf{DROP}~\cite{DBLP:conf/naacl/DuaWDSS019}: We use part of the publicly available train split as the part of the training set for generating tool-integrated solution trajectories, and a subset of the test split for evaluation. We decontextualize all questions to be understood independently from their associated passages in the original dataset.
    \item \textbf{IIRC}~\cite{DBLP:conf/emnlp/FergusonGHKD20}: We use part of the publicly available train split as the part of the training set for generating tool-integrated solution trajectories, and a subset of the test split for evaluation. We decontextualize all questions to be understood independently from their associated passages in the original dataset.
    \item \textbf{\dataset (ours)}: We use our full training split only for finetuning the action generator only. We use human-verified examples in the test split as part of the evaluation set.
\end{itemize}

\section{Tool Details}
\subsection{Code}
We use a Python interpreter with Python 3.9, and prepend the following set of import statements which enumerate a default list of libraries for \agent to use.
\begin{lstlisting}
import math
import numpy as np
import sympy
from datetime import datetime
from math import comb, gcd, lcm
from scipy.optimize import minimize
from sympy import symbols, Eq, solve, expand, factor, simplify, Matrix
from sympy.solvers.inequalities import solve_univariate_inequality
from sympy.core.relational import LessThan
\end{lstlisting}
We post-process the execution results by rounding all float outputs up to four digits after the decimal.

\subsection{Search}
\label{sec:search-info}
We use the Python library for SERP API (\texttt{v0.1.5}) to obtain Google Search results for search queries.
We first search for the answer box and extract the snippet within the box if it exists.
Otherwise, we take the topmost organic search result and extract the title and the snippet from the search result.
\begin{figure}[!h]
    \centering
    \includegraphics[scale=0.35, clip, trim=0.5cm 1.0cm 0.5cm 1.0cm]{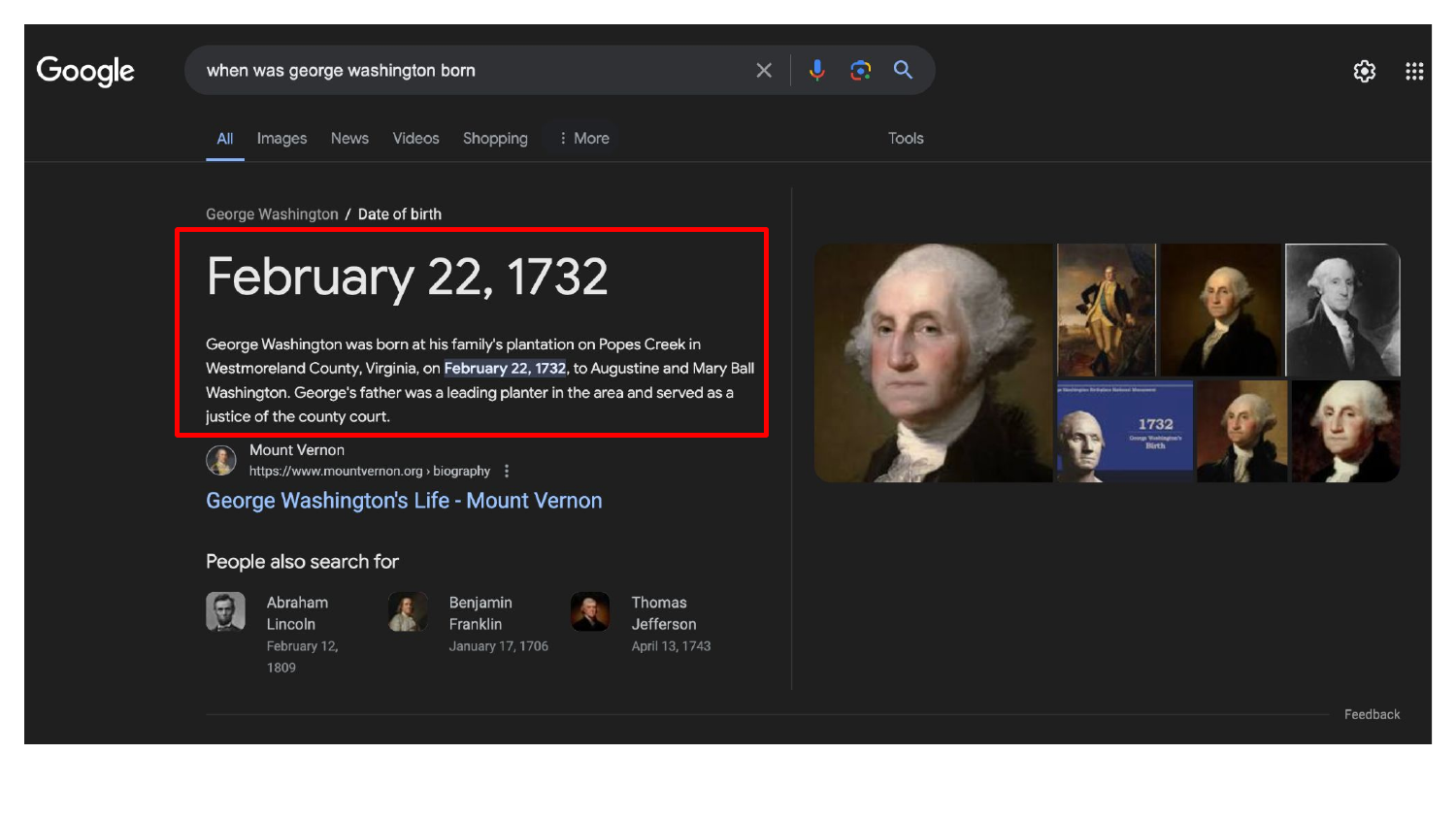}
    \caption{Visualization of a Google Search result (equally returned by SERP API) for the search query "when was george washington born". We use the information presented in the red box.}
    \label{fig:google-search-result}
\end{figure}

\section{Training Details}
\label{sec:husky-training-details}
\subsection{Dataset Composition for Tool-Integrated Solution Trajectories} 
Table~\ref{tab:training-data-composition} summarizes the dataset composition of the solution trajectories collected for training various modules in \agent, as well as the composition for training the action generator $\mathcal{A}$.
We use a subset of the training sets for tasks that require knowledge-based reasoning, including CWQ, MusiQue, StrategyQA, DROP and IIRC, due to rate limits imposed by SERP API.

\begin{table}[]
    \centering
    \setlength{\tabcolsep}{5.0pt} 
    \renewcommand{\arraystretch}{1.0}
    \begin{tabular}{lccc}
    \toprule
        \textbf{Dataset} & \textbf{\# Solutions} & \textbf{\# Correct Solutions} & \textbf{\# Actions} \\\midrule
        GSM-8K & 7500 & 7436 & 22595 \\
        MATH & 7475 & 6338 & 21368 \\
        TabMWP & 3000 & 2550 & 6839 \\
        FinQA & 6203 & 4568 & 14314 \\
        CWQ & 6000 & 2840 & 7359 \\
        MusiQue & 6000 & 2777 & 9198 \\
        StrategyQA & 1603 & 1279 & 4743 \\
        DROP & 6330 & 2932 & 9347 \\
        IIRC & 6000 & 3067 & 9265 \\
        \dataset & 1350 & 1350 & 6302 \\\midrule
        Total & 51,461 & 35,137 & 111,330 \\
    \bottomrule
    \end{tabular}
    \vspace{0.3cm}
    \caption{Number of solution trajectories, successful trajectories and action examples collected from each training dataset spanning numerical, tabular, knowledge-based and mixed reasoning.}
    \label{tab:training-data-composition}
\end{table}
\subsection{Training Hyperparameters}
For all training runs, we use the DeepSpeed ZeRO-3 Optimizer~\cite{DBLP:conf/sc/RajbhandariRRH20} in BF16 precision over 3 epochs with a learning rate of 5e-6, weight decay 1e-2, a linear scheduler for the learning rate, max length of 2048, warmup ratio of 0.03, and a total batch size of 32.

\textbf{Action generator.}
We use the provided hyperparameter configuration and fine-tune Llama-2-7B and 13B across 2 NVIDIA A100 80GB GPUs, and Llama-3-8B across 4 NVIDIA A100 80GB GPUs.

\textbf{Code generator.}
We use the provided hyperparameter configuration and fine-tune CodeTulu-7B and DeepSeekCoder-7B-Instruct-v1.5 across 2 NVIDIA A100 80GB GPUs, and Llama-3-8B across 4 NVIDIA A100 80GB GPUs.

\textbf{Math reasoner.}
We use the provided hyperparameter configuration and fine-tune Tulu-2-7B and DeepSeekMath-7B-Instruct across 2 NVIDIA A100 80GB GPUs, and Llama-3-8B across 4 NVIDIA A100 80GB GPUs.

\textbf{Query generator.}
We use the provided hyperparameter configuration and fine-tune Llama-2-7B across 2 NVIDIA A100 80GB GPUs, and Llama-3-8B across 4 NVIDIA A100 80GB GPUs.
\section{Evaluation Details}
\subsection{Inference}
Algorithm~\ref{alg:husky-inference} presents a detailed overview of the inference procedure.
We run all modules with batch size 16, temperature 0 (0.3 for the non-finetuned $\mathcal{M}_r$), and a strict maximum of 10 iterations for solving each problem.
In practice, we take a distinctively efficient approach to executing our agent modules by running all modules in parallel and use vLLM~\cite{DBLP:conf/sosp/KwonLZ0ZY0ZS23} to perform efficient inference.
\vspace{-0.5em}
\begin{algorithm}
\caption{\agent Inference}
\label{alg:husky-inference}
\begin{algorithmic}[1]
    \REQUIRE input $x$, action generator $\mathcal{A}$, modules $\mathcal{M}_c$, $\mathcal{M}_m$, $\mathcal{M}_q$, $\mathcal{M}_r$
    \STATE step index $i=0$, solution history $h_0 = \emptyset$
    \WHILE {$\mathcal{A}(x, h_i) \neq \text{\textbf{[finish]}} \text{ AND } i < 10$}
        \STATE $s_i, t_i = \mathcal{A}(x, h_i)$
        \IF {$t_i = \textbf{[code]}$}
            \STATE $c_i = \mathcal{M}_c(x, h_i, s_i)$
            \STATE $e_i = \text{PYTHON}(c_i)$
            \STATE $o_i = \mathcal{M}_r(x, s_i, c_i, e_i)$
        \ELSIF {$t_i = \textbf{[math]}$}
            \STATE $o_i = \mathcal{M}_m(x, h_i, s_i)$
        \ELSIF {$t_i = \textbf{[search]}$}
            \STATE $q_i = \mathcal{M}_q(x, h_i, s_i)$
            \STATE $e_i = \text{GOOGLE}(q_i)$
            \STATE $o_i = \mathcal{M}_r(s_i, e_i)$
        \ELSIF {$t_i = \textbf{[commonsense]}$}
            \STATE $o_i = \mathcal{M}_r(x, h_i, s_i)$
        \ELSIF {$t_i = \textbf{[finish]}$}
            \STATE $a = \text{EXTRACT}(s_i)$
        \ENDIF
        \STATE $i \mathrel{+}= 1$, $h_i =$ "$h_{i-1}\texttt{\textbackslash n}s_i\texttt{\textbackslash n}o_i$"
    \ENDWHILE
    \RETURN $a$
\end{algorithmic}
\end{algorithm}
\vspace{-0.5em}
\subsection{Evaluation Datasets and Metrics}
We share additional details into how the datasets are evaluated. All random seeds are initialized at 42.
\begin{itemize}[left=0pt, itemsep=0.2em]
    \item \textbf{GSM-8K}, \textbf{MATH}, \textbf{Bamboogle}, \textbf{FinQA}, \textbf{TAT-QA}: We use the full test split containing 1,319 examples, 5,000 examples, 125 examples and 1,133 examples, respectively.
    \item \textbf{TabMWP}: We use the 1K version of the test split, which contains 1,000 examples.
    \item \textbf{StrategyQA}: We use a subset of 300 randomly selected examples from the test split, following \cite{DBLP:journals/corr/abs-2311-05657}. We use the same subset for all our evaluations.
    \item \textbf{HotpotQA}: We use a subset of 1,000 randomly selected examples from the test split, following \cite{DBLP:journals/corr/abs-2311-05657}. We use the same subset for all our evaluations.
    \item \textbf{Google-DeepMind Mathematics}: We combine both the train and test splits from the algebra (1,679), basic math (300), calculus (269), muldiv (446) and number theory (1,034) subsets.
    \item \textbf{MathQA}: We use the full test split for the gain (391), general (777), geometry (117) and physics (488) subsets, and we combine both the train and test splits for the probability (301) subset due to its lack of test examples.
    \item \textbf{MultimodalQA}: We use 362 examples in the development set that are labeled as compositional questions using both text and table modalities.
    \item \textbf{DROP*}: We use 200 examples from the test split as part of the evaluation set, and we decontextualize the questions to remove the solution's dependency on the input passage.
    \item \textbf{IIRC*}: We use 200 examples from the test split as part of the evaluation set, and we decontextualize the questions to remove the solution's dependency on the input passage.
    \item \textbf{\dataset}: We use 292 examples from the test split that are evaluated by humans to be well-formed questions with the labeled answers matching human outputs.
\end{itemize}

\subsection{GPT-4 performance}
Table~\ref{tab:gpt-4-performance} lists the GPT-4 performances across all evaluation tasks for reference.
We report our replicated results with \texttt{gpt-4-0125-preview} for all evaluation tasks except GSM-8K and MATH, for which we use the reported numbers from DeepSeekMath~\cite{DBLP:journals/corr/abs-2402-03300} to save API costs.

\begin{table}
  \centering
  \caption{
  GPT-4 results across our evaluation tasks, compared to \agent.
  }
  \vspace{0.2cm}
  \label{tab:gpt-4-performance}
  \footnotesize
  \begin{tabular}{c}
    \begingroup
    \setlength{\tabcolsep}{2.5pt} 
    \renewcommand{\arraystretch}{1.0}
    {
      \begin{tabular}{lcccccccc}
        \toprule
        \multirow{3}{*}{Agent} & \multicolumn{4}{c}{\scriptsize \textbf{Numerical reasoning}} & \multicolumn{4}{c}{\scriptsize \textbf{Tabular reasoning}} \\ 
         & \cellcolor{lightredalpha}GSM8K & \cellcolor{lightredalpha}MATH & \cellcolor{lightbluealpha}GDM Math. & \cellcolor{lightbluealpha}MathQA & \cellcolor{lightredalpha}TabMWP & \cellcolor{lightredalpha}FinQA & \cellcolor{lightbluealpha}TAT-QA & \cellcolor{lightbluealpha}MMQA \\
         & acc & acc & acc & acc & \scriptsize EM & \scriptsize EM & \scriptsize EM & \scriptsize EM / F1 \\
        \midrule
        \agent$_{\text{Llama2-7B}}$ & 77.9 & 40.9 & \textbf{58.2} & 49.1 & 77.6 & 20.8 & 42.2 & 46.7/52.5 \\
        \agent$_{\text{Llama2-13B}}$ & 79.4 & 41.9 & 57.9 & \textbf{51.2} & 75.6 & 20.0 & 42.3 &     41.2/48.2 \\
        \agent$_{\text{Llama3-8B}}$ & 79.9 & 42.1 & 55.8 & 51.0 & 76.6 &  20.9 & 41.8 & 43.7/50.5   \\\midrule
        GPT-4 & \textbf{92.0} & \textbf{52.9} & 47.5 & 43.4 & \textbf{96.2} & \textbf{33.0} & \textbf{73.1} & \textbf{66.0}/\textbf{72.5}   \\
        \bottomrule
      \end{tabular}
      }
    \endgroup
    \\
    \\
    \begingroup
    \setlength{\tabcolsep}{5.4pt} 
    \renewcommand{\arraystretch}{1.0}
    {
      \begin{tabular}{lcccccc}
        \toprule
        \multirow{3}{*}{Agent} & \multicolumn{3}{c}{\scriptsize \textbf{Knowledge reasoning}} & \multicolumn{3}{c}{\scriptsize \textbf{Mixed-tool reasoning}} \\
        & \cellcolor{lightbluealpha}Bamboogle & \cellcolor{lightbluealpha}HotpotQA & \cellcolor{lightredalpha}Strat.QA & \cellcolor{lightredalpha}DROP* & \cellcolor{lightredalpha}IIRC* & \dataset \\
        & \scriptsize EM / F1 & \scriptsize EM/F1 & acc & \scriptsize EM / F1 & \scriptsize EM / F1 & acc \\
        \midrule
        \agent$_{\text{Llama2-7B}}$ & 54.4/65.8 & 32.7/46.6 & 70.0 & 26.0/30.9 & 33.0/37.9 & 20.9 \\
        \agent$_{\text{Llama2-13B}}$ & 56.0/66.8 & 35.1/48.8 & \textbf{71.3} & \textbf{27.5}/\textbf{32.1} & 32.5/37.2 & \textbf{25.0} \\
        \agent$_{\text{Llama3-8B}}$ & \textbf{58.4}/\textbf{70.2} & 33.7/49.0 & 67.3 & 26.0/31.2 & \textbf{33.5}/\textbf{39.1} & 20.9 \\\midrule
        GPT-4 & 52.8/68.41 & \textbf{39.4}/\textbf{53.1} & 71.0 & 22.0/25.5 & 28.0/33.73 & 20.2 \\
        \bottomrule
      \end{tabular}
      }
    \endgroup
  \end{tabular}
\end{table}

\section{Tool-Integrated Solution Trajectories}
\label{sec:tool-integrated-solution-trajectories}
\subsection{Inference}
We use \texttt{gpt-4-0125-preview} to generate the tool-integrated solution trajectories for all datasets as summarized in Table~\ref{tab:training-data-composition}.
Our inference consists of two prompts: the first prompt generates the solution trajectory (refer to Section~\ref{sec:solution-trajectory-examples}) until a \textbf{[code]} or \textbf{[search]} tool call, and the second prompt re-writes the tool-executed output into natural language.
We generate all solution trajectories with max length 4096, max new tokens 1024 and temperature 0.3.
We generate all tool-executed output rewrites with max length 2048, max new tokens 256 and temperature 0.3.
We devise a separate solution trajectory generation prompt for each dataset (or groups of datasets), as well as a separate tool-execution writer prompt for numerical, tabular and knowledge-based reasoning tasks, respectively.
\subsection{Instructions}
We provide instructions for generating the tool-integrated solution trajectories for each category in Figures~\ref{fig:trajectory-instruction-numerical},~\ref{fig:trajectory-instruction-tabular},~\ref{fig:trajectory-instruction-knowledge} and ~\ref{fig:trajectory-instruction-mixed}.
Refer to Section~\ref{sec:solution-trajectory-prompts} for the complete set of prompts with the few-shot demonstrations.

\begin{figure}[h!]

    \centering
\tcbset{colback=verylightgray, colframe=verylightgray, boxrule=0.5pt, arc=4pt}
\begin{tcolorbox}
\lstset{
    basicstyle=\ttfamily\color{black}\tiny,
    keywordstyle=\color{black},
    identifierstyle=\color{black},
    commentstyle=\color{black},
    stringstyle=\color{black},
    breaklines=true,
    backgroundcolor=\color{verylightgray},
    frame=none,
    numbers=none
}
\begin{lstlisting}
Given a math problem, integrate step-by-step reasoning and Python code to obtain the solution. Make sure to generate 1) the next step to be taken, 2) the tool to be used (either [code] or [math]), and 3) the associated code or math solution for that step.
- The format should be "Step N:" for the Nth step. The first sentence in each step should be a high-level summary of what to do for that step. It should begin with an imperative verb such as "calculate", "compute", "determine", "find" or "identify", and end with a newline character.
- If this sentence involves writing equations, define the variables in this sentence as well. For example, instead of "Write equations for the problem.", write "Write equations for the problem, using $x$ to represent the number of apples and $y$ to represent the number of oranges.
- After the step, choose between [code] and [math] to either return a code-based solution for the step, or a math-based solution for the step.
- For the code-based solution, write the code between lines occupied by the expressions '```python' and '```'.
- For the code-based solution, do not attempt to predict the executed output of the code - just write the code.
- For the code-based solution, do not use variables that are undefined within the same code snippet. If the variable is mentioned in an earlier code snippet, then copy the value of the variable over to the current code snippet.
- For the math-based solution, present the final result for the step in LaTeX using '\\boxed{}' without any units.
- If the final answer ANS has been reached in the output of the previous step, simply return "The answer is: <answer>ANS</answer>.", with the final answer between the tags <answer> and </answer>. Do not write the unit between the tags.
- Utilize the 'pi' symbol and 'Rational' from sympy for $\pi$ and fractions, and simplify all fractions and square roots without converting them to decimal values.
- Do not generate more than one code snippet at a time.
- Example imports are provided below. Import any of these packages in the code snippet, as well as additional packages as needed.
import math
import numpy as np
import sympy
from math import comb, gcd, lcm
from scipy.optimize import minimize
from sympy import symbols, Eq, solve, expand, factor, Matrix
from sympy.solvers.inequalities import solve_univariate_inequality
from sympy.core.relational import LessThan
Below are a few examples of the generated output. Adhere to the format shown in the examples.
\end{lstlisting}
\end{tcolorbox}
\caption{Instruction for numerical reasoning tasks}
\label{fig:trajectory-instruction-numerical}
\end{figure}

\begin{figure}[h!]
    \centering
\tcbset{colback=verylightgray, colframe=verylightgray, boxrule=0.5pt, arc=4pt}
\begin{tcolorbox}
\lstset{
    basicstyle=\ttfamily\color{black}\tiny,
    keywordstyle=\color{black},
    identifierstyle=\color{black},
    commentstyle=\color{black},
    stringstyle=\color{black},
    breaklines=true,
    backgroundcolor=\color{verylightgray},
    frame=none,
    numbers=none
}
\begin{lstlisting}
Given a question input that contains some passage, a table and the actual question, integrate step-by-step reasoning and Python code to obtain the solution. Make sure to generate 1) the next step to be taken, 2) the tool to be used (either [code] or [commonsense]), and 3) the associated code or text solution for that step.
- The format should be "Step N:" for the Nth step. The first sentence in each step should be a high-level summary of what to do for that step. It should begin with an imperative verb such as "identify", "find", "calculate", "compute" or "determine", and end with a newline character.
- After the step, choose between [code] and [commonsense] to either return a code-based solution or a commonsense-based solution for the step.
- Use the [code] tool to convert stem-and-leaf plots into lists, or to perform precise computations with large numbers or decimals.
- Use the [commonsense] tool for all other tasks, such as identifying values from the table or performing commonsense reasoning such as comparing two numbers.
- For the code-based solution, write the code between lines occupied by the expressions '```python' and '```'.
- For the code-based solution, do not attempt to predict the executed output of the code - just write the code.
- Use the Fraction class in the fractions library for code-based solutions if the question asks to return a fraction.
- For the code-based solution, do not use variables that are undefined within the same code snippet. If the variable is mentioned in an earlier code snippet, then copy the value of the variable over to the current code snippet.
- If the final answer ANS has been reached in the output of the previous step, simply return "The answer is: <answer>ANS</answer>.", with the final answer between the tags <answer> and </answer>.
Below are a few examples of the generated output. Adhere to the format shown in the examples.
\end{lstlisting}
\end{tcolorbox}
\caption{Instruction for tabular reasoning tasks}
\label{fig:trajectory-instruction-tabular}
\end{figure}

\begin{figure}[h!]
    \centering
\tcbset{colback=verylightgray, colframe=verylightgray, boxrule=0.5pt, arc=4pt}
\begin{tcolorbox}
\lstset{
    basicstyle=\ttfamily\color{black}\tiny,
    keywordstyle=\color{black},
    identifierstyle=\color{black},
    commentstyle=\color{black},
    stringstyle=\color{black},
    breaklines=true,
    backgroundcolor=\color{verylightgray},
    frame=none,
    numbers=none
}
\begin{lstlisting}
Given a factoid question, integrate step-by-step reasoning and search queries to obtain the solution. Make sure to generate 1) the next step to be taken, 2) the tool to be used ([search], [code] or [commonsense]), and 3) the associated search query or text solution for that step.
- The format should be "Step N:" for the Nth step. The first sentence in each step should be a high-level summary of what to do for that step. It should begin with an imperative verb such as "search", "recall", "determine", "find" or "identify", and end with a newline character.
- After the step, choose between [search], [code] and [commonsense] to either return a search-based solution, a code-based solution, or a commonsense-based solution for the step.
- Use the search tool to obtain answers for questions that cannot be answered by your own knowledge.
- For the search-based solution, write the search query between lines occupied by the expressions '```google' and '```'.
- For the commonsense-based solution, present the final result for the step between the tags '<output>' and '</output>'.
- If the search result does not solve the current step, then retry that step with an improved query.
- If the final answer ANS has been reached in the output of the previous step, simply return "The answer is: <answer>ANS</answer>.", with the final answer between the tags <answer> and </answer>.
- Do not generate more than one search query at a time.
Below are a few examples of the generated output. Adhere to the format shown in the examples.
\end{lstlisting}
\end{tcolorbox}
\caption{Instruction for knowledge-based reasoning tasks}
\label{fig:trajectory-instruction-knowledge}
\end{figure}

\begin{figure}[h!]
    \centering
\tcbset{colback=verylightgray, colframe=verylightgray, boxrule=0.5pt, arc=4pt}
\begin{tcolorbox}
\lstset{
    basicstyle=\ttfamily\color{black}\tiny,
    keywordstyle=\color{black},
    identifierstyle=\color{black},
    commentstyle=\color{black},
    stringstyle=\color{black},
    breaklines=true,
    backgroundcolor=\color{verylightgray},
    frame=none,
    numbers=none
}
\begin{lstlisting}
Given an input question, integrate step-by-step reasoning with math problem-solving, Python code and search queries to obtain the solution. Make sure to generate 1) the next step to be taken, 2) the tool to be used ([math], [code], [search] or [commonsense]), and 3) the associated search math, code, search or commonsense-based solution for that step.
- The format should be "Step N:" for the Nth step. The first sentence in each step should be a high-level summary of what to do for that step. It should begin with an imperative verb such as "search", "find", "calculate", "compute", "determine" or "identify", and end with a newline character.
- After the step, choose between [math], [search], [code] and [commonsense] to return a math-based solution, a search-based solution, a code-based solution, or a commonsense reasoning-based solution.
- Use the [math] tool to perform mathematical reasoning, given that all the necessary information is provided in the question and the solution history so far.
- Use the [search] tool to obtain information that is not provided in the question and cannot be answered by your own knowledge.
- For the search-based solution, write the search query between lines occupied by the expressions '```google' and '```'.
- Use the [code] tool to perform precise numerical computations with large numbers or decimals, to count long lists of numbers or perform operations between two different dates.
- For the code-based solution, write the code between lines occupied by the expressions '```python' and '```'.
- For the code-based solution, do not attempt to predict the executed output of the code - just write the code.
- For the code-based solution, do not use variables that are undefined within the same code snippet. If the variable is mentioned in an earlier code snippet, then copy the value of the variable over to the current code snippet.
- Do not generate more than one code snippet at a time.
- Use the [commonsense] tool for all other common reasoning tasks (e.g., comparing two numbers) that do not require the usage of math/search/code.
- If the step contains a short answer in its output, present the short answer to the step between the tags '<output>' and '</output>'.
- If the final answer "ANS" has been reached in the output of the previous step, simply return "The answer is: <answer>ANS</answer>.", with the final answer between the tags <answer> and </answer>.
- Example imports are provided below. Import any of these packages in the code snippet, as well as additional packages as needed.
import math
import numpy as np
import sympy
from math import comb, gcd, lcm
from scipy.optimize import minimize
from sympy import symbols, Eq, solve, expand, factor, Matrix
from sympy.solvers.inequalities import solve_univariate_inequality
from sympy.core.relational import LessThan
Below are a few examples of the generated output. Adhere to the format shown in the examples.
\end{lstlisting}
\end{tcolorbox}
\caption{Instruction for mixed reasoning tasks}
\label{fig:trajectory-instruction-mixed}
\end{figure}

\clearpage
\subsection{Examples}
\label{sec:solution-trajectory-examples}

\begin{figure}[h!]
    \centering
\tcbset{colback=verylightgray, colframe=verylightgray, boxrule=0.5pt, arc=4pt}
\begin{tcolorbox}
\lstset{
    basicstyle=\ttfamily\color{black}\tiny,
    keywordstyle=\color{black},
    identifierstyle=\color{black},
    commentstyle=\color{black},
    stringstyle=\color{black},
    breaklines=true,
    backgroundcolor=\color{verylightgray},
    frame=none,
    numbers=none,
    escapechar=@
}
\begin{lstlisting}
@\textbf{Question}@: A line passes through $(2,2,1)$ and $(5,1,-2).$  A point on this line has an $x$-coordinate of 4. Find the $z$-coordinate of the point.

@\textbf{Solution}@:
@\textbf{Step 1}@: Calculate the direction vector of the line.
Tool: [code]
```python
import numpy as np

# define the coordinates of the two points
point1 = np.array([2, 2, 1])
point2 = np.array([5, 1, -2])

# calculate the direction vector
direction_vector = point2 - point1

# print the output
out = direction_vector
print(out)
```
```output
[ 3 -1 -3]
```
The direction vector of the line is [3, -1, -3].

@\textbf{Step 2}@: Find the parameter $t$ when the $x$-coordinate is 4.
Tool: [code]
```python
from sympy import symbols, Eq, solve

# define the symbol for the parameter
t = symbols('t')

# define the initial x-coordinate and the direction vector's x-component
x_initial = 2
direction_x = 3

# equation for the x-coordinate at point t
eq = Eq(x_initial + direction_x*t, 4)

# solve for t
t_value = solve(eq, t)[0]

# print the output
out = t_value
print(out)
```
```output
2/3
```
The parameter $t$ when the $x$-coordinate is 4 is $2/3.

@\textbf{Step 3}@: Compute the $z$-coordinate of the point when $t = 2/3$.
Tool: [code]
```python
from sympy import Rational

# define the initial z-coordinate and the direction vector's z-component
z_initial = 1
direction_z = -3

# compute the z-coordinate at point t
z_coordinate = z_initial + direction_z * Rational(2, 3)

# print the output
out = z_coordinate
print(out)
```
```output
-1.0
```
The $z$-coordinate of the point on the line when $t = 2/3$ is -1.0.

The answer is: @\textbf{<answer>-1.0</answer>}@.
\end{lstlisting}
\end{tcolorbox}
\caption{Example tool-integrated solution trajectory from MATH}
\label{fig:trajectory-example-MATH}
\end{figure}

\begin{figure}[h!]
    \centering
\tcbset{colback=verylightgray, colframe=verylightgray, boxrule=0.5pt, arc=4pt}
\begin{tcolorbox}
\lstset{
    basicstyle=\ttfamily\color{black}\tiny,
    keywordstyle=\color{black},
    identifierstyle=\color{black},
    commentstyle=\color{black},
    stringstyle=\color{black},
    breaklines=true,
    backgroundcolor=\color{verylightgray},
    frame=none,
    numbers=none,
    escapechar=@
}
\begin{lstlisting}
@\textbf{Question}@: table of contents the following discussion of nonoperating income and expense excludes the results of the merger in order to provide a more meaningful year-over-year comparison . interest expense , net of capitalized interest decreased $ 249 million in 2014 from 2013 primarily due to a $ 149 million decrease in special charges recognized year-over-year as further described below , as well as refinancing activities that resulted in $ 100 million less interest expense recognized in 2014 . ( 1 ) in 2014 , we recognized $ 33 million of special charges relating to non-cash interest accretion on bankruptcy settlement obligations . in 2013 , we recognized $ 138 million of special charges relating to post-petition interest expense on unsecured obligations pursuant to the plan and penalty interest related to american 2019s 10.5% ( 10.5 % ) secured notes and 7.50% ( 7.50 % ) senior secured notes . in addition , in 2013 we recorded special charges of $ 44 million for debt extinguishment costs incurred as a result of the repayment of certain aircraft secured indebtedness , including cash interest charges and non-cash write offs of unamortized debt issuance costs . ( 2 ) as a result of the 2013 refinancing activities and the early extinguishment of american 2019s 7.50% ( 7.50 % ) senior secured notes in 2014 , we recognized $ 100 million less interest expense in 2014 as compared to 2013 . other nonoperating expense , net in 2014 consisted of $ 114 million of net foreign currency losses , including a $ 43 million special charge for venezuelan foreign currency losses , and $ 56 million in other nonoperating special charges primarily due to early debt extinguishment costs related to the prepayment of our 7.50% ( 7.50 % ) senior secured notes and other indebtedness . the foreign currency losses were driven primarily by the strengthening of the u.s . dollar relative to other currencies during 2014 , principally in the latin american market , including a 48% ( 48 % ) decrease in the value of the venezuelan bolivar and a 14% ( 14 % ) decrease in the value of the brazilian real . other nonoperating expense , net in 2013 consisted principally of net foreign currency losses of $ 56 million and early debt extinguishment charges of $ 29 million . reorganization items , net reorganization items refer to revenues , expenses ( including professional fees ) , realized gains and losses and provisions for losses that are realized or incurred as a direct result of the chapter 11 cases . the following table summarizes the components included in reorganization items , net on aag 2019s consolidated statement of operations for the year ended december 31 , 2013 ( in millions ) : . ( 1 ) in exchange for employees 2019 contributions to the successful reorganization , including agreeing to reductions in pay and benefits , we agreed in the plan to provide each employee group a deemed claim , which was used to provide a distribution of a portion of the equity of the reorganized entity to those employees . each employee group received a deemed claim amount based upon a portion of the value of cost savings provided by that group through reductions to pay and benefits as well as through certain work rule changes . the total value of this deemed claim was approximately $ 1.7 billion . ( 2 ) amounts include allowed claims ( claims approved by the bankruptcy court ) and estimated allowed claims relating to ( i ) the rejection or modification of financings related to aircraft and ( ii ) entry of orders treated as unsecured claims with respect to facility agreements supporting certain issuances of special facility revenue bonds . the debtors recorded an estimated claim associated with the rejection or modification of a financing .
['', '2013']
['labor-related deemed claim ( 1 )', '$ 1733']
['aircraft and facility financing renegotiations and rejections ( 2 ) ( 3 )', '325']
['fair value of conversion discount ( 4 )', '218']
['professional fees', '199']
['other', '180']
['total reorganization items net', '$ 2655']
what is the percent of the labor-related deemed claim as part of the total reorganization items net in 2013
@\textbf{Solution}@:
@\textbf{Step 1}@: Identify the total value of labor-related deemed claim in 2013.
Tool: [commonsense]
According to the table, the labor-related deemed claim in 2013 is $1733 million.

@\textbf{Step 2}@: Identify the total reorganization items net in 2013.
Tool: [commonsense]
According to the table, the total reorganization items net in 2013 is $2655 million.

@\textbf{Step 3}@: Calculate the percentage of the labor-related deemed claim as part of the total reorganization items net in 2013.
Tool: [code]
```python
labor_related_claim = 1733
total_reorganization_items = 2655
percentage = (labor_related_claim / total_reorganization_items) * 100
# print the output
print(percentage)
```
```output
65.2731
```
The labor-related deemed claim constitutes 65.2731 percent of the total reorganization items net in 2013.

The answer is: @\textbf{<answer>65.2731\%</answer>}@.
\end{lstlisting}
\end{tcolorbox}
\caption{Example tool-integrated solution trajectory from FinQA}
\label{fig:trajectory-example-finqa}
\end{figure}

\begin{figure}[h!]
    \centering
\tcbset{colback=verylightgray, colframe=verylightgray, boxrule=0.5pt, arc=4pt}
\begin{tcolorbox}
\lstset{
    basicstyle=\ttfamily\color{black}\tiny,
    keywordstyle=\color{black},
    identifierstyle=\color{black},
    commentstyle=\color{black},
    stringstyle=\color{black},
    breaklines=true,
    backgroundcolor=\color{verylightgray},
    frame=none,
    numbers=none,
    escapechar=@
}
\begin{lstlisting}
@\textbf{Question}@: When did the organization that gives out the Frank P. Brown Medal open?
@\textbf{Solution}@:
@\textbf{Step 1}@: Identify the organization that gives out the Frank P. Brown Medal.
Tool: [search]
```google
organization that gives out the Frank P. Brown Medal
```
```output
The Frank P. Brown Medal was formerly awarded by the Franklin Institute for excellence in science, engineering, and structures. It was established by the 1938 will of Franklin Pierce Brown, a member of the Master Plumbers Association.
```
The organization that gives out the Frank P. Brown Medal is <output>the Franklin Institute</output>.

@\textbf{Step 2}@: Find the year the Franklin Institute was founded.
Tool: [search]
```google
year the Franklin Institute was founded
```
```output
On February 5, 1824, Samuel Vaughan Merrick and William H. Keating founded The Franklin Institute of the State of Pennsylvania for the Promotion of the Mechanic Arts.
```
The Franklin Institute was founded in <output>1824</output>.

The answer is: @\textbf{<answer>1824</answer>}@.
\end{lstlisting}
\end{tcolorbox}
\caption{Example tool-integrated solution trajectory from MusiQue}
\label{fig:trajectory-example-strategyqa}
\end{figure}

\begin{figure}[h!]
    \centering
\tcbset{colback=verylightgray, colframe=verylightgray, boxrule=0.5pt, arc=4pt}
\begin{tcolorbox}
\lstset{
    basicstyle=\ttfamily\color{black}\tiny,
    keywordstyle=\color{black},
    identifierstyle=\color{black},
    commentstyle=\color{black},
    stringstyle=\color{black},
    breaklines=true,
    backgroundcolor=\color{verylightgray},
    frame=none,
    numbers=none,
    escapechar=@
}
\begin{lstlisting}
@\textbf{Question}@: Would Kurt Cobain have benefited from Project Semicolon?
@\textbf{Solution}@:
@\textbf{Step 1}@: Understand what Project Semicolon is.
Tool: [search]
```google
what is Project Semicolon?
```
```output
About Project Semicolon Organization: Project Semicolon is a compassionate online community that centers around mental health awareness and suicide prevention. Our platform serves as a safe ...
```
Project Semicolon is an organization focused on mental health awareness and suicide prevention.

@\textbf{Step 2}@: Determine the main issues Kurt Cobain faced.
Tool: [search]
```google
what issues did Kurt Cobain face?
```
```output
He was also prone to alcoholism, suffered from depression, and regularly used drugs and inhalants. Cobain had two uncles who killed themselves using guns. On March 4, 1994, Cobain was hospitalized in Rome following an overdose of Rohypnol (Flunitrazepam) and alcohol.
```
The main issues Kurt Cobain faced were alcoholism, depression, and regular use of drugs and inhalants.

@\textbf{Step 3}@: Determine if Project Semicolon's focus aligns with the issues Kurt Cobain faced.
Tool: [commonsense]
Project Semicolon focuses on mental health awareness and suicide prevention. Kurt Cobain faced issues such as alcoholism, depression, and regular use of drugs and inhalants. Given the alignment between Cobain's issues and Project Semicolon's focus, it is plausible that Kurt Cobain could have benefited from Project Semicolon.

The answer is: @\textbf{<answer>yes</answer>}@.
\end{lstlisting}
\end{tcolorbox}
\caption{Example tool-integrated solution trajectory from StrategyQA}
\label{fig:trajectory-example-strategyqa}
\end{figure}

\begin{figure}[h!]
    \centering
\tcbset{colback=verylightgray, colframe=verylightgray, boxrule=0.5pt, arc=4pt}
\begin{tcolorbox}
\lstset{
    basicstyle=\ttfamily\color{black}\tiny,
    keywordstyle=\color{black},
    identifierstyle=\color{black},
    commentstyle=\color{black},
    stringstyle=\color{black},
    breaklines=true,
    backgroundcolor=\color{verylightgray},
    frame=none,
    numbers=none,
    escapechar=@
}
\begin{lstlisting}
@\textbf{Question}@: How many yards longer was Marcedes Lewis's longest touchdown compared to his shortest in the game between the Dallas Cowboys and the Jaguars where David Buehler hit a 34-yard field goal?
@\textbf{Solution}@:
@\textbf{Step 1}@: Identify the date of the game between the Dallas Cowboys and the Jaguars where David Buehler hit a 34-yard field goal.
Tool: [search]
```google
precise date of the game between the Dallas Cowboys and the Jaguars where David Buehler hit a 34-yard field goal
```
```output
Jaguars 35-17 Cowboys (1 Nov, 2010) Final Score: Dallas Cowboys. FG. 10:51. David Buehler 34 Yard Field Goal is Good. 9 plays, 39 yards, 4:14. 0 ; Jacksonville Jaguars. TD. 7:55. Mike Sims-Walker 10 Yard Pass ...
```
The game between the Dallas Cowboys and the Jaguars, where David Buehler hit a 34-yard field goal, took place on <output>November 1, 2010</output>.

@\textbf{Step 2}@: Retrieve information about the game.
Tool: [search]
```google
game between the Dallas Cowboys and the Jaguars on November 1, 2010
```
```output
['Quarter', 'Time', 'Tm', 'Detail', 'JAX', 'DAL']
['1','10:46','Cowboys','David Buehler 34 yard field goal','0','3']
['1','7:50','Jaguars','Mike Sims-Walker 10 yard pass from David Garrard (Josh Scobee kick)','7','3']
['2','9:43','Jaguars','Marcedes Lewis 42 yard pass from David Garrard (Josh Scobee kick)','14','3']
['3','12:14','Jaguars','Mike Thomas 15 yard pass from David Garrard (Josh Scobee kick)','21','3']
['3','9:01','Jaguars','Marcedes Lewis 9 yard pass from David Garrard (Josh Scobee kick)','28','3']
['4','12:29','Cowboys','Marion Barber III 1 yard rush (David Buehler kick)','28','10']
['4','5:50','Jaguars','David Garrard 2 yard rush (Josh Scobee kick)','35','10']
['4','2:29','Cowboys','Jason Witten 8 yard pass from Jon Kitna (David Buehler kick)','35','17']
```

@\textbf{Step 3}@: Calculate the difference in yards between Marcedes Lewis's longest and shortest touchdowns in the game.
Tool: [code]
```python
longest_touchdown = 42
shortest_touchdown = 9
yard_difference = longest_touchdown - shortest_touchdown
print(yard_difference)
```
```output
33.0
```
Marcedes Lewis's longest touchdown was 33 yards longer than his shortest touchdown in the game between the Dallas Cowboys and the Jaguars.

The answer is: @\textbf{<answer>33</answer>}@ yards.
\end{lstlisting}
\end{tcolorbox}
\caption{Example tool-integrated solution trajectory from DROP*}
\label{fig:question-rewrite-drop-sports-fewshot-1}
\end{figure}

\clearpage
\section{DROP* and IIRC*}
\label{sec:drop-iirc-details}
\subsection{Question re-writing}
We decontextualize a subset of questions from DROP and IIRC using \texttt{gpt-3.5-turbo-0125} with max length 2048, max new tokens 128 and temperature 0.3.
We provide few-shot prompts for decontextualizing both DROP and IIRC below.

\begin{figure}[h!]
    \centering
\tcbset{colback=verylightgray, colframe=verylightgray, boxrule=0.5pt, arc=4pt}
\begin{tcolorbox}
\lstset{
    basicstyle=\ttfamily\color{black}\tiny,
    keywordstyle=\color{black},
    identifierstyle=\color{black},
    commentstyle=\color{black},
    stringstyle=\color{black},
    breaklines=true,
    backgroundcolor=\color{verylightgray},
    frame=none,
    numbers=none,
    escapechar=@
}
\begin{lstlisting}
Given a passage that describes a sports game, rewrite the old question by integrating specific information from the passage such that the user can answer the new question without reading the passage and using Google Search instead.
- Indicate the two teams that played in the given game, using the phrase "the game between A and B" for teams A and B.
- Choose one piece of information from the passage and add it to the question to better specify the game between the two teams.
- This information should be related to statistical details of the game related to a single player regarding 1) scoring a field goal or 2) a TD run.
- This information should NOT be related to other aspects of the game such as passing, injuries or other celebratory events.
- Make sure that the new information added does not directly answer the question itself.
- Make sure that the new question is grammatical.
Below are a few examples of the generated output. Adhere to the format shown in the examples.
---
Passage: This game involved a scary moment, after Seattle's  Ricardo Lockette was hit during a kick return. He lied on the ground, motionless, for about 7 minutes before he was taken off the field on a cart. X-rays later revealed that Lockette had a broken neck. The injury ended his career. The Cowboys would only kick field goals in this game, as Dan Bailey was 4 for 4 on field goals. Dallas lead 12-10 with under 2 minutes to go. However, the Seahawks would march down the field and would take a 13-12 lead after Steven Hauschka drilled a 24-yard field goal. Dallas tried to come back, but Seattle forced a turnover on downs to end the game.
Old question: How many total points were scored in the game?
New question: How many total points were scored in the game between the Dallas Cowboys and the Seattle Seahawks where Steven Hauschka scored a 24-yard field goal?
---
Passage: Hoping to rebound from their loss to the Patriots, the Raiders stayed at home for a Week 16 duel with the Houston Texans.  Oakland would get the early lead in the first quarter as quarterback JaMarcus Russell completed a 20-yard touchdown pass to rookie wide receiver Chaz Schilens.  The Texans would respond with fullback Vonta Leach getting a 1-yard touchdown run, yet the Raiders would answer with kicker Sebastian Janikowski getting a 33-yard and a 30-yard field goal.  Houston would tie the game in the second quarter with kicker Kris Brown getting a 53-yard and a 24-yard field goal. Oakland would take the lead in the third quarter with wide receiver Johnnie Lee Higgins catching a 29-yard touchdown pass from Russell, followed up by an 80-yard punt return for a touchdown.  The Texans tried to rally in the fourth quarter as Brown nailed a 40-yard field goal, yet the Raiders' defense would shut down any possible attempt.
Old question: Who scored the first touchdown of the game?
New question: Who scored the first touchdown of the game between the Oakland Raiders and the Houston Texans where Vonta Leach got a 1-yard touchdown run in the first quarter?
---
Passage: Still searching for their first win, the Bengals flew to Texas Stadium for a Week 5 interconference duel with the Dallas Cowboys.  In the first quarter, Cincinnati trailed early as Cowboys kicker Nick Folk got a 30-yard field goal, along with RB Felix Jones getting a 33-yard TD run.  In the second quarter, Dallas increased its lead as QB Tony Romo completed a 4-yard TD pass to TE Jason Witten.  The Bengals would end the half with kicker Shayne Graham getting a 41-yard and a 31-yard field goal. In the third quarter, Cincinnati tried to rally as QB Carson Palmer completed an 18-yard TD pass to WR T. J. Houshmandzadeh.  In the fourth quarter, the Bengals got closer as Graham got a 40-yard field goal, yet the Cowboys answered with Romo completing a 57-yard TD pass to WR Terrell Owens.  Cincinnati tried to come back as Palmer completed a 10-yard TD pass to Houshmandzadeh (with a failed 2-point conversion), but Dallas pulled away with Romo completing a 15-yard TD pass to WR Patrick Crayton.
Old question: Which team scored the final TD of the game?
New question: Which team scored the final TD of the game between the Cincinnati Bengals and the Dallas Cowboys where Nick Folk scored a 30-yard field goal in the first quarter?
---
Passage: Trying to snap a six-game losing skid, the Lions returned home for an NFC North rematch the-now 2-time NFC North champion Chicago Bears.  In the first quarter, the Bears struck first with kicker Robbie Gould nailing a 36-yard field goal.  Afterwards, the Lions took the lead with QB Jon Kitna completing a 23-yard TD pass to TE Dan Campbell.  In the second quarter, Chicago bounced back with QB Rex Grossman completing a 13-yard TD pass to WR Bernard Berrian.  Afterwards, RB Adrian Peterson got a 2-yard TD run.  In the third quarter, Detroit retook the lead with Kitna completing a 20-yard TD pass to WR Mike Furrey and a 2-yard TD pass to WR Roy Williams.  However, in the fourth quarter, the inconsistency that continues to plague the Lions showed as the Bears won with Gould getting a 36-yard field goal, a 39-yard field goal, and a 44-yard field goal and on a dropped pass by Mike Williams in the endzone on the last play of the game.  With their seventh-straight loss, the Lions fell to 2-13 as they were swept by their division rivals.
Old question: How many field goals did Robbie Gould make in the 4th quarter?
New question: How many field goals did Robbie Gould make in the 4th quarter in the game between the Detroit Lions and the Chicago Bears where he nailed a 36-yard field goal in the first quarter?
\end{lstlisting}
\end{tcolorbox}
\caption{Few-shot prompt for decontextualizing DROP (sports)}
\label{fig:question-rewrite-drop-sports-fewshot-2}
\end{figure}

\begin{figure}[h!]
    \centering
\tcbset{colback=verylightgray, colframe=verylightgray, boxrule=0.5pt, arc=4pt}
\begin{tcolorbox}
\lstset{
    basicstyle=\ttfamily\color{black}\tiny,
    keywordstyle=\color{black},
    identifierstyle=\color{black},
    commentstyle=\color{black},
    stringstyle=\color{black},
    breaklines=true,
    backgroundcolor=\color{verylightgray},
    frame=none,
    numbers=none,
    escapechar=@
}
\begin{lstlisting}
Given a passage that describes a historical event, rewrite the old question by integrating specific information from the passage such that the events in the question are less ambiguous and the user can answer the new question without reading the passage and using Google Search instead.
- Replace ambiguous entities (e.g., "the city") with the specific entity (e.g., "New York City") from the passage.
- For historical events which happened multiple times (e.g., Battle of Hastings), add more details (e.g., Battle of Hastings in the 11th century) to narrow it down to a single event.
- Do NOT generate questions that include the years for both of the historical events. If necessary, add the year to the event that is more ambiguous without the year description.
- Make sure that any new information added does not directly answer the question itself.
- Make sure that the new question asks for the same information as the old question in the passage.
- Leave the question as it is if it is already specific enough to solve on its own without the passage.
- Make sure that the new question is grammatical.
Below are a few examples of the generated output. Adhere to the format shown in the examples.
---
Passage: In 1085, Guadalajara was retaken by the Christian forces of Alfonso VI . The chronicles say that the Christian army was led by Alvar Fanez de Minaya, one of the lieutenants  of El Cid. From 1085 until the Battle of Las Navas de Tolosa in 1212, the city suffered wars against the Almoravid and the Almohad Empires. In spite of the wars, the Christian population could definitely settle down in the area thanks to the repopulation with people from the North  who received their first fuero in 1133 from Alfonso VII.In 1219, the king Fernando III gave a new fuero to the city .During the reign of Alfonso X of Castile, the protection of the king allowed the city to develop its economy by protecting merchants and allowing markets.
Old question: How many years after the people of the North received their first fuero from Alfonso VII did king Fernando III give a new fuero to the city?
New question: How many years after the people of the North in Guadalajara received their first fuero from Alfonso VII did king Fernando III give a new fuero to Guadalajara?
---
Passage: The Lithuanian Civil War of 1432-1438 was a conflict over the succession to the throne of the Grand Duchy of Lithuania, after Vytautas the Great died in 1430 without leaving an heir.  The war was fought on the one side by \u0160vitrigaila, allied with the Teutonic Knights, and on the other by Sigismund K\u0119stutaitis, backed by the Kingdom of Poland. The war threatened to sever the Union of Krewo, the personal union between Poland and Lithuania. \u0160vitrigaila's alliance with the Grand Master of the Teutonic Order, Paul von Rusdorf, launched the Polish-Teutonic War  but failed to secure victory for \u0160vitrigaila. When Sigismund captured power in Lithuania by staging a coup in 1432, Lithuania split into two opposing camps, and there began three years of devastating hostilities. To prevent the Knights from continuing their support of \u0160vitrigaila, Poland backed a Hussite invasion of Prussia in 1433. The war ended in a decisive defeat for \u0160vitrigaila and his ally, the Livonian branch of the Teutonic Knights, at the Battle of Pabaiskas in September 1435. \u0160vitrigaila eventually surrendered in 1437; Sigismund K\u0119stutaitis ruled Lithuania for only eight years before he was assassinated in 1440.
Old question: How many years did the Lithuanian Civil War last?
New question: How many years did the Lithuanian Civil War that started in 1432 last?
---
Passage: Before Hunyadi could assemble his forces, the army of Mehmed II arrived at Belgrade. The siege began on July 4, 1456. Szil\u00e1gyi could rely on a force of only 5,000-7,000 men in the castle. Mehmed set up his siege on the neck of the headland and started heavily bombarding the city's walls on June 29. He arrayed his men in three sections: The Rumelian corps had the majority of his 300 cannons, while his fleet of 200 river war vessels had the rest of them. The Rumelians were arrayed on the right wing and the Anatolian corps were arrayed on the left. In the middle were the personal guards of the Sultan, the Janissaries, and his command post. The Anatolian corps and the Janissaries were both heavy infantry troops. Mehmed posted his river vessels mainly to the northwest of the city to patrol the marshes and ensure that the fortress was not reinforced. They also kept an eye on the Sava river to the southwest to avoid the infantry from being outflanked by Hunyadi's army. The zone from the Danube eastwards was guarded by the Sipahi, the Sultan's feudal heavy cavalry corps, to avoid being outflanked on the right.
Old question: Which event happened first, the siege, or Hunyadi assembling his forces?
New question: Which event happened first, the siege of Belgrade by the army of Mehmed II or the assembly of Hunyadi's forces to aid Belgrade?
---
Passage: Since the end of World War II, in part due to industrial size and the onset of the Cold War, the United States has often been a proponent of reduced tariff-barriers and free trade. The U.S. helped establish the General Agreement on Tariffs and Trade  and later the World Trade Organization ; although it had rejected an earlier version in the 1950s . Since the 1970s, U.S. governments have negotiated managed-trade agreements, such as the North American Free Trade Agreement  in the 1990s, the Dominican Republic-Central America Free Trade Agreement  in 2006, and a number of bilateral agreements . In Europe, six countries formed the European Coal and Steel Community  in 1951 which became the European Economic Community  in 1958. Two core objectives of the EEC were the development of a common market, subsequently renamed the single market, and establishing a customs union between its member states. After expanding its membership, the EEC became the European Union  in 1993. The European Union, now the world's largest single market, has concluded free trade agreements with many countries around the world.
Old question: What Trade agreement happened first, North American Free Trade Agreement or Dominican Republic-Central America Free Trade Agreement?
New question: What Trade agreement happened first, North American Free Trade Agreement or Dominican Republic-Central America Free Trade Agreement?
---
\end{lstlisting}
\end{tcolorbox}
\caption{Few-shot prompt for decontextualizing DROP (history)}
\label{fig:question-rewrite-drop-history-fewshot-1}
\end{figure}

\begin{figure}[h!]
    \centering
\tcbset{colback=verylightgray, colframe=verylightgray, boxrule=0.5pt, arc=4pt}
\begin{tcolorbox}
\lstset{
    basicstyle=\ttfamily\color{black}\tiny,
    keywordstyle=\color{black},
    identifierstyle=\color{black},
    commentstyle=\color{black},
    stringstyle=\color{black},
    breaklines=true,
    backgroundcolor=\color{verylightgray},
    frame=none,
    numbers=none,
    escapechar=@
}
\begin{lstlisting}
Passage: The French king, John II, had been held captive in England. The Treaty of Br\u00e9tigny set his ransom at 3\u00a0million\u00a0crowns and allowed for hostages to be held in lieu of John. The hostages included two of his sons, several princes and nobles, four inhabitants of Paris, and two citizens from each of the nineteen principal towns of France. While these hostages were held, John returned to France to try and raise funds to pay the ransom. In 1362 John's son Louis of Anjou, a hostage in English-held Calais, escaped captivity. So, with his stand-in hostage gone, John felt honor-bound to return to captivity in England. The French crown had been at odds with Navarre  since 1354, and in 1363 the Navarrese used the captivity of John II in London and the political weakness of the Dauphin to try to seize power. Although there was no formal treaty, Edward III supported the Navarrese moves, particularly as there was a prospect that he might gain control over the northern and western provinces as a consequence. With this in mind, Edward deliberately slowed the peace negotiations. In 1364, John II died in London, while still in honourable captivity. Charles V succeeded him as king of France. On 7 May 1364, one month after the dauphin's accession and three days before his coronation as Charles V, the Navarrese suffered a crushing defeat at the Battle of Cocherel.
Old question: How many years passed between the French being at odds with Navarre and the Navarrese attempting to seize power?
New question: How many years passed between the French being at odds with Navarre in 1354 and the Navarrese attempting to seize power using the weakness of John II?
---
Passage: News of the two battles reached England in August. After several months of negotiations, the government of the Duke of Newcastle decided to send an army expedition the following year to dislodge the French. They chose Major General Edward Braddock to lead the expedition. Word of the British military plans leaked to France well before Braddock's departure for North America. In response, King Louis XV dispatched six regiments to New France under the command of Baron Dieskau in 1755. The British sent out their fleet in February 1755, intending to blockade French ports, but the French fleet had already sailed. Admiral Edward Hawke detached a fast squadron to North America in an attempt to intercept them. In a second British action, Admiral Edward Boscawen fired on the French ship Alcide on June 8, 1755, capturing her and two troop ships. The British harassed French shipping throughout 1755, seizing ships and capturing seamen. These actions contributed to the eventual formal declarations of war in spring 1756.
Old question: How many months were there between the blockade on the French ports and Admiral firing on Alcide
New question: How many months were there between the British attempt to blockade the French ports in 1755 and Admiral Edward Boscawen firing on the French ship Alcide?
---
Passage: Upper Austria had been rebellious for centuries, with 62 known uprisings between 1356 and 1849, 14 of which occurred in the 16th century. However, the Peasants' War of 1626 was the costliest in terms of human life and damage to livestock and property. The war caused Martin Aichinger to lose his farm and begin roaming the country. He eventually became a religious leader who led a popular revolt against aristocratic rule. His revolutionary ideas frightened the rulers so much that they tried to arrest him, leading to another series of uprisings that ended in the Battle on the Frankenberg  in 1636. All of Aichinger's followers were slaughtered during the battle, including the remaining women and children who had been in hiding.
Old question: How many years after the Peasants' War was the Battle on the Frankenberg?
New question: How many years after the Peasants' War of 1626 was the Battle on the Frankenberg?
---
Passage: In the United States, conscription began in 1917 and was generally well received, with a few pockets of opposition in isolated rural areas. The administration decided to rely primarily on conscription, rather than voluntary enlistment, to raise military manpower for when only 73,000 volunteers enlisted out of the initial 1 million target in the first six weeks of the war. In 1917 10 million men were registered. This was deemed to be inadequate, so age ranges were increased and exemptions reduced, and so by the end of 1918 this increased to 24 million men that were registered with nearly 3 million inducted into the military services. The draft was universal and included blacks on the same terms as whites, although they served in different units. In all 367,710 black Americans were drafted , compared to 2,442,586 white . Forms of resistance ranged from peaceful protest to violent demonstrations and from humble letter-writing campaigns asking for mercy to radical newspapers demanding reform. The most common tactics were dodging and desertion, and many communities sheltered and defended their draft dodgers as political heroes. Many socialists were jailed for \"obstructing the recruitment or enlistment service\". The most famous was Eugene Debs, head of the Socialist Party of America, who ran for president in 1920 from his prison cell. In 1917 a number of radicals and anarchists challenged the new draft law in federal court, arguing that it was a direct violation of the Thirteenth Amendment's prohibition against slavery and involuntary servitude. The Supreme Court unanimously upheld the constitutionality of the draft act in the Selective Draft Law Cases on January 7, 1918.
Old question: How many years after conscription began did Eugene Debs run for president?
New question: How many years after conscription began in the United States for World War I did Eugene Debs run for president?
---
\end{lstlisting}
\end{tcolorbox}
\caption{Few-shot prompt for decontextualizing DROP (history - continued)}
\label{fig:question-rewrite-drop-history-fewshot-2}
\end{figure}

\begin{figure}[h!]
    \centering
\tcbset{colback=verylightgray, colframe=verylightgray, boxrule=0.5pt, arc=4pt}
\begin{tcolorbox}
\lstset{
    basicstyle=\ttfamily\color{black}\tiny,
    keywordstyle=\color{black},
    identifierstyle=\color{black},
    commentstyle=\color{black},
    stringstyle=\color{black},
    breaklines=true,
    backgroundcolor=\color{verylightgray},
    frame=none,
    numbers=none,
    escapechar=@
}
\begin{lstlisting}
Given a Wikipedia passage and its title, rewrite the old question by integrating specific information from the passage such that the events in the question are less ambiguous and the user can answer the new question without reading the passage and using Google Search instead.
- For individuals that cannot be uniquely identified by their names, include small pieces of information that help to narrow the individual down.
- Make sure that any new information added does not directly answer the question itself.
- Make sure that the new question asks for the same information as the old question in the passage.
- Leave the question as it is if it is already specific enough to solve on its own without the passage.
- Do NOT mention "the passage" in the new question.
- Make sure that the new question requires multiple steps of reasoning to solve. For example, if the question asks about the birth date of a person, do NOT include the actual name of the person but other details of the person instead.
- Make sure that the new question is grammatical.
Below are a few examples of the generated output. Adhere to the format shown in the examples.
---
Title: Fredell Lack
Passage: Fredell Lack was born in Tulsa, Oklahoma, the oldest of three children of Jewish Eastern European (Latvian) immigrants, Abram I. Lack and Sarah Stillman Lack (who was a sister of noted painter Ary Stillman). She began violin lessons at age six, studying with Tosca Berger. When Fredell was 10, she moved with her family to Houston, Texas. There she studied with Josephine Boudreaux, the concertmaster of the Houston Symphony. At age 11, she first soloed with orchestra, performing the Wieniawski Concerto No. 2 with the Tulsa Philharmonic. At 12, Lack was accepted into the New York City studio of the legendary violinist and pedagogue Louis Persinger, whose other students included such artists as Yehudi Menuhin, Isaac Stern, and Ruggiero Ricci. She moved to New York and completed her pre-college schooling at the Bentley School while continuing her violin lessons with Persinger. At 17, she made her professional solo debut, playing the Mendelssohn Violin Concerto with the St. Louis Symphony. Subsequently she received a full scholarship to the Juilliard School in New York. She continued studying violin with Persinger there and also was deeply influenced by her study of chamber music with Felix Salmond. She received the Diploma from Juilliard at age 21.
Question: What is the population of the city where Lack was born?
Rewrite: What is the population of the city where Fredell Lack was born?
---
Title: John Ford filmography
Passage: John Ford (1894\u20131973) was an American film director whose career spanned from 1913 to 1971. During this time he directed more than 140 films. Born in Maine, Ford entered the filmmaking industry shortly after graduating from high school with the help of his older brother, Francis Ford, who had established himself as a leading man and director for Universal Studios. After working as an actor, assistant director, stuntman, and prop man \u2013 often for his brother \u2013 Universal gave Ford the opportunity to direct in 1917. Initially working in short films, he quickly moved into features, largely with Harry Carey as his star. In 1920 Ford left Universal and began working for the Fox Film Corporation. During the next ten years he directed more than 30 films, including the westerns The Iron Horse (1924) and 3 Bad Men (1926), both starring George O'Brien, the war drama Four Sons and the Irish romantic drama Hangman's House (both 1928 and both starring Victor McLaglen). In the same year of these last two films, Ford directed his first all-talking film, the short Napoleon's Barber. The following year he directed his first all-talking feature, The Black Watch.
Question: How old was Francis Ford when John Ford started his career in filmmaking?
Rewrite: How old was Francis Ford, John Ford's older brother, when John Ford started his career in filmmaking?
---
Title: George Glossop Walker
Passage: Walker's debut match for Derbyshire in the 1881 season was against Yorkshire when he never had the chance to bowl and scored 2 runs in each innings. He did not play again in that season nor in the 1882 season, and only played in two games in the 1883 season. In the 1884 and 1885 season, when William Cropper lead the bowling, he played more frequently and in 1885 took 7\u2013105 against Nottinghamshire in one match and 5\u201387 in the other. In 1886 Walker was selected for two Gentlemen of England teams, in one of which against Australia he was in the team with his hero W.G. Grace. For the county he took 6\u201326 against Marylebone Cricket Club (MCC), and 7\u201338 and 5\u201375 in the same match against Surrey. In the 1887 season Walker took 5\u201354 for Derbyshire against Lancashire and 5\u201349 against Surrey. He continued playing regularly for the Derbyshire club between 1888 and 1893 when it was without first-class status. In 1894 took 7\u2013108 for Gentlemen against Players with W. G. Grace in the side again although he never had the opportunity to bowl against him in any of his first-class games. He also took 5\u201324 for Derbyshire against Lancashire. In the 1896 season he took 9\u201385 against Leicestershire although his average was deteriorating. He played four games in the 1897 season and six in the 1898 season by which time his bowling made little impression, while Billy Bestwick was beginning to star.
Question: How many total games did Walker play in during the 1884 and 1885 seasons?
Rewrite: How many total games did George Glossop Walker play in during the 1884 and 1885 seasons?
---
Title: 446th Operations Group
Passage: The group was occasionally diverted from strategic missions to carry out air support and interdiction missions. It supported Operation Overlord, the invasion of Normandy by attacking transportation targets, including bridges, along with airfields and strong points in France. On D Day, the squadron and the rest of the 446th Group led the first heavy bomber mission of the day. The 446th aided ground forces at Caen and Saint-L\u00f4 during July by hitting bridges, gun batteries, and enemy troops. During Operation Market Garden, the attempt to seize a bridgehead across the Rhine in the Netherlands, the 704th dropped supplies to allied troops near Nijmegen. It struck lines of communications during the Battle of the Bulge. During Operation Varsity in March 1945, it supplied ground and airborne troops near Wesel. The squadron flew its last combat mission on 25 April 1945 against Salzburg, Austria. The group had flown 273 missions and had lost 58 aircraft during the war,\n
Question: When did the operation during which the 704th dropped supplies to allied troops near Nijmegen begin?
Rewrite: When did the operation during which the 704th dropped supplies to allied troops near Nijmegen begin?
---
\end{lstlisting}
\end{tcolorbox}
\vspace{-0.7em}
\caption{Few-shot prompt for decontextualizing IIRC}
\label{fig:question-rewrite-iirc-fewshot-1}
\end{figure}

\begin{figure}[h!]
    \centering
\tcbset{colback=verylightgray, colframe=verylightgray, boxrule=0.5pt, arc=4pt}
\begin{tcolorbox}
\lstset{
    basicstyle=\ttfamily\color{black}\tiny,
    keywordstyle=\color{black},
    identifierstyle=\color{black},
    commentstyle=\color{black},
    stringstyle=\color{black},
    breaklines=true,
    backgroundcolor=\color{verylightgray},
    frame=none,
    numbers=none,
    escapechar=@
}
\begin{lstlisting}
Title: Heartbreak on a Full Moon
Passage: After selling 25,000 copies and earning 68,000 album-equivalent units within three days, Heartbreak on a Full Moon debuted at number three on the US Billboard 200, becoming Brown's ninth consecutive top 10 album on the chart. The album was Brown's seventh solo album to debut at number one on the Billboard Top R&B/Hip-Hop Albums chart. On November 8, 2017, Heartbreak on a Full Moon was certified gold by the Recording Industry Association of America for combined sales and album-equivalent units of over 500,000 units in the United States (in this case, 250,000 double album sets, which are double-counted by the RIAA). Brown became the first R&B male artist that went gold in a week since Usher's Confessions in 2004. In its second chart week, the album remained at number three on Billboard 200, with 73,000 album-equivalent units. In Australia, it entered the ARIA Albums Chart at number five, becoming his first top ten in the nation since X in 2014. In the United Kingdom, the album debuted at number 10 on the UK Albums Chart, Brown's sixth non-consecutive top 10 album on the chart. The album was eventually certified Silver by the British Phonographic Industry (BPI) for sales of over 60,000 copies in the UK. In New Zealand, the album debuted at number three on the RMNZ Albums Chart, giving Brown his seventh top ten album on the chart. Until June 2018, the album has accumulated over 3\u00a0billion streams worldwide.
Question: When was the company that certified the song's \"gold\" status founded?
Rewrite: When was the company that certified the \"gold\" status of Heartbreak on a Full Moon founded?
---
Title: Rail transport in Israel
Passage: Rail infrastructure in what is now Israel was first envisioned and realized during the Ottoman period. Sir Moses Montefiore, in 1839, was an early proponent of trains in the land of Israel. However, the first railroad in Eretz Yisrael, was the Jaffa-Jerusalem railway, which opened on September 26, 1892. A trip along the line took 3 hours and 30 minutes. The line was initiated by the Jewish entrepreneur Joseph Navon and built by the French at 1\u00a0m gauge. The second line in what is now Israel was the Jezreel Valley railway from Haifa to Beit She\u2019an, which had been built in 1904 as part of the Haifa-Daraa branch, a 1905-built feeder line of the Hejaz Railway which ran from Medina to Damascus. At the time, the Ottoman Empire ruled the Levant, but was a declining power and would succumb in World War I. During the Ottoman era, the network grew: Nablus, Kalkiliya, and Beersheba all gained train stations. The First World War brought yet another rail line: the Ottomans, with German assistance, laid tracks from Beersheba to Kadesh Barnea, somewhere on the Sinai Peninsula. (This line ran through trains from Afula through Tulkarm.) This resulted in the construction of the eastern and southern railways.
Question: Was the first railroad line in Israel longer than the second line?
Rewrite: Was the first railroad line in Israel longer than the second railroad line in Israel?
---
Title: List of Indianapolis Colts starting quarterbacks
Passage: In 1998 the Colts, for the 4th time in 15 years, held the 1st overall pick in the draft and for the 3rd time in 15 years selected a quarterback \u2013 this time University of Tennessee's Peyton Manning. Manning started the first game of his rookie season and started every single Colts game since until the start of the 2011 season, when a recurring neck injury sidelined him. Despite a difficult rookie season, where he threw a league high 28 interceptions, Manning and the Colts responded by finishing 13\u20133 in 1999. The 10 game turnaround from the previous year set an NFL record. Even with this turnaround, the Colts lost in the playoffs. The following years would be marked by a near constant pattern. The Colts and Manning successes in the regular season were matched only by their failures in the post season. Manning was named to the Pro Bowl in 1999, 2000, 2002, 2003 and 2004, as well as winning the NFL MVP award in both 2003 and 2004. In 2004 Manning set a then NFL record when he threw 49 touchdowns in a single season. In spite of this the team failed in the playoffs, including early round exits in 1999, 2000, 2002 and 2005. In both 2003 and 2004 the Colts would lose to eventual Super Bowl winning New England Patriots in the AFC Championship Game and the Divisional Round respectively. In 2006 the Colts and Manning were finally able to beat the Patriots and their quarterback Tom Brady in the AFC Championship Game on their way to a victory in Super Bowl XLI against the Chicago Bears. Manning was named the Super Bowl MVP. The Colts and Manning would continue to have success, with Manning winning two further MVP awards in 2008 and 2009. In 2009 the Colts would return to the Super Bowl where they would lose to the New Orleans Saints.
Question: Who was the head coach of the team that the Colts lost to in the 2009 Superbowl?
Rewrite: Who was the head coach of the team that the Colts lost to in the 2009 Superbowl?
---
Title: 2008 Arizona Cardinals season
Passage: The 2008 Arizona Cardinals season was the 89th season for the team in the National Football League and their 21st season in Arizona. The season marked the Cardinals' first Super Bowl appearance, coming as a result of their victory against the Philadelphia Eagles in the NFC Championship. The Cardinals slogan for the season was \"Shock The World!\" Riding the back of quarterback Kurt Warner, who had gone from being a backup for the St. Louis Rams in 1999 to leading the Greatest Show on Turf to a Super Bowl XXXIV victory, and franchise wide receiver Larry Fitzgerald, the Cardinals went on a playoff run for the ages after having won just one playoff game in the last sixty years, as Warner once again recreated the magic he had captured with the Rams. (Coincidentally, both teams were based in St Louis at one point or another, only to relocate to different cities.)
Question: How many Super Bowls did the team the Cardinals beat to make their first appearance in the Super Bowl win?
Rewrite: How many Super Bowls did the team the Arizona Cardinals beat in the 2008 season to make their first appearance in the Super Bowl win?
---
\end{lstlisting}
\end{tcolorbox}
\caption{Few-shot prompt for decontextualizing IIRC (continued)}
\label{fig:question-rewrite-iirc-fewshot-2}
\end{figure}

\clearpage
\subsection{Examples}
We provide examples of question re-writes which result in DROP* and IIRC*.
\begin{figure}[h!]
    \centering
\tcbset{colback=verylightgray, colframe=verylightgray, boxrule=0.5pt, arc=4pt}
\begin{tcolorbox}
\lstset{
    basicstyle=\ttfamily\color{black}\tiny,
    keywordstyle=\color{black},
    identifierstyle=\color{black},
    commentstyle=\color{black},
    stringstyle=\color{black},
    breaklines=true,
    backgroundcolor=\color{verylightgray},
    frame=none,
    numbers=none,
    escapechar=@
}
\begin{lstlisting}
Passage: Hoping to rebound from their loss to the Patriots, the Raiders stayed at home for a Week 16 duel with the Houston Texans.  Oakland would get the early lead in the first quarter as quarterback JaMarcus Russell completed a 20-yard touchdown pass to rookie wide receiver Chaz Schilens.  The Texans would respond with fullback Vonta Leach getting a 1-yard touchdown run, yet the Raiders would answer with kicker Sebastian Janikowski getting a 33-yard and a 30-yard field goal.  Houston would tie the game in the second quarter with kicker Kris Brown getting a 53-yard and a 24-yard field goal. Oakland would take the lead in the third quarter with wide receiver Johnnie Lee Higgins catching a 29-yard touchdown pass from Russell, followed up by an 80-yard punt return for a touchdown.  The Texans tried to rally in the fourth quarter as Brown nailed a 40-yard field goal, yet the Raiders' defense would shut down any possible attempt.
old_question: Who scored the first touchdown of the game?
new_question: Who scored the first touchdown of the game between the Oakland Raiders and the Houston Texans where Sebastian Janikowski got a 33-yard and a 30-yard field goal?
\end{lstlisting}
\end{tcolorbox}
\caption{Example question re-write from DROP (sports)}
\label{fig:question-rewrite-drop-sports-example}
\end{figure}

\begin{figure}[h!]
    \centering
\tcbset{colback=verylightgray, colframe=verylightgray, boxrule=0.5pt, arc=4pt}
\begin{tcolorbox}
\lstset{
    basicstyle=\ttfamily\color{black}\tiny,
    keywordstyle=\color{black},
    identifierstyle=\color{black},
    commentstyle=\color{black},
    stringstyle=\color{black},
    breaklines=true,
    backgroundcolor=\color{verylightgray},
    frame=none,
    numbers=none,
    escapechar=@
}
\begin{lstlisting}
Passage: In 1905, 1,003 Korean immigrants, which included 802 men and 231 women and children, departed from the port of Chemulpo, Incheon aboard the ship Ilford to Salina Cruz, Oaxaca, Mexico. The journey took 45 days, after which they took a train to Coatzacoalcos, Veracruz. In the Veracruz port, another boat was taken to the port of Progreso with the final destination being the capital city of M\u00e9rida, Yucatan. They arrived in May 1905, with previously signed contracts for four years' work as indentured laborers on the Yucat\u00e1n henequen haciendas. Many of these Koreans were distributed throughout the Yucat\u00e1n in 32 henequen haciendas. The town of Motul, Yucatan, located in the heart of the henequen zone, was a destination for many of the Korean immigrants. Subsequently, in 1909, at the end of their contracts, they began a new stage in which they scattered even further  Thus, the majority of those who came were single men who made or remade their family lives with Yucatecan especially Maya women. While Korean girls were much more subject to marriages arranged by Korean parents, males had greater freedom when it came to making a family. This rapid intermarriage by Koreans, coupled with geographic dispersal, prevented the establishment of close social networks among these migrants and therefore provided the basis for Korean descendants among the Yucatan Peninsula. After that 1905 ship, no further entries of Koreans into Mexico were recorded, until many years later, leading to a new community of Koreans with completely different characteristics from those who entered in 1905. These descendants have started the Museo Conmemorativo de la Inmigraci\u00f3n Coreana a Yucat\u00e1n, a museum for the remembrance of their ancestors journey.
old question: How many years did the immigrants have to work as indentured laborers?
new_question: How many years did the Korean immigrants have to work as indentured laborers on the Yucat\u00e1n henequen haciendas after arriving in 1905?
\end{lstlisting}
\end{tcolorbox}
\caption{Example question re-write from DROP (history)}
\label{fig:question-rewrite-drop-history-example}
\end{figure}

\begin{figure}[h!]
    \centering
\tcbset{colback=verylightgray, colframe=verylightgray, boxrule=0.5pt, arc=4pt}
\begin{tcolorbox}
\lstset{
    basicstyle=\ttfamily\color{black}\tiny,
    keywordstyle=\color{black},
    identifierstyle=\color{black},
    commentstyle=\color{black},
    stringstyle=\color{black},
    breaklines=true,
    backgroundcolor=\color{verylightgray},
    frame=none,
    numbers=none,
    escapechar=@
}
\begin{lstlisting}
Title: 1965 New Zealand Grand Prix
Passage: It was the 12th New Zealand Grand Prix, doubled as the opening round of the 1965 Tasman Series. The race attracted 19 starters, including several overseas based drivers and teams. A large contingent of cars from Australia competed, including Frank Gardner competing for Alec Mildren Racing. Lex Davison and Leo Geoghegan brought across their own teams, while 1962 Formula One world champion, British racer Graham Hill race a Brabham for David McKay's Scuderia Veloce team. Star attraction though was the appearance of Team Lotus with their lead driver, 1963 World Champion, Jim Clark. Local honour was upheld by Bruce McLaren, who in an early iteration of the later McLaren team brought a pair of factory supported Coopers to race with American racer, the 1961 World Champion Phil Hill as his number two. The race was won by Graham Hill, his first victory in the NZGP. Gardner finished second to be the first 'antipodean' while first New Zealander was domestic series racer Jim Palmer in a career highlight as Brabham racing cars clean swept the podium.
old question: How many more wins did the 1961 World Champion have than the 1963 world champion?
new_question: How many more wins did the 1961 World Champion have than the 1963 World Champion in the 1965 New Zealand Grand Prix?
\end{lstlisting}
\end{tcolorbox}
\caption{Example question re-write from IIRC}
\label{fig:question-rewrite-iirc-example}
\end{figure}

\section{\dataset}
\label{sec:husky-curation}
\subsection{Dataset Construction}
We construct \dataset from a manually curated list of 40 topics as shown in Table~\ref{tab:huskyqa-topics}.
We first generate 10 factoid questions for each topic-subtopic pair in Table~\ref{tab:huskyqa-topics} using \texttt{gpt-3.5-turbo-0125}.
Figure~\ref{fig:huskyqa-question-generation-fewshot} shows our few-shot prompt for obtaining the seed questions.

Next, we use SERP API to obtain the answer to each factoid question, similarly to how we use SERP API in Appendix~\ref{sec:search-info}.
We then use the few-shot prompt provided in Figure~\ref{fig:huskyqa-factoid-generation-fewshot} to generate a factoid statement for each (question, answer) pair in \dataset.

Finally, we generate complex numerical reasoning questions from pairs of factoid statements within the same (topic, subtopic) category.
We use all unique pairs of the 10 factoid statements within each category, resulting in 45 candidate final questions from each category and a total of 1,800 questions across all categories.
We use the prompt provided in Figure~\ref{fig:huskyqa-final-question-generation-fewshot} to generate the final questions.

The final questions are generated such that the answer within each factoid statement is not provided in the final question and must be determined separately before solving the final question.
As a result, our final questions contain pieces of missing information that must first be identified by the agent, and then must be integrated in numerical reasoning for the agent to solve the final questions successfully.

We split the train and test set by topics. 
Our test set includes a total of 8 topics -- population (country), economics (revenue - company), buildings (skyscraper - height), geography (state - area), sports (soccer - goals), politics (number of senators - country), books (number of chapters) and chemistry (melting point - solid).
By splitting the dataset across separate topics, we ensure that the any model or agent trained on the dataset has not seen information related to topics in the evaluation set and is forced to generalize to topics outside of its training set.

We collect a total of 1,350 questions for the train set which is only used to provide examples for the action generator (see Table~\ref{tab:training-data-composition}) upon generating the solution trajectory, and 450 questions for the test set.
After filtering the questions in the test set with human annotators checking the soundness of the question as well as labeling the correct answer to each question, we have 292 questions remaining.

\begin{table}[]
    \setlength{\tabcolsep}{3pt}
    \centering
    \footnotesize
    \begin{tabular}{ll}
    \toprule
        \textbf{topic} & \textbf{subtopics} \\\midrule
        price & house, car, market value (soccer player), net worth (football/basketball/baseball player)\\
        population & country, state, city, college, company\\
        economics & GDP (country), revenue (company), income (household)\\
        buildings & stadium (capacity), skyscraper (height), skyscraper (number of floors), bridges (length) \\
        geography & country (area), state (area), river (length), lake (area), sea (area) \\
        sports & soccer (goals), football (touchdowns), basketball (points), baseball (home runs) \\
        history & founding year (country) \\
        politics & number of senators (country) \\
        animals & number of legs, lifespan \\
        books & number of chapters, number of pages \\
        instruments & number of strings/keyboards/holes \\
        astronomy & planet (number of moons) \\
        chemistry & melting point (solid), boiling point (liquid), atomic number \\
        biology & human anatomy (number of X's) \\
        law & maximum sentence (federal law) \\
        \bottomrule
    \end{tabular}
    \vspace{0.5em}
    \caption{List of seed topics and sub-topics for \dataset.}
    \label{tab:huskyqa-topics}
\end{table}

\begin{figure}[h!]
    \centering
\tcbset{colback=verylightgray, colframe=verylightgray, boxrule=0.5pt, arc=4pt}
\begin{tcolorbox}
\lstset{
    basicstyle=\ttfamily\color{black}\tiny,
    keywordstyle=\color{black},
    identifierstyle=\color{black},
    commentstyle=\color{black},
    stringstyle=\color{black},
    breaklines=true,
    backgroundcolor=\color{verylightgray},
    frame=none,
    numbers=none,
    escapechar=@
}
\begin{lstlisting}
Given a topic, generate a list of 10 related factual questions that result in numerical answers. The questions should be of the same topic but should not be the exact same questions. Make sure that the units are the same for the answers to the questions.
Below are a few examples of the generated output. Adhere to the format shown in the examples.
---
Topic: geography (state - area)
Questions
Q1: What is the area of California in square miles?
Q2: What is the area of Texas in square miles?
Q3: What is the area of New York in square miles?
Q4: What is the area of Florida in square miles?
Q5: What is the area of Washington in square miles?
Q6: What is the area of Arizona in square miles?
Q7: What is the area of Wyoming in square miles?
Q8: What is the area of Colorado in square miles?
Q9: What is the area of Pennsylvania in square miles?
Q10: What is the area of Ohio in square miles?
---
Topic: price (car)
Questions
Q1: What is the MSRP of the 2024 Hyundai Kona in U.S. dollars?
Q2: What is the MSRP of the 2024 Toyota Camry in U.S. dollars?
Q3: What is the MSRP of the 2024 Honda Accord in U.S. dollars?
Q4: What is the MSRP of the 2024 Chevrolet Trax in U.S. dollars?
Q5: What is the MSRP of the 2024 Lucid Air in U.S. dollars?
Q6: What is the MSRP of the 2024 Hyundai Sonata in U.S. dollars?
Q7: What is the MSRP of the 2024 Kia Seltos in U.S. dollars?
Q8: What is the MSRP of the 2024 Honda Ridgeline in U.S. dollars?
Q9: What is the MSRP of the 2024 Kia Telluride in U.S. dollars?
Q10: What is the MSRP of the 2024 Ford F-150 in U.S. dollars?
---
Topic: sports (soccer)
Questions
Q1: How many goals did Kylian Mbappe score in Ligue 1 during the 2023-2024 season?
Q2: How many goals did Erling Haaland score in the Premier League during the 2022-2023 season?
Q3: How many goals did Kevin De Bruyne score in the Premier League during the 2022-2023 season?
Q4: How many goals did Harry Kane score in Bundesliga during the 2023-2024 season?
Q5: How many goals did Son Heung-min score in the Premier League during the 2023-2024 season?
Q6: How many goals did Vinicius Jr. score in La Liga during the 2023-2024 season?
Q7: How many goals did Leroy Sane score in Bundesliga during the 2022-2023 season?
Q8: How many goals did Jude Bellingham score in La Liga during the 2023-2024 season?
Q9: How many goals did Florian Wirtz score in Bundesliga during the 2023-2024 season?
Q10: How many goals did Cole Palmer score in the Premier League during the 2023-2024 season?
---
Topic: price (house)
Questions
Q1: What is the median house price in New York in 2024?
Q2: What is the median house price in California in 2024?
Q3: What is the median house price in Texas in 2024?
Q4: What is the median house price in Florida in 2024?
Q5: What is the median house price in Georgia in 2024?
Q6: What is the median house price in Massachusetts in 2024?
Q7: What is the median house price in Pennsylvania in 2024?
Q8: What is the median house price in Washington in 2024?
Q9: What is the median house price in Illinois in 2024?
Q10: What is the median house price in Colorado in 2024?
---
\end{lstlisting}
\end{tcolorbox}
\caption{Few-shot prompt for generating seed questions for \dataset}
\label{fig:huskyqa-question-generation-fewshot}
\end{figure}

\begin{figure}[h!]
    \centering
\tcbset{colback=verylightgray, colframe=verylightgray, boxrule=0.5pt, arc=4pt}
\begin{tcolorbox}
\lstset{
    basicstyle=\ttfamily\color{black}\tiny,
    keywordstyle=\color{black},
    identifierstyle=\color{black},
    commentstyle=\color{black},
    stringstyle=\color{black},
    breaklines=true,
    backgroundcolor=\color{verylightgray},
    frame=none,
    numbers=none,
    escapechar=@
}
\begin{lstlisting}
Given a factual question and the corresponding answer, rewrite the answer into a complete sentence.
Below are a few examples of the generated output. Adhere to the format shown in the examples.
---
Question: What is the area of Texas in square miles?
Answer: 268,597 square miles
Rewrite: The area of Texas is 268,597 square miles.
---
Question: What is the MSRP of the 2024 Hyundai Kona in U.S. dollars?
Answer: $24,250
Rewrite: The MSRP of the 2024 Hyundai Kona in U.S. dollars is $24,250.
---
Question: How many goals did Kylian Mbappe score in Ligue 1 during the 2023-2024 season?
Answer: 26 goals
Rewrite: Kylian Mbappe scored 26 goals in Ligue 1 during the 2023-2024 season.
---
Question: What is the median house price in New York in 2024 in U.S. dollars?
Answer: $785,000
Rewrite: The median house price in New York in 2024 in U.S. dollars is $785,000.
---
\end{lstlisting}
\end{tcolorbox}
\caption{Few-shot prompt for generating factoid statements for \dataset}
\label{fig:huskyqa-factoid-generation-fewshot}
\end{figure}

\begin{figure}[h!]
    \centering
\tcbset{colback=verylightgray, colframe=verylightgray, boxrule=0.5pt, arc=4pt}
\begin{tcolorbox}
\lstset{
    basicstyle=\ttfamily\color{black}\tiny,
    keywordstyle=\color{black},
    identifierstyle=\color{black},
    commentstyle=\color{black},
    stringstyle=\color{black},
    breaklines=true,
    backgroundcolor=\color{verylightgray},
    frame=none,
    numbers=none,
    escapechar=@
}
\begin{lstlisting}
Given two pieces of factual statements, create a math question of moderate difficulty (at the level of middle school or high school math) that integrates the numbers from each statement.
Below are a few examples of the generated output. Adhere to the format shown in the examples.
---
Fact 1: The median value of a home in the United States in 2000 was $119,600.
Fact 2: The median value of a home in the United States in 2020 was $452,400.
Question: Joe bought a house in the year 2000 at the median price of the United States that year. Then, he sold the house at the median price in 2020. How much did the price of the house increase over this period on average per year?
---
Fact 1: George Washington was born in 1732.
Fact 2: Abraham Lincoln was born in 1809.
Question: How many years between the year George Washington was born and the year Abraham Lincoln was born are multiples of 4?
---
Fact 1: The Camp Nou has a capacity of 99,354.
Fact 2: The Santiago Bernabeu has a capacity of 81,044.
Question: FC Barcelona decides to allocate 85,000 seats for its home fans during its soccer match against Real Madrid, leaving the rest of the seats for the away fans. Real Madrid agrees to allocate the same number of away seats during its home match against FC Barcelona at the Santiago Bernabeu. How many seats are available in the Santiago Bernabeu for the home fans?
---
Fact 1: New York City had a population of 8.336 million as of 2022.
Fact 2: New York City had a population of 8.042 million as of 2002.
Question: If New York City grew by equal numbers of people between the years 2002 and 2022, what would have been its population in 2006?
---
\end{lstlisting}
\end{tcolorbox}
\caption{Few-shot prompt for generating final questions for \dataset}
\label{fig:huskyqa-final-question-generation-fewshot}
\end{figure}

\subsection{Examples}
We provide examples in the \dataset evaluation set in Table~\ref{tab:huskyqa-examples}.

\begin{table}[]
    \setlength{\tabcolsep}{3pt}
    \centering
    \scriptsize
    \begin{tabular}{lp{1.7cm}p{10cm}}
    \toprule
    \textbf{topic} & \textbf{subtopic} & \textbf{Question} \\\midrule
    price & market value - soccer player & A soccer team wants to purchase both Lionel Messi and Kevin De Bruyne. If they already have 20,000,000 euros and can save 15,000,000 euros every year, how many years will it take for them to have enough money to buy both players?\\
    population & country & If a plane can carry 300 passengers and is tasked with transporting 1 percent of the difference between the populations of Indonesia and Brazil in 2022 to a global conference, how many flights will it take to transport all these passengers?\\
    economics & revenue (company) & If Amazon decided to invest an amount equal to Tesla's total revenue in 2021 into new projects, what percentage of Amazon's total revenue for 2021 would this investment represent?\\
    buildings & skyscraper (height) & If a model of the Willis Tower is made at a scale of 1:100 and a model of the One World Trade Center is made at the same scale, by how many feet is the model of the One World Trade Center taller than the model of the Willis Tower?\\
    buildings & stadium (capacity) & If a football match between the Dallas Cowboys and an international team was organized where they decided to split the capacity evenly between home and away fans in the AT\&T Stadium, and then a rematch was scheduled at the Allianz Arena with the same ticket allocation policy, how many fewer seats would be available for each team's fans in the Allianz Arena compared to the AT\&T Stadium?\\
    sports & soccer (goals) & If Kylian Mbappe and Erling Haaland continue to score at their 2022-2023 season rates for the next 3 seasons, how many more goals will Haaland have scored than Mbappe after these 3 seasons?\\
    politics & number of senators (country) & If a conference is being held where only 40 percent of United States senators and 25 percent of Mexican senators are invited, how many senators in total will attend the conference?\\
    books & number of chapters & If a book club decides to read \"Pride and Prejudice\" and \"1984\" back to back, dedicating 3 days to discuss each chapter of \"Pride and Prejudice\" and 4 days to discuss each chapter of \"1984\", how many total days will the book club spend on both books?\\
    chemistry & melting point (solid) & A blacksmith decides to create an alloy consisting of 60 percent iron and 40 percent copper by mass. If the melting point of the alloy is directly proportional to the melting points of the individual metals based on their percentages in the alloy, what is the melting point of the alloy?",Find the melting point of iron in degrees Celsius.\\
    geography & state (area) & If a map designer wants to create a single page in an atlas with both California and New York at the same scale, and the page has an area of 310 square inches, how many square inches of the page would New York take up if California takes up its proportional share?\\
    
    \bottomrule
    \end{tabular}
    \vspace{0.5em}
    \caption{Example questions in \dataset.}
    \label{tab:huskyqa-examples}
\end{table}

\clearpage
\section{Training Data Examples}
\label{sec:training-data-examples}
\subsection{Action Generator}
We provide examples of the training data for the action generator in Figures~\ref{fig:training-data-action-generator-math} and~\ref{fig:training-data-action-generator-knowledge}.
\begin{figure}[h!]
    \centering
\tcbset{colback=verylightgray, colframe=verylightgray, boxrule=0.5pt, arc=4pt}
\begin{tcolorbox}
\lstset{
    basicstyle=\ttfamily\color{black}\tiny,
    keywordstyle=\color{black},
    identifierstyle=\color{black},
    commentstyle=\color{black},
    stringstyle=\color{black},
    breaklines=true,
    backgroundcolor=\color{verylightgray},
    frame=none,
    numbers=none,
    escapechar=@
}
\begin{lstlisting}
@\textbf{Prompt}@
Given the input question and the solution history that consists of steps for solving the input question and their corresponding outputs, decide whether the solution history has already solved the original question. If the original question has not been solved yet, assign a tool (either [math], [code], [search] or [commonsense]) and generate the next step that needs to be answered to solve the original question. Do not generate a step that has already been written in the solution history. Otherwise, if the original question has already been solved, return the [finish] tool, along with the final answer to the original question based on the solution history.
- [math] is for: 1) solving math questions, writing or re-organizing equations, performing abstract reasoning such as case-by-case analysis, or identifying the conditions given in the question.
- [code] is for: 1) computing large numbers (at least 100), fractions or decimals. 2) counting or averaging long lists of numbers. 3) performing date-related operations, such as counting the number of days between two dates.
- [search] is for: retrieving specific knowledge from the Web to answer questions related to history, sports, culture, geography, medicine, science, etc.
- [commonsense] is for: applying commonsense knowledge to reason about a relatively simple step, such as comparing two numbers or recalling a widely-known fact.
- [finish] is for: indicating that the question has been solved, and it is followed by the answer to the question.
---
Question: A triangle in a Cartesian coordinate plane has vertices (5, -2), (10, 5) and (5, 5). How many square units are in the area of the triangle? Express your answer as a decimal to the nearest tenth.
Solution history:
Step: Calculate the base and height of the triangle.
Output: The base of the triangle is 5 units, and the height is 7 units.
Next step or final answer: 

@\textbf{Completion}@
[math] Calculate the area of the triangle using the base and height.
\end{lstlisting}
\end{tcolorbox}
\caption{Example training data for the action generator (math)}
\label{fig:training-data-action-generator-math}
\end{figure}

\begin{figure}[h!]
    \centering
\tcbset{colback=verylightgray, colframe=verylightgray, boxrule=0.5pt, arc=4pt}
\begin{tcolorbox}
\lstset{
    basicstyle=\ttfamily\color{black}\tiny,
    keywordstyle=\color{black},
    identifierstyle=\color{black},
    commentstyle=\color{black},
    stringstyle=\color{black},
    breaklines=true,
    backgroundcolor=\color{verylightgray},
    frame=none,
    numbers=none,
    escapechar=@
}
\begin{lstlisting}
@\textbf{Prompt}@
Given the input question and the solution history that consists of steps for solving the input question and their corresponding outputs, decide whether the solution history has already solved the original question. If the original question has not been solved yet, assign a tool (either [math], [code], [search] or [commonsense]) and generate the next step that needs to be answered to solve the original question. Do not generate a step that has already been written in the solution history. Otherwise, if the original question has already been solved, return the [finish] tool, along with the final answer to the original question based on the solution history.
- [math] is for: 1) solving math questions, writing or re-organizing equations, performing abstract reasoning such as case-by-case analysis, or identifying the conditions given in the question.
- [code] is for: 1) computing large numbers (at least 100), fractions or decimals. 2) counting or averaging long lists of numbers. 3) performing date-related operations, such as counting the number of days between two dates.
- [search] is for: retrieving specific knowledge from the Web to answer questions related to history, sports, culture, geography, medicine, science, etc.
- [commonsense] is for: applying commonsense knowledge to reason about a relatively simple step, such as comparing two numbers or recalling a widely-known fact.
- [finish] is for: indicating that the question has been solved, and it is followed by the answer to the question.
---
Question: How many world championships have been won by the team that the Reds outscored 22-8 in the 1976 World Series?
Solution history:
Step: Identify the team that the Reds outscored 22-8 in the 1976 World Series.
Output: The team that the Reds outscored 22-8 in the 1976 World Series is the New York Yankees.
Next step or final answer:

@\textbf{Completion}@
[search] Find the number of World Championships won by the New York Yankees.
\end{lstlisting}
\end{tcolorbox}
\caption{Example training data for the action generator (knowledge)}
\label{fig:training-data-action-generator-knowledge}
\end{figure}

\subsection{Code Generator}
We provide examples of the training data for the code generator in Figures~\ref{fig:training-data-code-generator-math} and~\ref{fig:training-data-code-generator-knowledge}.
\begin{figure}[h!]
    \centering
\tcbset{colback=verylightgray, colframe=verylightgray, boxrule=0.5pt, arc=4pt}
\begin{tcolorbox}
\lstset{
    basicstyle=\ttfamily\color{black}\tiny,
    keywordstyle=\color{black},
    identifierstyle=\color{black},
    commentstyle=\color{black},
    stringstyle=\color{black},
    breaklines=true,
    backgroundcolor=\color{verylightgray},
    frame=none,
    numbers=none,
    escapechar=@
}
\begin{lstlisting}
@\textbf{Prompt}@
Given the input question, the solution history that consists of steps for solving the input question and their corresponding outputs, and the current step that must be addressed to solve the input question, write code that solves the current step.
- Write the code in Python.
- Do not attempt to write code that directly answers the question. Write code that answers the given step.
- For math questions, utilize the 'pi' symbol and 'Rational' from the sympy package for $\\pi$ and fractions, and simplify all fractions and square roots without converting them to decimal values.
- Example imports are provided below. Import any of these packages, as well as additional packages as needed.
```imports omitted```
---
Question: Al, Betty, and Clare split $\\$1000$ among them to be invested in different ways. Each begins with a different amount. At the end of one year they have a total of $\\$1500$. Betty and Clare have both doubled their money, whereas Al has managed to lose $\\$100$. What was Al's original portion?
Solution history:
Step: Write equations based on the given information, using $A$ to represent Al's original portion, $B$ to represent Betty's original portion, and $C$ to represent Clare's original portion.
Output: The total initial amount is $A + B + C = 1000$. After one year, the total amount is $A - 100 + 2B + 2C = 1500$. Betty and Clare have both doubled their money, whereas Al has lost $100. \\boxed{A + B + C = 1000, A - 100 + 2B + 2C = 1500}$
Current step: Solve the system of equations to find Al's original portion.
Code:

@\textbf{Completion}@
from sympy import symbols, Eq, solve
A, B, C = symbols('A B C')
eq1 = Eq(A + B + C, 1000)
eq2 = Eq(A - 100 + 2*B + 2*C, 1500)
solution = solve((eq1, eq2), (A, B, C))
out = solution[A]
print(out)
\end{lstlisting}
\end{tcolorbox}
\caption{Example training data for the code generator (math)}
\label{fig:training-data-code-generator-math}
\end{figure}
\begin{figure}[h!]
    \centering
\tcbset{colback=verylightgray, colframe=verylightgray, boxrule=0.5pt, arc=4pt}
\begin{tcolorbox}
\lstset{
    basicstyle=\ttfamily\color{black}\tiny,
    keywordstyle=\color{black},
    identifierstyle=\color{black},
    commentstyle=\color{black},
    stringstyle=\color{black},
    breaklines=true,
    backgroundcolor=\color{verylightgray},
    frame=none,
    numbers=none,
    escapechar=@
}
\begin{lstlisting}
@\textbf{Prompt}@
Given the input question, the solution history that consists of steps for solving the input question and their corresponding outputs, and the current step that must be addressed to solve the input question, write code that solves the current step.
- Write the code in Python.
- Do not attempt to write code that directly answers the question. Write code that answers the given step.
- For math questions, utilize the 'pi' symbol and 'Rational' from the sympy package for $\\pi$ and fractions, and simplify all fractions and square roots without converting them to decimal values.
- Example imports are provided below. Import any of these packages, as well as additional packages as needed.
```imports omitted```
---
Question: How many months did the Anglo-French offensive of the Battle of the Somme in 1916 last?
Solution history:
Step: Find the start and end dates of the Anglo-French offensive of the Battle of the Somme in 1916.
Output: The Anglo-French offensive of the Battle of the Somme in 1916 started on July 1 and ended on November 18, 1916.
Current step: Calculate the number of months that elapsed between July 1, 1916, and November 18, 1916.
Code:

@\textbf{Completion}@
from datetime import datetime
# define the start and end dates
start_date = datetime(1916, 7, 1)
end_date = datetime(1916, 11, 18)
# calculate the difference in months
total_months = (end_date.year - start_date.year) * 12 + end_date.month - start_date.month
print(total_months)
\end{lstlisting}
\end{tcolorbox}
\caption{Example training data for the code generator (knowledge)}
\label{fig:training-data-code-generator-knowledge}
\end{figure}

\subsection{Math Reasoner}
We provide an example of the training data for the math reasoner in Figure~\ref{fig:training-data-math-reasoner}.
\begin{figure}[h!]
    \centering
\tcbset{colback=verylightgray, colframe=verylightgray, boxrule=0.5pt, arc=4pt}
\begin{tcolorbox}
\lstset{
    basicstyle=\ttfamily\color{black}\tiny,
    keywordstyle=\color{black},
    identifierstyle=\color{black},
    commentstyle=\color{black},
    stringstyle=\color{black},
    breaklines=true,
    backgroundcolor=\color{verylightgray},
    frame=none,
    numbers=none,
    escapechar=@
}
\begin{lstlisting}
@\textbf{Prompt}@
Given the original question, the solution history that consists of steps for solving the input question and their corresponding outputs, and the current step that must be addressed to solve the input question, answer the current step by reasoning step-by-step.
- Make sure to answer the subquestion and not the original question.
- Do not attempt to directly answer the original question unless the subquestion asks for the same thing as the original question.
- Present the answer \"ANS\" to the subquestion in LaTeX using the format 'The answer is \\boxed{ANS}.' without any units in the box.
---
Question: Simplify $\\left(\\dfrac{-1+i\\sqrt{3}}{2}\\right)^6+\\left(\\dfrac{-1-i\\sqrt{3}}{2}\\right)^6.$
Solution history:
None
Current step: Recognize that the expressions are complex conjugates and use De Moivre's theorem to simplify the expression.
Solution:

@\textbf{Completion}@
The given expressions $\\left(\\dfrac{-1+i\\sqrt{3}}{2}\\right)^6$ and $\\left(\\dfrac{-1-i\\sqrt{3}}{2}\\right)^6$ are indeed complex conjugates of each other. The complex number $\\dfrac{-1+i\\sqrt{3}}{2}$ can be recognized as the polar form $re^{i\\theta}$ where $r=\\sqrt{\\left(\\frac{-1}{2}\\right)^2+\\left(\\frac{\\sqrt{3}}{2}\\right)^2}=1$ and $\\theta=\\frac{2\\pi}{3}$, since it corresponds to the cosine and sine of $\\frac{2\\pi}{3}$ in the complex plane.

By De Moivre's theorem, $(re^{i\\theta})^n = r^n e^{in\\theta}$. Since $r=1$, $r^n=1^n=1$ for any integer $n$. Therefore, we can simplify the expressions as follows:
1. $\\left(\\dfrac{-1+i\\sqrt{3}}{2}\\right)^6 = e^{i\\cdot 6 \\cdot \\frac{2\\pi}{3}} = e^{i4\\pi} = \\cos(4\\pi) + i\\sin(4\\pi) = 1 + i\\cdot 0 = 1$.
2. Similarly, the complex conjugate $\\left(\\dfrac{-1-i\\sqrt{3}}{2}\\right)^6$ can be seen as having an angle of $-\\frac{2\\pi}{3}$ (or equivalently, $2\\pi - \\frac{2\\pi}{3} = \\frac{4\\pi}{3}$ when considering positive angles in the standard form). Therefore, it simplifies to $e^{i\\cdot 6 \\cdot (-\\frac{2\\pi}{3})} = e^{i\\cdot 6 \\cdot \\frac{4\\pi}{3}} = e^{i4\\pi} = \\cos(4\\pi) + i\\sin(4\\pi) = 1 + i\\cdot 0 = 1$.

Adding these two results together, we get $1 + 1 = 2$.
The answer is $\\boxed{2}$.
\end{lstlisting}
\end{tcolorbox}
\caption{Example training data for the math reasoner}
\label{fig:training-data-math-reasoner}
\end{figure}
\subsection{Query Generator}
We provide an example of the training data for the query generator in Figure~\ref{fig:training-data-query-generator}.
\begin{figure}[h!]
    \centering
\tcbset{colback=verylightgray, colframe=verylightgray, boxrule=0.5pt, arc=4pt}
\begin{tcolorbox}
\lstset{
    basicstyle=\ttfamily\color{black}\tiny,
    keywordstyle=\color{black},
    identifierstyle=\color{black},
    commentstyle=\color{black},
    stringstyle=\color{black},
    breaklines=true,
    backgroundcolor=\color{verylightgray},
    frame=none,
    numbers=none,
    escapechar=@
}
\begin{lstlisting}
@\textbf{Prompt}@
Given the input question, the solution history that consists of steps for solving the input question and their corresponding outputs, and the current step that needs to be taken to solve the question, write a concise, informative Google Search query for obtaining information regarding the current step.
---
Question: In which county is the school where Matt Langel worked with Dunphy in 2006?
Solution history:
Step: Identify the school where Matt Langel worked with Dunphy in 2006.
Output: The school where Matt Langel worked with Dunphy in 2006 is the University of Pennsylvania.
Current step: Determine in which county the University of Pennsylvania is located.

@\textbf{Completion}@
University of Pennsylvania located in which county
\end{lstlisting}
\end{tcolorbox}
\caption{Example training data for the query generator}
\label{fig:training-data-query-generator}
\end{figure}

\clearpage
\section{\agent Inference Examples}
\label{sec:husky-inference-examples}
We provide examples of \agent inference in Figures~\ref{fig:husky-inference-numerical},~\ref{fig:husky-inference-tabular},~\ref{fig:husky-inference-knowledge} and~\ref{fig:husky-inference-mixed}.
\vspace{-0.5em}
\subsection{Numerical Reasoning}
\vspace{-0.5em}
\begin{figure}[h!]
    \centering
\tcbset{colback=verylightgray, colframe=verylightgray, boxrule=0.5pt, arc=4pt}
\begin{tcolorbox}
\lstset{
    basicstyle=\ttfamily\color{black}\tiny,
    keywordstyle=\color{black},
    identifierstyle=\color{black},
    commentstyle=\color{black},
    stringstyle=\color{black},
    breaklines=true,
    backgroundcolor=\color{verylightgray},
    frame=none,
    numbers=none,
    escapechar=@
}
\begin{lstlisting}
@\textbf{Question}@: Find the sum of all integers that satisfy these conditions: \\[\n|x|+1>7\\text{ and }|x+1|\\le7.\n\\]
@\textbf{Solution}@:
Step 1: [math] Translate the given inequalities into mathematical expressions that can be solved.
1. For the first inequality, $|x| + 1 > 7$, we subtract 1 from both sides to get $|x| > 6$. This means that the absolute value of $x$ must be greater than 6. In terms of real numbers, this translates to $x > 6$ or $x < -6$.
2. For the second inequality, $|x + 1| \\le 7$, we consider the definition of absolute value, which states that $|x + 1| \\le 7$ means that $-7 \\le x + 1 \\le 7$. Subtracting 1 from all parts of the inequality gives $-8 \\le x \\le 6$.
Combining these two sets of conditions, we find that $x$ must satisfy both $x > 6$ and $-8 \\le x \\le 6$. The intersection of these conditions is $-8 \\le x \\le 6$, which means $x$ can range from -8 to 6, inclusive.
Therefore, the answer is $\\boxed{-8 \\le x \\le 6}$.

Step 2: [math] Identify the integers that satisfy the given conditions.
The conditions given are $-8 \\le x \\le 6$. This means that $x$ can be any integer from -8 up to 6, inclusive. 
To list all the integers that satisfy these conditions, we start from the smallest integer that is greater than -8, which is -8 itself, and go up to the largest integer that is less than or equal to 6, which is 6.
Therefore, the integers that satisfy the given conditions are $\\boxed{-8, -7, -6, -5, -4, -3, -2, -1, 0, 1, 2, 3, 4, 5, 6}$.

Step 3: [code] Compute the sum of all integers that satisfy the conditions.
```python
integer_list = [-8, -7, -6, -5, -4, -3, -2, -1, 0, 1, 2, 3, 4, 5, 6]
out = sum(integer_list)
print(out)
```
The sum of all integers that satisfy the given conditions is -15.

The answer is: -15.
\end{lstlisting}
\end{tcolorbox}
\vspace{-0.5em}
\caption{Example \agent solution (numerical, MATH)}
\label{fig:husky-inference-numerical}
\end{figure}
\vspace{-1em}
\subsection{Tabular Reasoning}
\vspace{-0.5em}
\begin{figure}[h!]
    \centering
\tcbset{colback=verylightgray, colframe=verylightgray, boxrule=0.5pt, arc=4pt}
\begin{tcolorbox}
\lstset{
    basicstyle=\ttfamily\color{black}\tiny,
    keywordstyle=\color{black},
    identifierstyle=\color{black},
    commentstyle=\color{black},
    stringstyle=\color{black},
    breaklines=true,
    backgroundcolor=\color{verylightgray},
    frame=none,
    numbers=none,
    escapechar=@
}
\begin{lstlisting}
@\textbf{Question}@: undesignated hedges was $ 41.2 million and $ 42.1 million , respectively . the fair value of these hedging instruments in the company 2019s consolidated balance sheets as of october 29 , 2011 and october 30 , 2010 was immaterial . interest rate exposure management 2014 on june 30 , 2009 , the company entered into interest rate swap transactions related to its outstanding 5.0% ( 5.0 % ) senior unsecured notes where the company swapped the notional amount of its $ 375 million of fixed rate debt at 5.0% ( 5.0 % ) into floating interest rate debt through july 1 , 2014 . ... (omitted) ... the total notional amounts of derivative instruments designated as hedging instruments as of october 29 , 2011 and october 30 , 2010 were $ 375 million of interest rate swap agreements accounted for as fair value hedges and $ 153.7 million and $ 139.9 million , respectively , of cash flow hedges denominated in euros , british pounds and analog devices , inc . notes to consolidated financial statements 2014 ( continued ) .
['statement of income classification', 'statement of income loss on swaps', 'statement of income gain on note', 'statement of income net income effect', 'statement of income gain on swaps', 'loss on note', 'net income effect']
['other income', '$ -4614 ( 4614 )', '$ 4614', '$ 2014', '$ 20692', '$ -20692 ( 20692 )', '$ 2014']
what is the percentage change in cash flow hedges in 2011 compare to the 2010?
@\textbf{Solution}@:
Step 1: [commonsense] Identify the notional amounts of cash flow hedges for 2011 and 2010.
The notional amounts of cash flow hedges for 2011 and 2010 are $153.7 million and $139.9 million, respectively.

Step 2: [code] Calculate the percentage change in cash flow hedges from 2010 to 2011.
```python
cash_flow_hedges_2010 = 139.9
cash_flow_hedges_2011 = 153.7
percentage_change = ((cash_flow_hedges_2011 - cash_flow_hedges_2010) / cash_flow_hedges_2010) * 100
# print the output
print(percentage_change)
```
The percentage change in cash flow hedges from 2010 to 2011 is 9.8642.

The answer is: 9.8642 percent.
\end{lstlisting}
\end{tcolorbox}
\vspace{-0.5em}
\caption{Example \agent solution (tabular, FinQA)}
\label{fig:husky-inference-tabular}
\end{figure}
\subsection{Knowledge-based Reasoning}
\begin{figure}[h!]
    \centering
\tcbset{colback=verylightgray, colframe=verylightgray, boxrule=0.5pt, arc=4pt}
\begin{tcolorbox}
\lstset{
    basicstyle=\ttfamily\color{black}\tiny,
    keywordstyle=\color{black},
    identifierstyle=\color{black},
    commentstyle=\color{black},
    stringstyle=\color{black},
    breaklines=true,
    backgroundcolor=\color{verylightgray},
    frame=none,
    numbers=none,
    escapechar=@
}
\begin{lstlisting}
@\textbf{Question}@: When was the company that built the first steam locomotive to carry passengers on a public rail line founded?
@\textbf{Solution}@:
Step 1: [search] Identify the company that built the first steam locomotive to carry passengers on a public rail line.
```google
company that built the first steam locomotive to carry passengers on a public rail line
```
The first steam locomotive to carry passengers on a public rail line was built by Robert Stephenson and Company.

Step 2: [search] Find when Robert Stephenson and Company was founded.
```google
when was Robert Stephenson and Company founded
```
Robert Stephenson and Company was founded in 1823.

The answer is: 1823.
\end{lstlisting}
\end{tcolorbox}
\vspace{-0.5em}
\caption{Example \agent solution (knowledge, Bamboogle)}
\label{fig:husky-inference-knowledge}
\end{figure}
\subsection{Mixed Reasoning}
\begin{figure}[h!]
    \centering
\tcbset{colback=verylightgray, colframe=verylightgray, boxrule=0.5pt, arc=4pt}
\begin{tcolorbox}
\lstset{
    basicstyle=\ttfamily\color{black}\tiny,
    keywordstyle=\color{black},
    identifierstyle=\color{black},
    commentstyle=\color{black},
    stringstyle=\color{black},
    breaklines=true,
    backgroundcolor=\color{verylightgray},
    frame=none,
    numbers=none,
    escapechar=@
}
\begin{lstlisting}
@\textbf{Question}@: A scientist has a 10 kg block of aluminum and decides to melt it down to create a mixture with zinc, aiming for the final mixture to have a melting point exactly halfway between the melting points of pure aluminum and pure zinc. What weight of zinc (in kg) must the scientist add to the aluminum to achieve this goal, assuming the melting point of the mixture can be linearly interpolated between the melting points of pure aluminum and zinc?
(Note: This question simplifies the complexities of alloy formation and assumes the melting point between mixtures of aluminum and zinc can be linearly interpolated based on their proportions by weight.)
@\textbf{Solution}@:
Step 1: [search] Find the melting point of aluminum in Celsius.
```google
aluminum melting point
```
The melting point of aluminum is 660 degrees Celsius.

Step 2: [search] Find the melting point of zinc in Celsius.
```google
zinc melting point
```
The melting point of zinc is 419.5 degrees Celsius.

Step 3: [math] Compute the desired melting point of the mixture.
To find the desired melting point of the mixture, we need to find the midpoint between the melting points of pure aluminum and pure zinc. The melting point of aluminum is 660 degrees Celsius, and the melting point of zinc is 419.5 degrees Celsius.
The midpoint between two numbers can be found by adding the two numbers together and dividing by 2. Therefore, the desired melting point of the mixture is:
\\[\\frac{660 + 419.5}{2} = \\frac{1079.5}{2} = 539.75\\]
Thus, the desired melting point of the mixture is $\\boxed{539.75}$ degrees Celsius.

Step 4: [math] Determine the proportion of zinc required to achieve the desired melting point.
To determine the proportion of zinc required to achieve the desired melting point, we can use the concept of weighted averages. The formula for weighted averages is given by:
\\[\\text{Average} = \\frac{w_1 \\cdot x_1 + w_2 \\cdot x_2}{w_1 + w_2}\\]
where $w_1$ and $w_2$ are the weights of the two substances, and $x_1$ and $x_2$ are their respective values (in this case, melting points). We want the average melting point of the mixture to be 539.75 degrees Celsius, and we know the melting points of aluminum and zinc are 660 and 419.5 degrees Celsius, respectively. Let $w_1$ be the weight of aluminum (10 kg) and $w_2$ be the weight of zinc to be added. We can set up the equation:
\\[539.75 = \\frac{10 \\cdot 660 + w_2 \\cdot 419.5}{10 + w_2}\\]
Solving for $w_2$, we get:
\\[539.75(10 + w_2) = 6600 + 419.5w_2\\]
\\[5397.5 + 539.75w_2 = 6600 + 419.5w_2\\]
\\[120.25w_2 = 1202.5\\]
\\[w_2 = \\frac{1202.5}{120.25} = 10\\]
Therefore, the scientist must add $\\boxed{10}$ kg of zinc to achieve the desired melting point.
\end{lstlisting}
\end{tcolorbox}
\vspace{-0.5em}
\caption{Example \agent solution (mixed, \dataset)}
\label{fig:husky-inference-mixed}
\end{figure}
\section{Tool-Integrated Solution Trajectory Prompts}
\label{sec:solution-trajectory-prompts}
We provide examples of few-shot prompts for the set of training tasks used for generating the tool-integrated solution trajectories in Figures~\ref{fig:solution-trajectory-math-1},~\ref{fig:solution-trajectory-math-2},~\ref{fig:solution-trajectory-math-3},~\ref{fig:solution-trajectory-tabular-1},~\ref{fig:solution-trajectory-tabular-2},~\ref{fig:solution-trajectory-tabular-3},~\ref{fig:solution-trajectory-knowledge-1} and~\ref{fig:solution-trajectory-knowledge-2}.
\begin{figure}[h!]
    \centering
\tcbset{colback=verylightgray, colframe=verylightgray, boxrule=0.5pt, arc=4pt}
\begin{tcolorbox}
\lstset{
    basicstyle=\ttfamily\color{black}\tiny,
    keywordstyle=\color{black},
    identifierstyle=\color{black},
    commentstyle=\color{black},
    stringstyle=\color{black},
    breaklines=true,
    backgroundcolor=\color{verylightgray},
    frame=none,
    numbers=none,
    escapechar=@
}
\begin{lstlisting}
Question: A bowling ball cannot weigh more than 16 pounds and must have a diameter of $8 \\frac{1}{2}$ inches. How many square inches are in the surface area of a bowling ball before the finger holes are drilled? Express your answer as a common fraction in terms of $\\pi$.
Solution history: 
None
Next solution:
Step 1: Compute the radius of the bowling ball.
Tool: [code]
```python
import sympy
# define the diameter
diameter = sympy.Rational(17, 2)
# compute the radius of the bowling ball
radius = diameter / 2
# print the output
out = radius
print(out)
```
---
Question: Tom has a red marble, a green marble, a blue marble, and three identical yellow marbles. How many different groups of two marbles can Tom choose?
Solution history:
Step 1: Compute the number of ways Tom can choose two yellow marbles.
Tool: [math]
Since the yellow marbles are identical, there is only \\boxed{1} way for Tom to choose two yellow marbles.

Step 2: Compute the number of ways Tom can choose two marbles of different colors.
Tool: [code]
```python
from math import comb
# define the number of colors and the number of ways to choose 2 colors
num_colors = 4
num_ways = comb(num_colors, 2)
# print the output
out = num_ways
print(out)
```
```output
6
```
There are 6 ways Tom can choose two marbles of different colors.
Next solution:
Step 3: Compute the total number of different groups of two marbles Tom can choose.
Tool: [math]
The total number of distinct pairs of marbles Tom can choose is $1+6= \\boxed{7}$.
---
Question: The points $(9, -5)$ and $(-3, -1)$ are the endpoints of a diameter of a circle. What is the sum of the coordinates of the center of the circle?
Solution history:
Step 1: Compute the coordinates of the center of the circle.
Tool: [code]
```python
# define the coordinates of the endpoints
x1, y1 = 9, -5
x2, y2 = -3, -1
# compute the coordinates of the center of the circle
center_x = (x1 + x2) / 2
center_y = (y1 + y2) / 2
# print the output
out = (center_x, center_y)
print(out)
```
```output
(3.0, -3.0)
```
The coordinates of the center of the circle are (3.0, -3.0).
Next solution:
\end{lstlisting}
\end{tcolorbox}
\vspace{-0.5em}
\caption{Few-shot prompt for GSM-8K/MATH solution trajectories (1)}
\label{fig:solution-trajectory-math-1}
\end{figure}

\begin{figure}[h!]
    \centering
\tcbset{colback=verylightgray, colframe=verylightgray, boxrule=0.5pt, arc=4pt}
\begin{tcolorbox}
\lstset{
    basicstyle=\ttfamily\color{black}\tiny,
    keywordstyle=\color{black},
    identifierstyle=\color{black},
    commentstyle=\color{black},
    stringstyle=\color{black},
    breaklines=true,
    backgroundcolor=\color{verylightgray},
    frame=none,
    numbers=none,
    escapechar=@
}
\begin{lstlisting}
Step 2: Compute the sum of the coordinates of the center of the circle.
Tool: [code]
```python
# define the coordinates of the center of the circle (based on outputs from step 1)
x_center, y_center = 3, -3
# compute the sum of the coordinates
sum_coordinates = x_center + y_center
# print the output
out = sum_coordinates
print(out)
```
---
Question: Find the number of $x$-intercepts on the graph of $y = \\sin \\frac{1}{x}$ (evaluated in terms of radians) in the interval $(0.0001, 0.001).$
Solution history:
Step 1: Identify the conditions for the function to have $x$-intercepts.
Tool: [math]
The intercepts occur where $\\sin \\frac{1}{x}= 0$, that is, where $x = \\frac{1}{k\\pi}$ and $k$ is a nonzero integer. $\\boxed{x = \\frac{1}{k\\pi}}$

Step 2: Solve the inequality for the given interval.
Tool: [math]
Solving \\[0.0001 < \\frac{1}{k\\pi} < 0.001\\] yields \\[\\frac{1000}{\\pi} < k < \\frac{10{,}000}{\\pi}.\\] $\\boxed{\\frac{1000}{\\pi} < k < \\frac{10{,}000}{\\pi}}$

Step 3: Compute the number of $x$-intercepts in the given interval.
Tool: [code]
```python
# define the lower and upper bounds
lower_bound = 1000/sympy.pi
upper_bound = 10000/sympy.pi
# compute the number of x-intercepts
num_intercepts = math.floor(upper_bound) - math.ceil(lower_bound) + 1
# print the output
out = num_intercepts
print(out)
```
```output
2865
```
There are 2865 $x$-intercepts in the interval $(0.0001, 0.001).$
Next solution:
The answer is: <answer>2865</answer>.
---
Question: A rectangular band formation is a formation with $m$ band members in each of $r$ rows, where $m$ and $r$ are integers. A particular band has less than 100 band members. The director arranges them in a rectangular formation and finds that he has two members left over. If he increases the number of members in each row by 1 and reduces the number of rows by 2, there are exactly enough places in the new formation for each band member. What is the largest number of members the band could have?
Solution history: 
None
Next solution:
Step 1: Write equations based on the given information, using $m$ to represent the number of members in each row and $r$ to represent the number of rows.
Tool: [math]
The total number of band members can be represented by $mr + 2$. When the director increases the number of members in each row by 1 and reduces the number of rows by 2, the total becomes $(m+1)(r-2)$. The equation is $\\boxed{mr + 2 = (m+1)(r-2)}$.
---
Question: Let \\[f(x) = \\left\\{\n\\begin{array}{cl} ax+3, &\\text{ if }x>2, \\\\\nx-5 &\\text{ if } -2 \\le x \\le 2, \\\\\n2x-b &\\text{ if } x <-2.\n\\end{array}\n\\right.\\]Find $a+b$ if the piecewise function is continuous (which means that its graph can be drawn without lifting your pencil from the paper).
Solution history:
\end{lstlisting}
\end{tcolorbox}
\vspace{-0.5em}
\caption{Few-shot prompt for GSM-8K/MATH solution trajectories (2)}
\label{fig:solution-trajectory-math-2}
\end{figure}

\begin{figure}[h!]
    \centering
\tcbset{colback=verylightgray, colframe=verylightgray, boxrule=0.5pt, arc=4pt}
\begin{tcolorbox}
\lstset{
    basicstyle=\ttfamily\color{black}\tiny,
    keywordstyle=\color{black},
    identifierstyle=\color{black},
    commentstyle=\color{black},
    stringstyle=\color{black},
    breaklines=true,
    backgroundcolor=\color{verylightgray},
    frame=none,
    numbers=none,
    escapechar=@
}
\begin{lstlisting}
Step 1: Determine the value of $a$ by setting the first two pieces of the function equal to each other at $x=2$.
Tool: [code]
```python
from sympy import symbols, Eq, solve
# define the variables from the given question
x, a, b = symbols('x a b')
# define the first two pieces of the function
piece1 = a*x + 3
piece2 = x - 5
# set them equal to each other at x=2
eq = Eq(piece1.subs(x, 2), piece2.subs(x, 2))
# solve for a
solution_a = solve(eq, a)
# print the output
out = solution_a[0]
print(out)
```
```output
-3
```
The value of $a$ is -3.
Next solution:
Step 2: Determine the value of $b$ by setting the last two pieces of the function equal to each other at $x=-2$.
Tool: [code]
```python
from sympy import symbols, Eq, solve
# define the variables from the given question
x, b = symbols('x b')
# define the equation
eq = Eq(x-5, 2*x-b)
# substitute x = -2 in the equation
eq = eq.subs(x, -2)
# solve the equation for b
b_value = solve(eq, b)[0]
# print the output
out = b_value
print(out)
```
---
\end{lstlisting}
\end{tcolorbox}
\vspace{-0.5em}
\caption{Few-shot prompt for GSM-8K/MATH solution trajectories (3)}
\label{fig:solution-trajectory-math-3}
\end{figure}

\begin{figure}[h!]
    \centering
\tcbset{colback=verylightgray, colframe=verylightgray, boxrule=0.5pt, arc=4pt}
\begin{tcolorbox}
\lstset{
    basicstyle=\ttfamily\color{black}\tiny,
    keywordstyle=\color{black},
    identifierstyle=\color{black},
    commentstyle=\color{black},
    stringstyle=\color{black},
    breaklines=true,
    backgroundcolor=\color{verylightgray},
    frame=none,
    numbers=none,
    escapechar=@
}
\begin{lstlisting}
Question: ['Stem', 'Leaf ']
['3', ['3', '3', '3', '5', '5']]
['4', ['6']]
['5', ['4', '5', '7', '8']]
['6', ['7', '8']]
['7', ['2', '3', '7', '9']]
['8', ['6', '8', '9']]
The members of the local garden club tallied the number of plants in each person's garden. How many gardens have at least 47 plants?
Solution history:
None
Next solution:
Step 1: Convert the Stem and Leaf plot into a list of numbers.
Tool: [code]
```python
def stem_leaf_to_list(stem_leaf_plot):
    stem_leaf_plot = "\n".split(stem_leaf_plot)
    result = []
    for row in stem_leaf_plot[1:]:  # Skip the header row
        stem = row[0]
        leaves = row[1]
        for leaf in leaves:
            number = int(stem + leaf)  # Concatenate stem and leaf, then convert to integer
            result.append(number)
    return result
stem_leaf_plot_input = "['Stem', 'Leaf ']\n['3', ['3', '3', '3', '5', '5']]\n['4', ['6']]\n['5', ['4', '5', '7', '8']]\n['6', ['7', '8']]\n['7', ['2', '3', '7', '9']]\n['8', ['6', '8', '9']]"
out = st''em_leaf_to_list(stem_leaf_plot_input)
print(out)
```
---
\end{lstlisting}
\end{tcolorbox}
\vspace{-0.5em}
\caption{Few-shot prompt for TabMWP solution trajectories (1)}
\label{fig:solution-trajectory-tabular-1}
\end{figure}

\begin{figure}[h!]
    \centering
\tcbset{colback=verylightgray, colframe=verylightgray, boxrule=0.5pt, arc=4pt}
\begin{tcolorbox}
\lstset{
    basicstyle=\ttfamily\color{black}\tiny,
    keywordstyle=\color{black},
    identifierstyle=\color{black},
    commentstyle=\color{black},
    stringstyle=\color{black},
    breaklines=true,
    backgroundcolor=\color{verylightgray},
    frame=none,
    numbers=none,
    escapechar=@
}
\begin{lstlisting}
Question: ['Donation level', 'Number of donors']
['Gold', '15']
['Silver', '68']
['Bronze', '58']
The Burlington Symphony categorizes its donors as gold, silver, or bronze depending on the amount donated. What fraction of donors are at the bronze level? Simplify your answer.
Solution history:
Step 1: Identify the number of donors at the bronze level.
Tool: [commonsense]
According to the table, there are 58 donors at the bronze level.

Step 2: Identify the total number of donors.
Tool: [code]
```python
num_gold = 15
num_silver = 68
num_bronze = 58
num_total = num_gold + num_silver + num_bronze
print(num_total)
```
```output
141
```
There are a total of 141 donors.
Next solution:
Step 3: Compute the fraction of donors at the bronze level.
Tool: [code]
```python
from fractions import Fraction
num_bronze = 58
num_total = 141
fraction_bronze = Fraction(num_bronze, num_total).limit_denominator()
print(fraction_bronze)
```
---
Question: ['Day', 'Number of tickets']
['Friday', '71']
['Saturday', '74']
['Sunday', '75']
['Monday', '72']
The transportation company tracked the number of train tickets sold in the past 4 days. On which day were the fewest train tickets sold?
Solution history:
None
Next solution:
Step 1: Identify the day the fewest train tickets were sold.
Tool: [commonsense]
According to the table, the smallest number of tickets sold on a particular day is 71. The corresponding day is Friday.
---
Question: '[\'Column 1\', \'Column 2\']
[\'shiny metal bead\', \'$0.04\']
[\'shiny red bead\', \'$0.05\']
[\'star-shaped silver bead\', \'$0.05\']
["brown cat\'s eye bead", \'$0.09\']
[\'orange glass bead\', \'$0.02\']
[\'round silver bead\', \'$0.04\']
Emmy has $0.05. Does she have enough to buy an orange glass bead and a round silver bead?
Solution history:
Step 1: Identify the price of an orange glass bead.
Tool: [commonsense]
According to the table, an orange glass bead costs $0.02.

Step 2: Identify the price of a round silver bead.
Tool: [commonsense]
According to the table, a round silver bead costs $0.04.
Next solution:
Step 3: Determine whether Emmy has enough to buy an orange glass bead and a round silver bead.
Tool: [code]
```python
emmy_budget = 0.05
price_orange_glass_bead = 0.02
price_round_silver_bead = 0.04
emmy_cost = price_orange_glass_bead + price_round_silver_bead
emmy_leftover = emmy_budget - emmy_cost
if emmy_leftover >= 0:
    print("Emmy has enough money.")
else:
    print("Emmy does not have enough money.")
```
---
\end{lstlisting}
\end{tcolorbox}
\vspace{-0.5em}
\caption{Few-shot prompt for TabMWP solution trajectories (2)}
\label{fig:solution-trajectory-tabular-2}
\end{figure}

\begin{figure}[h!]
    \centering
\tcbset{colback=verylightgray, colframe=verylightgray, boxrule=0.5pt, arc=4pt}
\begin{tcolorbox}
\lstset{
    basicstyle=\ttfamily\color{black}\tiny,
    keywordstyle=\color{black},
    identifierstyle=\color{black},
    commentstyle=\color{black},
    stringstyle=\color{black},
    breaklines=true,
    backgroundcolor=\color{verylightgray},
    frame=none,
    numbers=none,
    escapechar=@
}
\begin{lstlisting}
Question: ['Month', 'Number of magazines']
['November', '3,874']
['December', '1,300']
['January', '7,828']
['February', '5,211']
A magazine editor looked at her sales figures for the past 4 months. How many more copies were sold in February than in November?
Solution history:
Step 1: Identify the number of magazine copies sold in November.
Tool: [commonsense]
According to the table, 3,874 copies were sold in November.

Step 2: Identify the number of magazine copies sold in February.
Tool: [commonsense]
According to the table, 5,211 copies were sold in February.

Step 3: Compute the difference between the number of copies sold in November and February.
Tool: [code]
```python
copies_november = 3874
copies_february = 5211
diff_copies = copies_february - copies_november
print(diff_copies)
```
```output
1337
```
Next solution:
The answer is: <answer>1337 copies</answer>.
---
Question: ['Column 1', 'Column 2']
['2', '8']
['3', '12']
['4', '16']
['5', '20']
Terrence likes to make lemonade using the lemons from his grandmother's trees. There is a proportional relationship between the number of pitchers of lemonade Terrence wants to make, x, and the number of lemons he needs, y. What is the constant of proportionality? Write your answer as a whole number or decimal.
Solution history:
None
Next solution:
Step 1: Identify the proportional constant between the pairs of numbers in the table.
Tool: [code]
```python
# copy the table from the question
table_str = "['Column 1', 'Column 2']
['2', '8']
['3', '12']
['4', '16']
['5', '20']"
# convert the string into a list of lists
table_list = "\n".split(table_str)
# exclude the headers
table_list = table_list[1:]
# iterate over the table and identify the proportional constant
prop = set()
for xy_pair in table_list:
    x, y = xy_pair[0], xy_pair[1]
    prop.add(y / x)
print(prop)
```
---
\end{lstlisting}
\end{tcolorbox}
\vspace{-0.5em}
\caption{Few-shot prompt for TabMWP solution trajectories (3)}
\label{fig:solution-trajectory-tabular-3}
\end{figure}

\begin{figure}[h!]
    \centering
\tcbset{colback=verylightgray, colframe=verylightgray, boxrule=0.5pt, arc=4pt}
\begin{tcolorbox}
\lstset{
    basicstyle=\ttfamily\color{black}\tiny,
    keywordstyle=\color{black},
    identifierstyle=\color{black},
    commentstyle=\color{black},
    stringstyle=\color{black},
    breaklines=true,
    backgroundcolor=\color{verylightgray},
    frame=none,
    numbers=none,
    escapechar=@
}
\begin{lstlisting}
Question: When did the ocean surrounding Antarctica come into existence?
Solution history:
None
Next solution:
Step 1: Identify the ocean surrounding Antarctica.
Tool: [search]
```google
the ocean surrounding Antarctica
```
---
Question: Who was the first president from the party following the Anti-Administration Party?
Solution history:
Step 1: Find which party followed the Anti-Administration Party.
Tool: [search]
```google
the party that followed the Anti-Administration Party
```
```output
Democratic-Republican Party - Wikipedia
```
The Democratic-Republican Party followed the Anti-Administration Party.

Step 2: Identify the first president of the Democratic-Republican Party.
Tool: [search]
```google
the first president from the Democratic-Republican Party
```
```output
Thomas Jefferson defeated John Adams in the 1800 presidential election, thereby becoming the first Democratic-Republican president.
```
Thomas Jefferson was the first president from the Democratic-Republican Party.
Next solution:
Step 3: Return the answer to the question based on the outputs above.
Tool: [commonsense]
The first president from the party following the Anti-Administration Party is <output>Thomas Jefferson</output>.
---
Question: In which county is the Manning Bowl located?
Solution history:
None
Next solution:
Step 1: Find which county Manning Bowl is located in.
Tool: [search]
```google
the county is the Manning Bowl located in
```
---
Question: What movie stars Morgan Freeman, Robert De Niro and the producer of Flatliners?
Solution history:
Step 1: Identify the producer of Flatliners.
Tool: [search]
```google
the producer of Flatliners
```
```output
Rick Bieber, Michael Douglas, Scott Rudin
```
The producers of Flatliners are Michael Douglas, Scott Rudin and Rick Bieber.

Step 2: Find which movie Morgan Freeman, Robert Deniro and Rick Bieber starred in.
Tool: [search]
```google
movies that Morgan Freeman, Robert Deniro and Rick Bieber star in
```
```output
Last Vegas showcases the first-time onscreen collaboration of four legendary actors: Michael Douglas, Robert De Niro, Morgan Freeman, and Kevin Kline. The ...
```
The search result above does not mention Rick Bieber, so there is no movie where Morgan Freeman, Robert Deniro and Rick Bieber starred in together.
\end{lstlisting}
\end{tcolorbox}
\vspace{-0.5em}
\caption{Few-shot prompt for MusiQue solution trajectories (1)}
\label{fig:solution-trajectory-knowledge-1}
\end{figure}

\begin{figure}[h!]
    \centering
\tcbset{colback=verylightgray, colframe=verylightgray, boxrule=0.5pt, arc=4pt}
\begin{tcolorbox}
\lstset{
    basicstyle=\ttfamily\color{black}\tiny,
    keywordstyle=\color{black},
    identifierstyle=\color{black},
    commentstyle=\color{black},
    stringstyle=\color{black},
    breaklines=true,
    backgroundcolor=\color{verylightgray},
    frame=none,
    numbers=none,
    escapechar=@
}
\begin{lstlisting}
Step 3: Find which movie Morgan Freeman, Robert Deniro and Michael Douglas starred in.
Tool: [search]
```google
movies that Morgan Freeman, Robert Deniro and Michael Douglas star in
```
```output
Last Vegas is a 2013 American comedy film directed by Jon Turteltaub, written by Dan Fogelman and starring Michael Douglas, Robert De Niro, Morgan Freeman, ...
```
Morgan Freeman, Robert Deniro and Michael Douglas star in Last Vegas.

Step 4: Return the answer to the question based on the outputs above.
Tool: [commonsense]
<output>Last Vegas</output> stars Morgan Freeman, Robert De Niro and the producer of Flatliners.

Next solution:
The answer is: <answer>Last Vegas</answer>.
---
Question: When was the abolishment of the studio that distributed The Game?
Solution history:
Step 1: Identify the studio that distributed The Game.
Tool: [search]
```google
the studio that distributed the movie The Game
```
```output
After a wealthy San Francisco banker is given an opportunity to participate in a mysterious game, his life is turned upside down as he begins to question if it ...
```
I'm sorry, but it seems there was no information provided in the output regarding the studio that distributed The Game. Therefore, I cannot complete the answer based on the given output.
Next solution:
Step 1: Search again with a clearer query to identify the studio that distributed the movie "The Game."
Tool: [search]
```google
the Game 1997 movie distributor
```
---
Question: What percentage of households in New York City earns more than $200,000 every year?
Solution history:
Step 1: Find the number of households in New York City.
Tool: [search]
```google
number of households in New York City
```
```output
3.192 million
```
There are 3.192 million households in New York City.

Step 2: Find the number of households in New York City with income over $200,000 per year.
Tool: [search]
```google
number of households in New York City with income over $200,000 per year
```
```output
In 2020, there were about 386,292 households with an income of 200,000 U.S. dollars or more in New York City. In comparison, approximately 263,396 households had an income of less than 10,000 U.S. dollars during the same year.
```
There were 386,292 households with income over $200,000 per year in 2020.
Next solution:
Step 3: Compute the percentage of households that earn more than $200,000 in New York City.
Tool: [code]
```python
num_households_nyc = 3192000
num_households_over_200k_nyc = 386292
percentage = (num_households_over_200k_nyc / num_households_nyc) * 100
print(percentage)
```
---
\end{lstlisting}
\end{tcolorbox}
\vspace{-0.5em}
\caption{Few-shot prompt for MusiQue solution trajectories (2)}
\label{fig:solution-trajectory-knowledge-2}
\end{figure}

\end{document}